\documentclass{book}
\usepackage{roboto}

\usepackage[english]{babel}
\usepackage[letterpaper,top=2cm,bottom=2cm,left=3cm,right=3cm,marginparwidth=1.75cm]{geometry}

\usepackage{pdflscape}
\usepackage{lscape}
\usepackage{epigraph}

\usepackage{rotating}
\usepackage{longtable}
\usepackage{array}
\usepackage{float}
\usepackage{amsmath}
\usepackage{graphicx}
\usepackage{inconsolata}
\usepackage[colorlinks=true, allcolors=blue]{hyperref}
\usepackage{enumitem}

\usepackage{tikz} 

\usepackage{listings}
\usepackage{xcolor}

\usepackage[T1]{fontenc}
\usepackage{inconsolata}  

\definecolor{bgcolor}{rgb}{0.97,0.97,0.97}
\definecolor{codeblue}{rgb}{0.1,0.1,0.8}
\definecolor{codegreen}{rgb}{0,0.4,0}
\definecolor{codegray}{rgb}{0.4,0.4,0.4}
\definecolor{codepurple}{rgb}{0.5,0,0.5}
\definecolor{codered}{rgb}{0.6,0.2,0.2}
\definecolor{lightgray}{rgb}{0.9,0.9,0.9}
\definecolor{darkgray}{rgb}{0.6,0.6,0.6} 

\makeatletter
\renewcommand{\paragraph}{%
  \@startsection{paragraph}{4}{\z@}{1ex}{-1em}{\normalfont\normalsize\bfseries\color{gray}}}
\makeatother

\lstdefinestyle{python}{
    language=Python,
    basicstyle=\ttfamily\small\color{black}\usefont{T1}{zi4}{m}{n},  
    keywordstyle=\bfseries\color{codeblue},  
    stringstyle=\color{codegreen},  
    commentstyle=\slshape\color{codegray},  
    showstringspaces=false,
    numbers=left,
    numberstyle=\tiny\color{codegray},  
    stepnumber=1,
    numbersep=8pt,
    frame=single,
    rulecolor=\color{darkgray},  
    breaklines=true,
    backgroundcolor=\color{bgcolor},
    tabsize=4,
    captionpos=b,
    morekeywords={self}, 
}

\lstdefinestyle{text}{
    language=,
    basicstyle=\ttfamily\small\color{black}\usefont{T1}{zi4}{m}{n},  
    stringstyle=\color{codered},
    commentstyle=\color{codegray},
    showstringspaces=false,
    numbers=none,
    frame=single,
    rulecolor=\color{lightgray},  
    frameround=tttt,
    breaklines=true,
    backgroundcolor=\color{bgcolor},
    tabsize=4,
    captionpos=b,
}

\lstdefinestyle{cmd}{
    language=bash,
    basicstyle=\ttfamily\small\color{black}\usefont{T1}{zi4}{m}{n},  
    keywordstyle=\bfseries\color{blue},
    stringstyle=\color{codegreen},
    commentstyle=\itshape\color{gray},
    showstringspaces=false,
    numbers=none,
    frame=single,
    rulecolor=\color{darkgray},  
    breaklines=true,
    backgroundcolor=\color{bgcolor},
    tabsize=4,
    captionpos=b,
}

\lstdefinestyle{sql}{
    language=SQL,  
    basicstyle=\ttfamily\small\color{black}\usefont{T1}{zi4}{m}{n},  
    keywordstyle=\bfseries\color{codeblue},  
    stringstyle=\color{codegreen},  
    commentstyle=\slshape\color{codegray},  
    showstringspaces=false,  
    numbers=left,  
    numberstyle=\tiny\color{codegray},  
    stepnumber=1,  
    numbersep=8pt,  
    frame=single,  
    rulecolor=\color{darkgray},  
    breaklines=true,  
    backgroundcolor=\color{bgcolor},  
    tabsize=4,  
    captionpos=b,  
    morekeywords={
        SELECT, INSERT, DELETE, UPDATE, FROM, WHERE, AND, OR, JOIN, ON, CREATE, INDEX, TABLE, VALUES, INTO, AS, DISTINCT, ORDER, BY, GROUP, HAVING, LIMIT, OFFSET, UNION, DROP, ALTER, TRUNCATE, RENAME, PRIMARY, FOREIGN, KEY, CONSTRAINT, NULL, NOT, DEFAULT, AUTO_INCREMENT, UNIQUE, CHECK,
        db, find, insertOne, updateOne, deleteOne, collection, aggregate, match, project, sort, limit, pipeline, insertMany, updateMany, deleteMany, findOne, $lookup, $group,
        GET, SET, DEL, HGET, HSET, HDEL, LPUSH, RPUSH, LPOP, RPOP, SADD, SREM, PUBLISH, SUBSCRIBE, EXPIRE, TTL, FLUSHDB, FLUSHALL, INCR, DECR,
        SELECT, INSERT, UPDATE, DELETE, FROM, WHERE, AND, OR, USE, KEYSPACE, CREATE, ALTER, DROP, TRUNCATE, TABLE, INDEX, PRIMARY, FOREIGN, KEY, WITH, CLUSTERING, ORDER, BY, ASC, DESC, LIMIT, BATCH, APPLY, TOKEN, CONSISTENCY, QUORUM, LOCAL_QUORUM, ANY, ALL, ONE, TWO, THREE,
        MATCH, CREATE, MERGE, RETURN, DELETE, DETACH, REMOVE, SET, FOREACH, UNWIND, WITH, ORDER, BY, ASCENDING, DESCENDING, SKIP, LIMIT, UNION, ALL, OPTIONAL, DISTINCT, WHERE, AND, OR, IN, STARTS, ENDS, CONTAINS, EXISTS, IS, NULL,
        SELECT, INSERT, DELETE, UPDATE, FROM, WHERE, AND, OR, JOIN, ON, CREATE, INDEX, TABLE, VALUES, INTO, AS, DISTINCT, ORDER, BY, GROUP, HAVING, LIMIT, OFFSET, UNION, DROP, ALTER, TRUNCATE, RENAME, PRIMARY, FOREIGN, KEY, CONSTRAINT, NULL, NOT, DEFAULT, AUTO_INCREMENT, UNIQUE, CHECK, SEQUENCE, SYNONYM, PACKAGE, FUNCTION, PROCEDURE, TRIGGER, VIEW, GRANT, REVOKE, COMMIT, ROLLBACK,
        SELECT, INSERT, DELETE, UPDATE, FROM, WHERE, AND, OR, JOIN, ON, CREATE, INDEX, TABLE, VALUES, INTO, AS, DISTINCT, ORDER, BY, GROUP, HAVING, LIMIT, OFFSET, UNION, DROP, ALTER, TRUNCATE, RENAME, PRIMARY, FOREIGN, KEY, CONSTRAINT, NULL, NOT, DEFAULT, AUTO_INCREMENT, UNIQUE, CHECK, SERIAL, BIGSERIAL, RETURNING, DO, LANGUAGE, PLPGSQL, BEGIN, END, IMMUTABLE, VOLATILE,
        SELECT, INSERT, DELETE, UPDATE, FROM, WHERE, AND, OR, JOIN, ON, CREATE, INDEX, TABLE, VALUES, INTO, AS, DISTINCT, ORDER, BY, GROUP, HAVING, LIMIT, OFFSET, UNION, DROP, ALTER, TRUNCATE, RENAME, PRIMARY, FOREIGN, KEY, CONSTRAINT, NULL, NOT, DEFAULT, AUTO_INCREMENT, UNIQUE, CHECK, INCREMENT, ENGINE, CHARSET, COLLATE, COMMENT
    }  
}

\lstdefinestyle{html}{
    language=HTML,
    basicstyle=\ttfamily\small\color{black}\usefont{T1}{zi4}{m}{n},  
    keywordstyle=\bfseries\color{codeblue},  
    stringstyle=\color{codegreen},  
    commentstyle=\slshape\color{codegray},  
    showstringspaces=false,
    numbers=left,
    numberstyle=\tiny\color{codegray},  
    stepnumber=1,
    numbersep=8pt,
    frame=single,
    rulecolor=\color{darkgray},  
    breaklines=true,
    backgroundcolor=\color{bgcolor},
    tabsize=4,
    captionpos=b,
    morekeywords={<!DOCTYPE,html,head,body,div,span,a,img,href,src,script,style}, 
}

\title{Deep Learning and Machine Learning, Advancing Big Data Analytics and Management: Unveiling AI's Potential Through Tools, Techniques, and Applications}
\author{
    Pohsun Feng\textsuperscript{*†} \\ 
    \textit{National Taiwan Normal University} \\
    41075018h@ntnu.edu.tw
    \and
    Ziqian Bi\textsuperscript{*†} \\ 
    \textit{Indiana University} \\
    bizi@iu.edu
    \and
    Yizhu Wen \\ 
    \textit{University of Hawaii} \\
    yizhuw@hawaii.edu
    \and
    Qian Niu \\ 
    \textit{Kyoto University} \\
    niu.qian.f44@kyoto-u.ac.jp
    \and
    Junyu Liu \\ 
    \textit{Kyoto University} \\
    liu.junyu.82w@st.kyoto-u.ac.jp
    \and
    Tianyang Wang \\ 
    \textit{Xi’an Jiaotong-Liverpool University} \\
    Tianyang.Wang21@student.xjtlu.edu.cn
    \and
    Ming Li \\ 
    \textit{Georgia Institute of Technology} \\
    mli694@gatech.edu
    \and
    Sen Zhang \\ 
    \textit{Rutgers University} \\
    sen.z@rutgers.edu
    \and
    Benji Peng \\ 
    \textit{AppCubic} \\
    benji@appcubic.com
    \and
    Ming Liu \\ 
    \textit{Purdue University} \\
    liu3183@purdue.edu
    \and
    Xuanhe Pan \\ 
    \textit{University of Wisconsin-Madison} \\
    xpan73@wisc.edu
    \and
    Caitlyn Heqi Yin \\ 
    \textit{University of Wisconsin-Madison} \\
    hyin66@wisc.edu
    \and
    Jiawei Xu \\ 
    \textit{Purdue University} \\
    xu1644@purdue.edu
    \and
    Jinlang Wang \\ 
    \textit{University of Wisconsin-Madison} \\
    jinlang.wang@wisc.edu
    \and
    Keyu Chen \\ 
    \textit{Georgia Institute of Technology} \\
    kchen637@gatech.edu
    \and
    Xinyuan Song \\
    \textit{Emory University} \\
    xsong30@emory.edu
    \and
    Zekun Jiang \\
    \textit{Sichuan University} \\
    zekun\_jiang@163.com
}

\date{}

\begin{document}

\maketitle

\begingroup
\renewcommand\thefootnote{}\footnote{
    \textsuperscript{*} Equal contribution \\
    \textsuperscript{$\dagger$} Corresponding author
}
\addtocounter{footnote}{0}
\endgroup

\epigraph{"The more you buy,The more you save."}{\textit{— Jensen Huang, CEO of NVIDIA}}

\tableofcontents  
\cleardoublepage

\setcounter{part}{1} 
\part{Getting Started}

\chapter{A Journey into Artificial Intelligence}

Welcome to the world of deep learning and machine learning! In this era full of infinite possibilities, artificial intelligence (AI) is no longer just a concept from science fiction but is becoming an integral part of our daily lives. From voice assistants in smartphones to self-driving cars, AI is redefining the world around us. If you're a beginner looking for a gateway into this vast field, this chapter will open a window to the history, applications, and future trends of AI, giving you an exciting overview of what's to come.

\section{The Rise of Machine Learning and Deep Learning}

\subsection{From Basics to Intelligence}

The history of machine learning \cite{peng2024deeplearningmachinelearning} dates back to the mid-20th century when scientists began to explore the idea of whether machines could learn to solve problems. What once sounded like a fantasy is now a reality. The core idea of machine learning is to enable computers to identify patterns in data and use them to make predictions, classify objects, or solve complex tasks. Deep learning, a branch of machine learning inspired by neural networks in the human brain, has taken this concept even further, allowing machines to handle more intricate and abstract problems.

\subsection{Deep Learning: The Driving Force of the Future}

You may have heard stories like AlphaGo~\cite{Silver2016} defeating human Go masters, or perhaps you use an AI-powered app every day. These advances are driven by deep learning. Since 2010, deep learning has achieved remarkable results in areas such as image recognition, speech recognition, and natural language processing, gaining widespread attention in both academia and industry. In many ways, deep learning has become the engine of AI, powering its rapid advancement.

\section{Applications Everywhere}

\subsection{The Power to Transform the World}

Machine learning and deep learning are transforming numerous industries and impacting our everyday lives in ways we never imagined:

\begin{itemize}
    \item \textbf{Smartphones and Daily Technology}: Facial recognition on your phone, as well as voice assistants that answer your questions, rely on powerful AI-driven image recognition and natural language processing technologies.
    \item \textbf{Autonomous Driving and Transportation}: Self-driving cars are becoming a reality, thanks to breakthroughs in machine learning, which enable them to perceive their environment, make decisions, and plan routes.
    \item \textbf{Healthcare Revolution}: AI is helping doctors analyze medical images, predict disease trends, and develop personalized treatment plans, elevating healthcare to new heights.
    \item \textbf{Financial Services}: In the financial sector, deep learning is used for fraud detection, market prediction, and providing personalized investment advice.
\end{itemize}

\subsection{The Future is Here}

Whether it's helping doctors diagnose diseases, assisting lawyers with legal document analysis, or optimizing supply chains with big data, the applications of machine learning and deep learning have already moved beyond the technology sector and are driving change across multiple industries.

\section{The Future of AI: The Next Technological Wave}

\subsection{Automated Machine Learning: Democratizing AI}

In the future, machine learning will no longer be the exclusive domain of experts. With the advent of automated machine learning (AutoML), even individuals without deep technical knowledge will be able to build AI models. This democratization of AI will lower the barriers to entry, allowing more people to participate in this technological revolution.

\subsection{Edge Computing: Bringing AI Closer to You}

As the Internet of Things (IoT) expands, AI will shift from the cloud to local devices, known as “edge computing.” This means that your smartwatch, home appliances, and even your car will have AI capabilities, providing faster and more personalized services without relying on the cloud.

\subsection{Explainability and Fairness in AI}

As AI becomes increasingly integrated into society, transparency and fairness in its decision-making processes will be critical. The future of AI research will focus on ensuring that AI systems are not only powerful but also understandable and unbiased, promoting fairness across different social groups.

\section{Visualization: Seeing the World Behind Data}

For beginners, data and models can often seem complex and difficult to understand. Visualization, however, serves as a key to unlocking the mysteries of AI. With graphical tools, complex algorithms, and data structures become more intuitive. For example, visualization tools help illustrate how a deep learning model extracts features from images or text. Visualization not only aids learning but is also an essential tool for debugging and optimizing models.

\section{The Beginner’s Journey in AI}

For those just starting, learning deep learning and machine learning may seem daunting, but with a clear path, the journey can be fun and rewarding:

\begin{enumerate}
    \item \textbf{Solidify Your Math Foundation}: Concepts like linear algebra, calculus, and probability are the bedrock of machine learning. You don't need to be a mathematician, but mastering these tools will significantly enhance your understanding.
    \item \textbf{Learn to Code}: Python is the most popular language in machine learning, and with libraries like TensorFlow and PyTorch, you can easily build deep learning models.
    \item \textbf{Start with Simple Models}: Classical algorithms like linear regression and decision trees are great starting points. Once you understand the basics, you can move on to more complex deep-learning models.
    \item \textbf{Hands-On Projects}: Theory is important, but hands-on experience is where real learning happens. Start with publicly available datasets and gradually build and optimize your own AI models.
\end{enumerate}

\section{Conclusion}

Deep learning and machine learning are reshaping the world, and you have the opportunity to be part of this transformation. As you embark on this learning journey, you’ll not only master cutting-edge technologies but also discover how to apply them creatively and meaningfully. Whether your goal is to solve real-world problems with AI or explore the frontiers of research, the future holds endless possibilities. Let’s step into this intelligent era together and embrace a brighter future!

\chapter{Making the Best Use of Tools}

In today's digital age, artificial intelligence (AI) tools have become widely adopted. These tools significantly boost productivity and help us achieve more efficient outcomes in big data management and analysis, machine learning, and deep learning. Particularly, natural language processing (NLP) advancements have made multimodal AI tools incredibly powerful. These tools can understand and generate natural language and analyze multimodal inputs (such as text, images, and code) to assist users in solving complex tasks.

The utility of these AI tools goes beyond just coding. They can be used to ask questions, explore ideas, confirm research directions, and solve technical issues. In big data management and machine learning projects, AI tools are invaluable for quickly exploring data, designing and optimizing models, and increasing the overall efficiency of research and development. This chapter will introduce several of the most popular AI tools today and explain how to make the best use of them to enhance your work in big data and machine learning.

\section{ChatGPT}

ChatGPT~\cite{OpenAI2023}, developed by OpenAI, is a powerful conversational AI tool~\cite{brown2020languagemodelsfewshotlearners}. It can not only understand and generate natural language but also assist developers in data analysis, model design, and code generation~\cite{chen2021evaluatinglargelanguagemodels}. ChatGPT excels in providing detailed suggestions based on context and continuously refines its responses through ongoing dialogue with the user. This makes ChatGPT a valuable tool for solving complex problems and determining research directions.

\subsection{Data Analysis and Direction Setting}

In the process of big data analysis, ChatGPT can assist users in understanding data distributions, selecting features, and performing data cleaning tasks. Users can inquire about specific analysis methods, the details of algorithmic models, or how to select the best model for processing their data. ChatGPT can also help users narrow down their scope of analysis and develop effective strategies by discussing and answering specific questions.

\subsection{Machine Learning and Deep Learning Model Design}

ChatGPT is highly effective in the field of machine learning and deep learning. It can help users understand the workings of different algorithms and provide suggestions on adjusting model parameters to improve performance. During model design, ChatGPT can offer advice on selecting activation functions, optimizers, and loss functions, ensuring that the constructed model is well-tuned and logical.

\subsection{Code Writing and Optimization}

Beyond direction setting and model design, ChatGPT can also generate code snippets based on user requirements. For example, if a user describes a machine learning algorithm, ChatGPT can provide the corresponding code implementation. Supporting multiple programming languages, it helps developers quickly prototype algorithms. Additionally, in terms of code optimization, ChatGPT can help simplify complex code structures and improve execution efficiency.

\section{Claude}

Claude, developed by Anthropic~\cite{Anthropic2023Claude}, is another powerful conversational AI tool. It is particularly well-suited for projects that require a high level of security and robustness, making it a strong fit for big data analysis and machine learning tasks. Claude not only helps with code generation but is also adept at performing data processing, analysis, and security review in model design.

\subsection{Big Data Processing and Analysis}

In big data management and analysis, data security and accuracy are paramount~\cite{ELMESTARI2024103605}. Claude can analyze the characteristics of datasets and help developers design more robust algorithms, ensuring privacy protection during data processing~\cite{Anomaly_detection}. Developers can use Claude to design distributed data processing pipelines, identify key features, and detect potential anomalies in the data.

\subsection{Machine Learning Model Optimization and Tuning}

Claude stands out in model optimization and tuning, especially when aiming for robustness. It helps users understand how different algorithms apply to large-scale data and offers suggestions for parameter tuning, ensuring the generated models generalize well~\cite{madry2019deeplearningmodelsresistant}. When working with private or sensitive data, Claude's security insights are invaluable in ensuring the model is compliant and secure.

\section{Gemini}

Gemini, developed by Google DeepMind, is a multimodal AI tool that combines powerful language understanding with the ability to process multiple input types~\cite{DeepMind2023Gemini}. Gemini not only assists with code generation and optimization but also helps users perform big data analysis and multimodal reasoning, making it an ideal tool for complex data analysis and model design~\cite{baltrušaitis2017multimodalmachinelearningsurvey}.

\subsection{Multimodal Reasoning and Analysis}

Gemini's greatest strength lies in its multimodal reasoning capability. In big data analysis and machine learning projects, users can input textual descriptions, code snippets, or images, and Gemini will combine this information for deeper reasoning and analysis. This multimodal capability is particularly advantageous when working with diverse data or when designing models for complex, cross-disciplinary tasks.

\subsection{Deep Learning Model Design and Optimization}

For deep learning models, especially in image processing and natural language processing, Gemini assists users in designing and optimizing neural network structures. Users can seek advice on model architectures (such as convolutional neural networks, recurrent neural networks, or transformer models), and Gemini will provide suggestions for suitable architectures based on the input. It also helps with hyperparameter tuning to improve model performance.

\section{Other Multimodal AI Tools}

In addition to ChatGPT, Claude, and Gemini, there are many other multimodal conversational AI tools available that can also assist developers and data scientists in big data management, machine learning, and deep learning.

\subsection{CodeWhisperer}

CodeWhisperer, developed by Amazon, is an AI code assistant integrated into IDEs such as Visual Studio Code. It is particularly useful for cloud computing and machine learning tasks on big data~\cite{AmazonQ}. CodeWhisperer not only offers contextual code completions but is also seamlessly integrated with AWS services, making it an ideal choice for developers working on cloud-based machine learning and data processing tasks.

\subsection{Copilot}

GitHub Copilot, powered by OpenAI Codex, is an AI code assistant that can help write and optimize code, especially in complex machine learning projects~\cite{GitHub2021Copilot}. For data scientists, Copilot simplifies many time-consuming tasks, such as writing model code, by offering real-time code suggestions tailored to the user’s coding habits.

\subsection{Replit Ghostwriter}

Replit Ghostwriter is an AI code assistant on the Replit platform, offering online programming and real-time data analysis~\cite{Replit2023Ghostwriter}. It is particularly useful for interactive machine-learning experiments and coding within a browser. Ghostwriter provides instant feedback, helping users optimize model code and visualize data analysis on the fly.

\section{Conclusion}

The use of AI tools has become an essential skill for modern data scientists and developers. By effectively leveraging tools such as ChatGPT, Claude, Gemini, and others, developers can significantly enhance their efficiency in big data management, machine learning, and deep learning. From setting research directions to optimizing algorithmic models, these tools offer robust support. As AI technology continues to evolve, we can expect these tools to become even more intelligent and easier to use.

\chapter{Choosing Computer Hardware for Programming and ML}

In this chapter, we will explain the importance of selecting the right computer hardware for various tasks, from general programming to more demanding machine learning and deep learning workloads. The hardware you choose can significantly affect both your productivity and the performance of your applications.

\section{Example Configurations}

The x86 platform primarily utilizes CPUs from Intel and AMD. However, in recent years, Intel's introduction of big.LITTLE architecture and various blue screen issues have raised concerns. While Intel offers specific optimizations for gaming, it doesn't provide significant advantages for programming tasks. Therefore, we recommend AMD-based configurations, which offer better value for money and are generally more stable.

\textbf{About Overclocking:} Although overclocking is quite popular among gaming enthusiasts, it is not recommended for systems used in programming or deep learning tasks. These workloads prioritize stability, and deep learning in particular demands much longer processing times and heavier workloads compared to gaming. This means the computer’s endurance is more critical. For beginners, it is important to remember that maintaining system stability is key, especially when handling intensive computational tasks.

\subsection{A High-End DDR4 Configuration}

\subsubsection{Configurations and Analysis}

1. \textbf{Processor (CPU): AMD Ryzen 9 5950X}  
\begin{itemize}[noitemsep,topsep=0pt]
   \item 16 cores, 32 threads  
   \item Base frequency: 3.4 GHz, Max boost frequency: 4.9 GHz  
   \item TDP: 105W  
   \item Architecture: Zen 3  
\end{itemize}

\textbf{Why it suits deep learning}: A powerful multi-core processor is essential. The AMD Ryzen 9 5950X, with its 16 cores and 32 threads, efficiently handles tasks like data preprocessing, model compilation, and task distribution, which are CPU-intensive. 32 threads allows parallel processing, particularly in multi-task and high-load environments, such as training multiple models or handling different datasets simultaneously.

\noindent 2. \textbf{Graphics Card (GPU): NVIDIA GeForce RTX 3090}  
\begin{itemize}[noitemsep,topsep=0pt]
   \item Memory: 24 GB GDDR6X  
   \item CUDA Cores: 10,496  
   \item Tensor Cores: 328  
   \item Memory Bandwidth: 936.2 GB/s  
   \item Supports 8K resolution, ray tracing (RTX)  
\end{itemize}

\textbf{Why it suits deep learning:}  
The NVIDIA RTX 3090 is an excellent choice for deep learning:
\begin{itemize}[noitemsep,topsep=0pt]
   \item \textbf{Large VRAM}: 24 GB of VRAM allows for loading and processing large neural networks and datasets without frequent memory swapping, which is crucial for deep learning tasks such as image processing and NLP.
   \item \textbf{CUDA and Tensor Cores}: The 10,496 CUDA cores and 328 Tensor cores significantly enhance the parallel computing capability, accelerating matrix operations (such as matrix multiplication in deep learning) and convolution operations. Tensor cores further accelerate FP16/FP32 calculations, improving performance during training and inference.
   \item \textbf{High Memory Bandwidth}: The 936.2 GB/s memory bandwidth ensures rapid data transfer between memory and processing units, reducing latency and enhancing training efficiency.
\end{itemize}

\noindent3. \textbf{Memory (RAM): 32 GB x 4 DDR4 3200 MHz (128 GB total)}  
   \begin{itemize}[noitemsep,topsep=0pt]
       \item Type: DDR4  
       \item Frequency: 3200 MHz  
       \item CAS latency: CL16 or lower  
   \end{itemize}

\textbf{Why it suits deep learning:}  
Deep learning requires handling large datasets, especially in tasks involving computer vision and natural language processing. The 128 GB memory capacity allows loading large datasets into memory at once, reducing the need for frequent data swapping between memory and storage. This large memory capacity speeds up data preprocessing, batch loading, and in-memory computations, minimizing memory bottlenecks. Additionally, the 3200 MHz frequency ensures fast data transfer, improving system responsiveness.

\noindent4. \textbf{Motherboard: B550/x570 Chipset}  
   \begin{itemize}[noitemsep,topsep=0pt]
       \item Supports PCIe 4.0 (GPU slot)  
       \item Supports overclocking  
       \item Supports high-speed M.2 SSDs  
   \end{itemize}

\textbf{Why it suits deep learning:}  
The B550 chipset provides PCIe 4.0 support, which is critical for maximizing the performance of the RTX 3090 and NVMe SSD. While more affordable than the X570 chipset, the B550 still offers essential high-end features, including support for overclocking, high-speed memory, and fast storage, making it a cost-effective choice for deep learning tasks without sacrificing critical functionality.

\noindent5. \textbf{Storage (SSD): 2TB NVMe SSD}  
   \begin{itemize}[noitemsep,topsep=0pt]
       \item Read speed: $\sim$ 500 MB/s  
       \item Write speed: $\sim$000 MB/s  
       \item Interface: PCIe 4.0 (dependent on motherboard)  
   \end{itemize}

\textbf{Why it suits deep learning:}  
Deep learning involves frequent access to large datasets, and a high-speed NVMe SSD significantly accelerates data loading and storage operations. The speed of an NVMe SSD is crucial for tasks such as loading datasets, writing model weights, and saving checkpoints during training. The 2TB capacity is sufficient to store multiple projects, datasets, and model files, reducing the need to rely on external storage.

\noindent6. \textbf{Storage (HDD): 16TB SATA HDD}  
   \begin{itemize}[noitemsep,topsep=0pt]
       \item 7200 RPM, 256 MB cache  
       \item Large capacity for data storage  
   \end{itemize}

\textbf{Why it suits deep learning:}  
The 16TB HDD is ideal for storing large amounts of archival data, backups, and massive datasets that do not require frequent access. Although the read/write speed is slower compared to SSDs, HDDs offer a much larger capacity, which is essential for long-term storage in deep learning projects. This combination of SSD for performance and HDD for capacity provides an efficient solution for handling both hot and cold data.

\noindent7. \textbf{Power Supply: 1000W 80 PLUS Gold or Platinum Certified PSU}  
   \begin{itemize}[noitemsep,topsep=0pt]
       \item Provides sufficient power for high-end GPU and multi-core CPU  
       \item Power headroom for overclocking  
   \end{itemize}

\textbf{Why it suits deep learning:}  
The 1000W PSU ensures stable power delivery to high-performance components such as the RTX 3090 and Ryzen 9 5950X, which are power-hungry during deep learning workloads. The 80 PLUS Gold or Platinum certification guarantees high energy efficiency, which is crucial for maintaining system stability and reliability during long-duration, high-load tasks. Additionally, the extra power headroom allows for potential overclocking, ensuring future-proofing for more demanding workloads.

\noindent8. \textbf{Case: Full Tower Case}  
   \begin{itemize}[noitemsep,topsep=0pt]
       \item Excellent thermal design and airflow management  
       \item Supports multiple fans and liquid cooling solutions  
   \end{itemize}

   \textbf{Why it suits deep learning:}  
   Deep learning tasks often involve long periods of heavy GPU and CPU usage, generating significant heat. A full tower case provides ample space for effective cooling solutions, such as multiple fans or liquid cooling systems, ensuring optimal temperature control. Keeping components cool under heavy load helps maintain system performance and longevity, preventing thermal throttling and potential hardware damage.

\subsection{A High-End DDR4 ThreadRipper Configuration}

In this section, we will guide you through purchasing a high-end DDR4 ThreadRipper configuration on eBay, specifically focusing on the 3995WX, 5975WX, and 5995WX processors. Currently, DDR4-based CPUs are highly affordable, yet still deliver excellent performance, making them a cost-effective option. DDR4 memory and components are also readily available, making it a suitable choice for high-end users.

\subsubsection{Choosing the Processor}

ThreadRipper processors are ideal for high-end workstations or tasks requiring extreme multi-threading. Here are the three processors you can find on eBay, along with their price trends:

\begin{itemize}
    \item \textbf{3995WX}: 64 cores and 128 threads, suitable for massively parallel computing or virtualization environments. Originally priced at around \$5,500, it can now be found on eBay for around \$1,000 to \$1,500, offering excellent value.
    \item \textbf{5975WX}: 32 cores and 64 threads, ideal for applications requiring high parallel processing, such as video rendering and 3D modeling. Originally priced around \$3,200, current eBay prices are approximately \$1,200.
    \item \textbf{5995WX}: 64 cores and 128 threads, the most powerful in the ThreadRipper series, designed for extreme workloads. Originally priced at around \$6,500, it can now be found for about \$3,000 on eBay.
\end{itemize}

It's important to note that the 3995WX and other ThreadRipper 3000 series processors use the TRX40 chipset motherboards, while the ThreadRipper 5000 series (e.g., 5975WX and 5995WX) require the \textbf{WRX80} chipset motherboards. Be sure to choose the correct motherboard based on your processor.

When purchasing on eBay, be aware of vendor locks. For example, some 3995WX processors may have a \textbf{Lenovo lock}, meaning they will only work on specific Lenovo-branded motherboards. Avoid purchasing locked processors if you do not have a compatible motherboard.

\subsubsection{Motherboard Selection}

For the 3995WX, you should choose a motherboard that supports the TRX40 chipset, whereas the 5000 series ThreadRipper (like 5975WX and 5995WX) requires a WRX80 chipset motherboard. Here are some recommended options:

\begin{itemize}
    \item \textbf{TRX40 Motherboards} (for the 3995WX and other ThreadRipper 3000 series processors):
    \begin{itemize}
        \item \textbf{ASUS ROG Zenith II Extreme Alpha}
        \item \textbf{MSI Creator TRX40}
        \item \textbf{Gigabyte TRX40 AORUS XTREME}
    \end{itemize}
    \item \textbf{WRX80 Motherboards} (for the 5995WX, 5975WX, and other ThreadRipper 5000 series processors):
    \begin{itemize}
        \item \textbf{ASUS Pro WS WRX80E-SAGE SE WIFI}
        \item \textbf{Gigabyte WRX80 SU8}
    \end{itemize}
\end{itemize}

When searching for these models on eBay, ensure the product is free of vendor locks and comes with all necessary accessories (e.g., I/O shields, M.2 screws, etc.).

\subsubsection{Cooling Solution}

ThreadRipper processors generate significant heat, and \textbf{do not come with a stock cooler}, so you will need to choose a suitable cooling solution. The type of cooler you select will significantly affect the system's noise level. Below are common cooling options:

\begin{itemize}
    \item \textbf{Liquid Coolers (e.g., Corsair H150i RGB Pro XT)}: Liquid coolers can deliver efficient cooling while maintaining low noise levels, making them ideal for users requiring a quiet work environment. Liquid cooling systems typically produce minimal noise, perfect for home or office use.
    \item \textbf{High-Efficiency Air Coolers (e.g., Noctua NH-U14S TR4-SP3)}: Air coolers also provide good cooling but may generate more noise at higher fan speeds. Noctua, for example, offers high-end air coolers known for their low noise levels, suitable for users sensitive to noise.
\end{itemize}

\textbf{If using large, high-power fans (often referred to as "aggressive fans")}, they can offer extreme cooling efficiency but typically produce a significant amount of noise. This setup is often used for server-grade equipment or specialized high-performance workstations. If you plan to use such a cooling solution, we recommend installing your system in a \textbf{soundproof server room} or a dedicated rack to mitigate noise interference in your work environment.

When installing the cooler, ensure it comes with the \textbf{TR4 or SP3 mounting brackets}, as traditional coolers might not be compatible with the larger processor size.

\subsubsection{Memory Selection}

The ThreadRipper platform supports multi-channel memory, so we recommend starting with \textbf{32GB per DIMM} and using at least 128GB (4 x 32GB) or more to fully utilize the multi-channel capability. Here are some recommended memory options:

\begin{itemize}
    \item \textbf{Corsair Vengeance LPX DDR4 3200MHz 32GB}: A cost-effective option with good stability.
    \item \textbf{G.Skill Ripjaws V DDR4 3200MHz 32GB}: High-frequency memory, ideal for multi-tasking and rendering workloads.
\end{itemize}

You can choose a 4-channel or 8-channel memory configuration to maximize memory bandwidth and performance, depending on the motherboard's maximum supported memory channels.

\subsection{A High-End DDR4 Server Configuration}

Next, we will discuss configuring a high-end DDR4 server based on AMD EPYC 7763 and 7773X processors. With DDR4 memory prices dropping and the price of these processors significantly lower on the second-hand market, this makes for a very cost-effective solution.

\subsubsection{Choosing the Processor}

AMD's EPYC series processors are designed for server and data center environments. Here are two recommended processors, along with their price trends:

\begin{itemize}
    \item \textbf{EPYC 7763}: 64 cores and 128 threads, delivering excellent multi-threading performance, ideal for highly concurrent workloads. Originally priced at around \$7,800, it can now be found for about \$700 on eBay.
    \item \textbf{EPYC 7773X}: 64 cores and 128 threads, leveraging 3D V-Cache technology to provide larger cache sizes, making it ideal for workloads requiring huge data caches, such as database processing. Originally priced at \$8,500, it can now be found for around \$1,500 on eBay.
\end{itemize}

Be aware that these processors may have vendor locks (e.g., HP, Dell, or Lenovo lock), meaning they will only work on specific branded motherboards. Make sure the processor is compatible with your motherboard before purchasing.

\subsubsection{Motherboard Selection}

EPYC processors require motherboards with the SP3 socket. Here are some common motherboard options:

\begin{itemize}
    \item \textbf{Supermicro H12SSL-i}
    \item \textbf{Gigabyte MZ72-HB0}
    \item \textbf{ASUS KGPE-D16}
\end{itemize}

When searching for these motherboards on eBay, ensure that the product includes all necessary accessories, particularly the I/O shield and cooling components.

\subsubsection{Cooling Solution}

EPYC processors also do not come with stock coolers, so you will need to purchase a cooler that supports the SP3 socket. Below are recommended coolers:

\begin{itemize}
    \item \textbf{Noctua NH-U14S TR4-SP3}: An air cooler offering excellent cooling performance for the SP3 socket.
    \item \textbf{Dynatron A39 Server Cooler}: A small, efficient cooler suitable for server setups.
\end{itemize}

Ensure that the cooler comes with the correct SP3 mounting hardware to guarantee proper installation.

\subsubsection{Memory Selection}

EPYC processors support DDR4 RDIMM or LRDIMM. We recommend starting with \textbf{32GB per DIMM} and using an 8-channel configuration. The minimum recommended memory configuration is 256GB (8 x 32GB). Below are some memory options:

\begin{itemize}
    \item \textbf{Samsung DDR4 3200MHz RDIMM}
    \item \textbf{Micron DDR4 3200MHz LRDIMM}
\end{itemize}

Make sure that the memory type is compatible with your motherboard and processor, and we recommend using multi-channel memory configurations to maximize overall performance.

\subsubsection{Other Considerations}

For a server configuration, especially one based on the EPYC 7763 or 7773X processors, you should also consider the following hardware:

\begin{itemize}
    \item \textbf{Power Supply}: A high-wattage power supply is necessary for server systems. We recommend at least 1200W with 80 Plus Platinum certification.
    \item \textbf{Chassis}: Choose a chassis that supports multi-point cooling and large-sized motherboards. You can consider rackmount chassis from brands like Supermicro or Dell.
    \item \textbf{Cooling Systems}: If using high-power fans, we recommend installing the server in a dedicated soundproof server room or cabinet to avoid noise disruptions in your work environment.
\end{itemize}

By following these steps, you can easily complete your high-end ThreadRipper or EPYC server hardware purchases on eBay.

\subsection{A High-End DDR5 Configuration}

Another high-end configuration has AMD's Ryzen 9 7950X3D, DDR5 memory, and NVIDIA's RTX 4090. It is designed to provide exceptional performance for demanding tasks like deep learning, 3D rendering, gaming, and content creation. Below is the detailed configuration and an analysis of why each component is suited for these high-performance tasks.

\subsection{Configuration and Analysis}

\noindent1. \textbf{Processor (CPU): AMD Ryzen 9 7950X3D}  
   \begin{itemize}[noitemsep,topsep=0pt]
       \item 16 cores, 32 threads  
       \item Base frequency: 4.2 GHz, Max boost frequency: 5.7 GHz  
       \item TDP: 120W  
       \item 3D V-Cache technology for enhanced gaming and compute performance  
       \item Architecture: Zen 4  
   \end{itemize}

\textbf{Why it suits high-end performance:}  
The AMD Ryzen 9 7950X3D combines high core count and clock speed with AMD’s 3D V-Cache technology, which significantly enhances cache memory. This boosts performance in memory-bound tasks like gaming, rendering, and certain deep-learning models. The processor’s 16 cores and 32 threads make it highly capable of parallel processing, making it ideal for handling multiple demanding tasks simultaneously. Its high single-core performance also benefits tasks like gaming and workloads that depend on fast clock speeds.

\textbf{Or,}

\noindent1. \textbf{Processor (CPU): AMD Ryzen 9 9950X}  
   \begin{itemize}[noitemsep,topsep=0pt]
       \item 16 cores, 32 threads  
       \item Base frequency: 4.4 GHz, Max boost frequency: 5.9 GHz  
       \item TDP: 170W  
       \item Supports PCIe 5.0 and DDR5 memory  
       \item Architecture: Zen 5  
   \end{itemize}

\textbf{Why it suits high-end performance:}  
The AMD Ryzen 9 9950X combines high core count and clock speed with support for the latest PCIe 5.0 and DDR5 memory technologies. This delivers superior performance in memory-intensive tasks and faster data processing. With 16 cores and 32 threads, it excels at parallel processing, making it perfect for demanding tasks like rendering and AI workloads. Its AI optimizations and advanced architecture make it ideal for gaming and productivity, benefiting both single-core and multi-core performance.

\noindent2. \textbf{Graphics Card (GPU): NVIDIA GeForce RTX 4090}  
   \begin{itemize}[noitemsep,topsep=0pt]
       \item Memory: 24 GB GDDR6X  
       \item CUDA Cores: 16,384  
       \item Tensor Cores: 512  
       \item Memory Bandwidth: 1,008 GB/s  
       \item Supports 8K resolution, ray tracing (RTX), and DLSS 3.0  
   \end{itemize}

\textbf{Why it suits high-end performance:}  
The NVIDIA RTX 4090 is the most powerful GPU available in 2022, offering unparalleled performance for tasks such as deep learning, 3D rendering, and high-end gaming. The 24 GB GDDR6X memory allows the handling of large datasets and complex models without bottlenecks. With 16,384 CUDA cores and 512 Tensor cores, the RTX 4090 excels in parallel computing tasks, especially in deep learning where massive matrix operations and tensor calculations are needed. The advanced ray tracing and DLSS 3.0 technologies improve graphical fidelity and performance in games and visual simulations. The RTX 4090 is ideal for anyone requiring top-tier GPU performance for AI, creative, or gaming workloads.

\noindent3. \textbf{Memory (RAM): 96 GB DDR5}  
   \begin{itemize}[noitemsep,topsep=0pt]
       \item Type: DDR5  
       \item Frequency: 5200 MHz or slightly higher(up to 6000 MHz) 
       \item Configuration: 2 x 48 GB modules  
   \end{itemize}

\textbf{Why it suits high-end performance:}  
Although DDR5 offers higher bandwidth and improved efficiency over DDR4, it operates at slightly lower memory frequencies. The frequency might not reach its full potential due to DDR5's initial adoption phase, but the increased memory capacity—96 GB—more than compensates for this, allowing users to work with large datasets and multitask efficiently. This configuration is ideal for deep learning, large-scale simulations, and video editing, where vast amounts of data must be processed simultaneously. The 96 GB capacity is future-proof, offering ample headroom for memory-intensive applications.

\noindent4. \textbf{Motherboard: B650/X670 Chipset}  
   \begin{itemize}[noitemsep,topsep=0pt]
       \item Supports PCIe 5.0 (GPU slot)  
       \item Supports overclocking  
       \item Supports high-speed M.2 SSDs  
   \end{itemize}

\textbf{Why it suits deep learning:}  
The B650 and X670 chipsets provide PCIe 5.0 support, which is essential for maximizing the performance of next-generation GPUs and NVMe SSDs used in deep learning. These motherboards offer enhanced memory speed, overclocking capabilities, and support for high-bandwidth components, ensuring optimal performance for large datasets and compute-intensive tasks. The B650 offers a more affordable option with essential features, while the X670 provides additional PCIe lanes and connectivity options, making it a more robust choice for demanding deep learning environments.

\noindent5. \textbf{Storage (SSD): 4TB NVMe SSD}  
   \begin{itemize}[noitemsep,topsep=0pt]
       \item Read speed: $\sim$ 7000 MB/s  
       \item Write speed: $\sim$ 6800 MB/s  
       \item Interface: PCIe 4.0  
   \end{itemize}

\textbf{Why it suits high-end performance:}  
The 4TB NVMe SSD offers fast read and write speeds, significantly enhancing system responsiveness and reducing load times for large files and applications. This is especially important in deep learning and video editing tasks where large datasets and files need to be accessed and processed frequently. PCIe 4.0 support allows even faster data transfer rates, further reducing bottlenecks in data-heavy workflows.

\noindent6. \textbf{Storage (HDD): 20TB SATA HDD}  
   \begin{itemize}[noitemsep,topsep=0pt]
       \item 7200 RPM, 256 MB cache  
       \item Massive storage capacity  
   \end{itemize}

\textbf{Why it suits high-end performance:}  
The 20TB HDD provides an enormous amount of storage for archival data, backups, and large datasets that do not require frequent access. While not as fast as SSDs, the large capacity of the HDD is essential for long-term data storage in applications like video editing, deep learning, and content creation, where raw data, project files, and backup images can consume significant space.

\noindent7. \textbf{Power Supply: 1200W 80 PLUS Platinum Certified PSU}  
   \begin{itemize}[noitemsep,topsep=0pt]
       \item Provides ample power for high-end GPU and CPU  
       \item Efficiency ensures stable performance under heavy loads  
   \end{itemize}

\textbf{Why it suits high-end performance:}  
A 1200W power supply ensures sufficient and stable power delivery for the power-hungry RTX 4090 and Ryzen 9 7950X3D, especially during overclocking or peak loads. The 80 PLUS Platinum certification guarantees high efficiency, reducing power waste and ensuring reliability during prolonged high-load operations, which is critical for tasks like deep learning and rendering that require long periods of uninterrupted performance.

\noindent8. \textbf{Case: Full Tower Case}  
   \begin{itemize}[noitemsep,topsep=0pt]
       \item Supports excellent airflow and multiple cooling solutions  
       \item Ample space for large components  
   \end{itemize}

\textbf{Why it suits high-end performance:}  
A full tower case is essential for housing large components like the RTX 4090, and it provides space for advanced cooling systems such as liquid cooling. Given the high power consumption and heat generation of this configuration, efficient cooling is crucial to prevent thermal throttling and ensure long-term stability during extended workloads.

\section{Hardware Considerations for Programming}
When working on general programming tasks, the hardware requirements are relatively modest compared to machine learning or deep learning workloads. However, ensuring that you have appropriate resources is still crucial for efficient development. 

\subsection{Processor (CPU)}
For most programming tasks, a modern multi-core processor is sufficient. Compiling code and running applications will benefit from higher clock speeds and more cores, especially in multi-threaded programming environments. However, a balance should be struck between power and cost based on the complexity of your development needs.

\subsection{Memory (RAM)}
RAM is another critical factor for programming. While typical tasks may not be extremely memory-intensive, having at least 16 GB of RAM is recommended to avoid slowdowns due to memory paging, especially when running multiple applications or virtual machines.

\subsection{Storage}
Fast storage is essential for reducing load times when working with large files or compiling code. Solid-state drives (SSDs) are highly recommended over traditional hard disk drives (HDDs) to ensure faster data access and overall system responsiveness.

\section{Hardware for Machine Learning and Deep Learning}
Machine learning and deep learning tasks require considerably more powerful hardware due to the scale and complexity of the operations involved.

\subsection{Processor (CPU)}
While the CPU remains important for machine learning workflows, particularly for data preprocessing, it plays a secondary role compared to the GPU in deep learning tasks. A high-end CPU, such as AMD's Ryzen or Intel's i9 series, with multiple cores and high thread counts, is recommended to handle parallel data pipelines effectively.

\subsection{Graphics Processing Unit (GPU)}
The choice of GPU is critical for deep learning due to its ability to handle parallel computations efficiently. While NVIDIA GPUs with CUDA support remain widely used, there are now viable alternatives from Apple and AMD, offering diverse options for different users.

\subsubsection{NVIDIA GPUs}
NVIDIA continues to be a dominant player in deep learning, with their CUDA framework and GPUs such as the RTX series for consumer use and the A100 for enterprise-level tasks. These GPUs excel in handling large datasets and complex model training. Models like the RTX 4090, with 24 GB of memory, or enterprise-grade cards with even more VRAM, are ideal for heavy workloads.

\subsubsection{Apple’s Metal Performance Shaders (MPS)}
For users with modern Apple Silicon Macs, Apple’s Metal Performance Shaders (MPS) offer an integrated solution for GPU acceleration in deep learning. MPS support is built into frameworks such as PyTorch, enabling accelerated model training directly on Apple's M1, M2, and M3 chips. The unified memory architecture in Apple Silicon allows Apple GPUs to access the entire system's memory, making it an attractive option for local development without the need for external GPUs.

\subsubsection{AMD ROCm}
AMD's ROCm (Radeon Open Compute) platform is gaining traction as a strong competitor to CUDA, particularly with support for PyTorch, TensorFlow, and JAX. AMD’s recent GPUs, including the Radeon RX 7900 XTX and the MI250 series, are optimized for deep learning tasks. ROCm enables high-performance computing and offers alternatives for users looking for non-NVIDIA options, especially in Linux environments.

\subsubsection{Memory Considerations}
Regardless of the GPU brand, selecting a GPU with enough VRAM is critical. GPUs with 16 GB to 24 GB of memory, such as NVIDIA RTX cards and AMD’s Radeon Pro models, or Apple Silicon Macs with more than 32 GB of RAM are recommended to prevent memory bottlenecks during training.

Each of these platforms has its strengths, allowing users to choose based on their budget, hardware preferences, and compatibility with their development environments.

\subsection{Memory (RAM)}
RAM requirements for machine learning vary depending on the size of the dataset and the complexity of the models being trained. A minimum of 32 GB of RAM is generally recommended for most deep learning tasks, with 64 GB or more being preferable for larger datasets and more complex model architectures.

\subsection{Storage}
Deep learning tasks can involve large datasets that require fast access times. NVMe SSDs are ideal for this, providing high-speed data retrieval necessary for fast iteration and model training.

\subsection{Other Considerations}
For distributed training or working with very large datasets, systems with multiple GPUs or access to a dedicated server may be necessary. Additionally, cloud-based solutions such as AWS or Google Cloud can be explored as scalable options for handling massive workloads that exceed local hardware capabilities.

\section{Making Informed Decisions}
Choosing the right hardware for your specific needs depends on the type of tasks you expect to perform. While general programming can be accomplished on relatively modest systems, machine learning, and deep learning require significantly more powerful hardware, particularly in terms of GPUs and memory capacity. The goal of this chapter is to help you identify the right balance of components to ensure smooth development and efficient training workflows.

\chapter{Setting Up Your Development Environment}

Before embarking on any task, having the proper tools is crucial for success. In the realm of software development, setting up a proper development environment is the first critical step. With the right setup, developers can streamline their workflow, reduce errors, and focus on producing high-quality code. This chapter will guide you through the necessary steps to prepare your development environment, ensuring you are equipped to work efficiently with Python.

\section{Installing Python}

In this section, we will begin by setting up Python on your machine. Python is a versatile and widely used programming language, and installing it correctly is essential for any Python development. The following steps will guide you through downloading, installing, and verifying the Python installation on different operating systems.

\subsection{Downloading Python}

To download the latest version of Python, visit the official Python website at \url{https://www.python.org}. Navigate to the Downloads section, where you can choose the appropriate version for your operating system (Windows, macOS, or Linux). It is recommended to download the latest stable release for better compatibility with modern libraries and tools.

\subsection{Installing on Windows/Mac/Linux}

Once you have downloaded the Python installer, follow these instructions for installing Python on your system:

\begin{itemize}
    \item \textbf{Windows}: Run the installer and ensure the option "Add Python to PATH" is checked before continuing with the installation. This will allow you to run Python from the command line.
    \item \textbf{macOS}: macOS usually comes with Python pre-installed, but it may be an older version. To install the latest version, you can use a package manager like Homebrew (\texttt{brew install python}) or download it from the Python website.
    \item \textbf{Linux}: Most Linux distributions include Python by default, but if it's not present, you can install it manually using your package manager, such as \texttt{apt} for Ubuntu (\texttt{sudo apt install python3}) or \texttt{dnf} for Fedora.
\end{itemize}

\subsection{Verifying the Installation}

After installing Python, it is crucial to verify the installation. Open a terminal or command prompt and type the following command:

\begin{lstlisting}[style=cmd]
python --version
\end{lstlisting}

If the installation is successful, the terminal will display the installed Python version.

\begin{lstlisting}[style=text]
Python 3.11.3
\end{lstlisting}

Additionally, you can check whether the Python package installer, \texttt{pip}, is installed by typing:

\begin{lstlisting}[style=cmd]
pip --version
\end{lstlisting}

\textit{Note: If you are installing a newer version of Python 3.x, \texttt{pip} is usually installed automatically along with it.}

Once both versions are displayed, your Python installation is complete, and you are ready to begin coding!

\begin{lstlisting}[style=text]
pip 23.3.2 from C:\Users\user\AppData\Local\Programs\Python\Python311\Lib\site-packages\pip (python 3.11)
\end{lstlisting}

\section{Setting Up a Virtual Environment}

In this section, we will explore the significance of virtual environments in Python development and cover two popular ways to create and manage virtual environments: using \texttt{venv} and \texttt{conda}. Virtual environments allow you to isolate project dependencies and avoid conflicts between different versions of libraries.

\subsection{Importance of Virtual Environments}

A virtual environment is an isolated environment that contains its installation directories and libraries, separate from the global Python installation. This is essential when working on multiple projects that require different versions of Python packages. Virtual environments enable you to:

\begin{itemize}
    \item Avoid dependency conflicts between projects.
    \item Easily manage and track project-specific packages.
    \item Test code in different environments without affecting the system-wide installation.
\end{itemize}

\subsection{Creating a Virtual Environment with \texttt{conda}}

If you are using Anaconda or Miniconda, \texttt{conda} provides a powerful way to manage virtual environments. To create a virtual environment with \texttt{conda}, follow these steps:

\begin{enumerate}
    \item Open your terminal or Anaconda Prompt.
    \item Run the following command to create a new virtual environment with the desired Python version:
\begin{lstlisting}[style=cmd]
conda create --name myenv python=3.x
\end{lstlisting}
    \item To activate the environment, run:
\begin{lstlisting}[style=cmd]
conda activate myenv
\end{lstlisting}
    \item To deactivate the environment, run:
\begin{lstlisting}[style=cmd]
conda deactivate
\end{lstlisting}
\end{enumerate}

\subsection{Activating/Deactivating a Virtual Environment}

To activate or deactivate conda virtual environments:

\begin{itemize}
    \item \textbf{Activate}: \texttt{conda activate myenv}
    \item \textbf{Deactivate}: \texttt{conda deactivate}
\end{itemize}

\textit{Note: After activating the virtual environment, your terminal prompt will display the environment name in parentheses, e.g., \texttt{(myenv)}, indicating that the environment is active.}

\section{Installing Essential Packages}

In this section, we will discuss how to install Python packages using \texttt{pip}, introduce some commonly used packages, and provide tips on managing dependencies effectively. Packages are essential tools that help extend the functionality of Python and simplify various programming tasks.

\subsection{Using pip to Install Packages}

\texttt{pip} is the standard package manager for Python, used to install and manage libraries that are not part of the Python standard library. To install a package, use the following command:

\begin{lstlisting}[style=cmd]
pip install package_name
\end{lstlisting}

For example, to install \texttt{numpy} in a specific \texttt{conda} environment, follow these steps:

1. First, activate the environment where you want to install \texttt{numpy} by running:

\begin{lstlisting}[style=cmd]
conda activate myenv
\end{lstlisting}

2. Once the environment is activated, install \texttt{numpy} within that environment by running:

\begin{lstlisting}[style=cmd]
pip install numpy
\end{lstlisting}

This ensures that \texttt{numpy} is installed specifically in the activated environment.

To install multiple packages at once, use:

\begin{lstlisting}[style=cmd]
pip install numpy pandas matplotlib
\end{lstlisting}

To check if a package is installed or to see the installed version, use:

\begin{lstlisting}[style=cmd]
pip show package_name
\end{lstlisting}

To upgrade an already installed package to the latest version, use:

\begin{lstlisting}[style=cmd]
pip install --upgrade package_name
\end{lstlisting}

\subsection{Commonly Used Packages (e.g., numpy, pandas, matplotlib)}

There are many widely used Python packages, including:

\begin{itemize}
    \item \textbf{numpy}: A powerful library for numerical computations~\cite{Harris2020}.
    \item \textbf{pandas}: A data manipulation and analysis library~\cite{mckinney-proc-scipy-2010}.
    \item \textbf{matplotlib}: A plotting library for visualizations~\cite{Hunter2007Matplotlib}.
    \item \textbf{scikit-learn}: A machine learning library for data mining and analysis~\cite{scikit-learn}.
    \item \textbf{requests}: A simple HTTP library for making web requests.
    \item \textbf{tqdm}: A fast, extensible progress bar for loops and command-line applications.
\end{itemize}

\subsection{Installing Special Packages - PyTorch and CUDA}

\textit{Note: If you do not fully understand what CUDA is or do not need GPU acceleration at the moment, you can skip this section and simply install the CPU version of PyTorch using the following command:}

\begin{lstlisting}[style=cmd]
pip install torch torchvision torchaudio
\end{lstlisting}

\textbf{PyTorch} is a deep learning framework widely used in artificial intelligence and machine learning tasks~\cite{Paszke2019PyTorch}. PyTorch supports both CPU and GPU computations, leveraging \textbf{CUDA} for GPU acceleration. To utilize GPU for faster training and inference, it is essential to install the appropriate version of PyTorch that is compatible with CUDA. Below is an introduction to CUDA and the steps to install a CUDA-enabled PyTorch version.

\subsubsection{Introduction to CUDA}
\textbf{CUDA} (Compute Unified Device Architecture) is a parallel computing platform and API developed by NVIDIA that enables developers to use GPUs for high-performance computations. For tasks such as deep learning, where large-scale computations are required, GPU acceleration can significantly speed up model training and inference. To allow PyTorch to take advantage of CUDA, the correct version of the CUDA toolkit and a compatible version of PyTorch must be installed.

\subsubsection{Installing a CUDA-Enabled PyTorch}

Follow these steps to install PyTorch with CUDA support:

\begin{enumerate}
    \item Ensure that your system has the NVIDIA GPU driver and the appropriate version of the CUDA toolkit installed. You can check the CUDA version on your system using the command \texttt{nvidia-smi}.
    \item Visit the official PyTorch website (\url{https://pytorch.org/get-started/locally/}) and select the proper options for your operating system, a package manager (e.g., \texttt{pip} or \texttt{conda}), and the version of CUDA that matches your system configuration.
    \item For instance, if you want to install PyTorch with CUDA 11.8 on Windows using \texttt{pip}, you can follow the selection process and copy the generated command below into your Conda environment for execution.
    \begin{figure}[H]
        \centering
        \includegraphics[width=1.0\textwidth]{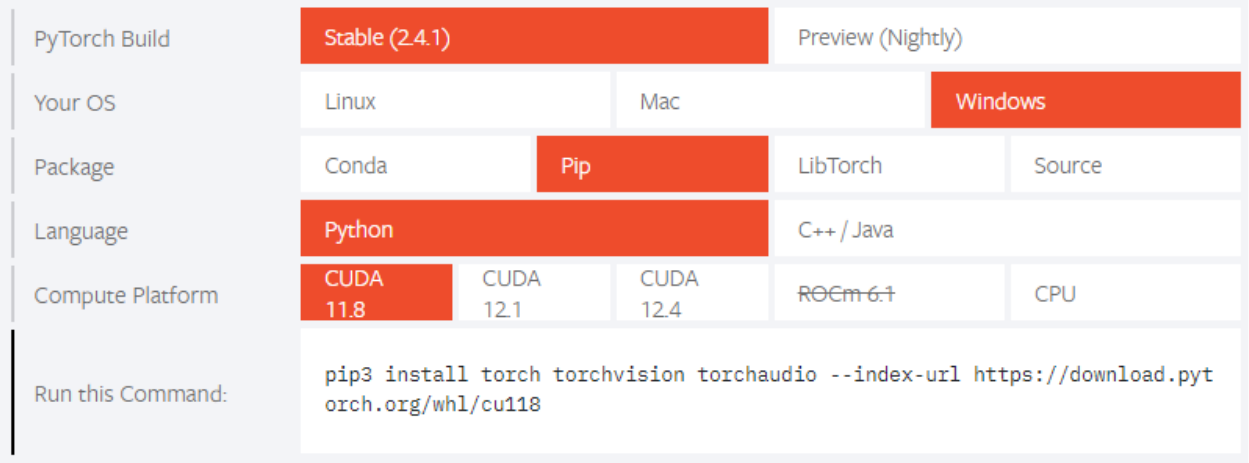}
        \caption{PyTorch installation guide for Windows using Pip with CUDA 11.8, showing the command to install PyTorch, torchvision, and torchaudio.}
    \end{figure}
    \item If the CUDA toolkit is not yet installed on your system, follow the instructions on the NVIDIA website to download and install the correct version.
\end{enumerate}

\subsubsection{Verifying PyTorch CUDA Support}

After installation, you can verify whether PyTorch detects CUDA support by running the following Python command in your conda environment:

\begin{lstlisting}[style=python]
import torch
print(torch.cuda.is_available())  # Returns True if CUDA is available
\end{lstlisting}

By installing PyTorch with CUDA support, you can take full advantage of the computational power of GPUs, dramatically improving the performance of deep learning models, especially when dealing with large datasets and complex architectures.

\subsection{Installing Special Packages - TensorFlow and CUDA}

\textit{Note: If you do not fully understand what CUDA is or do not need GPU acceleration at the moment, you can skip this section and simply install the CPU version of TensorFlow using the following command:}

\begin{lstlisting}[style=cmd]
pip install tensorflow
\end{lstlisting}

To use \textbf{TensorFlow}~\cite{abadi2016tensorflowlargescalemachinelearning} with GPU acceleration, you will need to install the correct version of TensorFlow that is compatible with your system’s CUDA version. TensorFlow leverages NVIDIA's CUDA framework for GPU computation, which can greatly accelerate the performance of deep learning models.

Here are the steps to install TensorFlow with CUDA support:

\begin{itemize}
    \item First, ensure that you have a compatible version of CUDA and cuDNN installed. TensorFlow supports specific versions of CUDA and cuDNN, so check TensorFlow’s compatibility chart to match the correct versions.
    
    \item Next, install the TensorFlow GPU version via pip:
    
    \begin{lstlisting}[style=cmd]
pip install tensorflow-gpu
    \end{lstlisting}
    
    \item After installation, verify that TensorFlow can access the GPU by running the following Python script:
    
    \begin{lstlisting}[style=python]
import tensorflow as tf
print("Num GPUs Available: ", len(tf.config.list_physical_devices('GPU')))
    \end{lstlisting}
    
    If the output shows the number of available GPUs, the installation was successful and TensorFlow is ready to use the GPU for computations.
    \begin{lstlisting}[style=cmd]
Num GPUs Available:  0
    \end{lstlisting}
    
\end{itemize}

Make sure you also have the necessary NVIDIA drivers installed on your system for CUDA to function properly. Additionally, it’s recommended to use a virtual environment to manage package dependencies effectively.

\subsection{Managing Dependencies}

Managing multiple packages and their versions is crucial for project stability. Some best practices include:

\begin{enumerate}
    \item \textbf{Using a \texttt{requirements.txt} file}: List the required packages and versions in this file.
    
\begin{lstlisting}[style=text]
// Basic version 
numpy
tqdm
opencv-python

// Specifying a minimum version
gradio>=4.21.0
matplotlib>=3.2.2
\end{lstlisting}

Once you've listed your required packages in the \texttt{requirements.txt} file, you can install all the packages at once with the following command:

\begin{lstlisting}[style=text]
pip install -r requirements.txt
\end{lstlisting}
    \item \textbf{Specifying version numbers}: Ensure correct package versions in the \texttt{requirements.txt} file to avoid issues from updates.
    \item \textbf{Using virtual environments}: Isolate your project’s dependencies from the global environment.
\end{enumerate}

\section{Setting Up an Integrated Development Environment (IDE)}

In this section, we will explore the differences between text editors and integrated development environments (IDEs), and discuss how to set up and configure a popular IDE for Python development. Having the right development environment is crucial for improving productivity, organizing code, and simplifying debugging processes.

\subsection{Introduction to Popular IDEs (e.g., PyCharm, VSCode)}

There are two primary types of tools for writing and managing code: \textbf{text editors} and \textbf{IDEs}. Both serve the purpose of editing code, but they differ in functionality and the level of integration.

\begin{itemize}
    \item \textbf{Text Editors}: Text editors are generally simpler and lighter, focusing on writing and editing text, with fewer built-in features. Some popular text editors include:
    \begin{itemize}
        \item \textbf{Vim}: A highly customizable, keyboard-driven text editor favored by developers for its efficiency.
        \item \textbf{Notepad++}: A free text editor for Windows that supports syntax highlighting and other basic features but lacks integrated debugging or project management.
        \item \textbf{VSCode (Visual Studio Code)}: A highly extensible text editor that, with the right extensions, can function similarly to an IDE. It supports syntax highlighting, debugging, version control, and more.
    \end{itemize}
    \item \textbf{IDEs}: Integrated Development Environments come with a full set of tools for coding, debugging, testing, and managing projects all in one place.
    \begin{itemize}
        \item \textbf{PyCharm}: A powerful IDE specifically designed for Python development. PyCharm provides tools like code completion, debugging, refactoring, and testing built into one package, making it ideal for larger, more complex projects.
    \end{itemize}
\end{itemize}

\subsection{Setting Up and Configuring Your IDE}

Once you've selected the right tool for your workflow, the next step is setting up your IDE or text editor. Below, we’ll outline how to configure \textbf{VSCode} and \textbf{PyCharm}.

\begin{itemize}
    \item \textbf{Setting Up VSCode}:
    \begin{enumerate}
        \item Download and install VSCode.
        \item Install the Python extension from the Extensions Marketplace.
        \item Set the correct Python interpreter via the command palette (\texttt{Ctrl+Shift+P}).
        \item Install additional extensions like Pylint and Black for linting and code formatting.
    \end{enumerate}
    \item \textbf{Setting Up PyCharm}:
    \begin{enumerate}
        \item Download and install PyCharm.
        \item Set up a new Python project and configure the Python interpreter.
        \item Customize settings for code style, Git integration, and the integrated terminal.
        \item Use PyCharm’s powerful debugging tools to step through your code and troubleshoot issues.
    \end{enumerate}
\end{itemize}

Proper configuration of your IDE or text editor can significantly improve your productivity and help you focus on writing better code.

\subsection{Recommended VSCode Extensions}

Here are some useful extensions for VSCode that can enhance your Python development workflow and help with specific tasks:

\begin{itemize}
    \item \textbf{JSON Crack}: A handy tool for visualizing JSON structures. This extension allows you to explore JSON data in a graphical format, making it easier to understand complex nested structures.
    
    \item \textbf{Rainbow CSV}: This extension highlights columns in CSV files, making it easier to recognize and differentiate columns when working with data. It's particularly helpful when dealing with large datasets or unformatted CSV files.
    
    \item \textbf{Copilot}: A powerful AI-driven extension developed by GitHub. Copilot assists by providing intelligent code suggestions, auto-completion, and even writing entire blocks of code based on your prompts, greatly improving coding speed and accuracy.
    
    \item \textbf{CodeSnap}: A simple but effective extension that allows you to take beautiful screenshots of your code snippets directly from VSCode. It includes customization options like background color and padding to enhance readability and aesthetics.
\end{itemize}

\chapter{Introduction to Python Programming}

In this chapter, we will start by writing a simple Python program to print a message, ``Hello, World!'' This is the first step in learning any programming language and will help you understand the basics of Python syntax.

\section{Hello World}

In this section, we will begin with the most basic Python program by printing a string to the screen.

\subsection{Hello World — Printing a String}

First, let’s look at a very simple program that prints ``Hello, World!'' to the screen.

\begin{lstlisting}[style=python]
print("Hello, World!")
\end{lstlisting}

When you run this code, the output on your screen will be:

\begin{lstlisting}[style=text]
Hello, World!
\end{lstlisting}

This program is very simple. It uses the ``print'' function to display the text inside the parentheses on the screen. This is your first Python program.

\subsection{Hello World — Defining a Function}

Next, we can place the above code inside a function. A function is like a small tool that you can use whenever you need to perform a specific task.

\begin{lstlisting}[style=python]
def hello_world():
    print("Hello, World!")
\end{lstlisting}

Now, if we want to display “Hello, World!” we just need to call this function:

\begin{lstlisting}[style=python]
hello_world()
\end{lstlisting}

When you run this code, the output on your screen will be:

\begin{lstlisting}[style=text]
Hello, World!
\end{lstlisting}

Here is the complete code:

\begin{lstlisting}[style=python]
def hello_world():
    print("Hello, World!")

hello_world()
\end{lstlisting}

\subsection{Hello World — Using \texttt{if \_\_name\_\_ == "\_\_main\_\_"}}

Next, let’s discuss a slightly more complex concept. Suppose you have two Python files: one called \texttt{hw.py} and another called \texttt{test.py}. Let’s first look at the \texttt{hw.py} file:

\begin{lstlisting}[style=python]
def hello_world():
    print("Hello, World!")

if __name__ == "__main__":
    hello_world()
\end{lstlisting}

Here’s what happens:

When you run the \texttt{hw.py} file directly, Python will execute the \texttt{hello\_world()} function, and the output on your screen will be:

\begin{lstlisting}[style=text]
Hello, World!
\end{lstlisting}

Now, suppose you want to use the \texttt{hello\_world} function from \texttt{hw.py} in another file, \texttt{test.py}. You can do this as follows:

\begin{lstlisting}[style=python]
import hw

hw.hello_world()
\end{lstlisting}

When you run \texttt{test.py}, the output on your screen will be:

\begin{lstlisting}[style=text]
Hello, World!
\end{lstlisting}

However, if you do not want the code in \texttt{hw.py} to run automatically when it is imported into another file, but only when it is run directly, then \texttt{if \_\_name\_\_ == "\_\_main\_\_"} is very useful. This ensures that \texttt{hello\_world()} is only executed when you run \texttt{hw.py} directly; if you import it into \texttt{main.py}, the code will not run unless you explicitly call it in \texttt{test.py}.

This is the purpose of \texttt{if \_\_name\_\_ == "\_\_main\_\_"} — it prevents code from running when it is not supposed to, giving you better control over the behavior of your program.

In debugging a program that consists of multiple source files, using the \texttt{if \_\_name\_\_ == "\_\_main\_\_"} statement can help you debug more effectively. Specifically, when your project includes multiple modules or script files, each file may have its functionality, logic, and interfaces. When you want to debug a specific module independently, you can add \texttt{if \_\_name\_\_ == "\_\_main\_\_"} to that module and write some modular test code or invoke functions within that module underneath it. This not only helps in verifying the module's independence but also ensures that each part can function correctly on its own when integrated with other modules, thereby improving debugging efficiency and program reliability.

On Windows, if you want to use multi-threading, you must also wrap the main program in the \texttt{if \_\_name\_\_ == "\_\_main\_\_"} construct. This is because, on Windows, when using multi-threading or multi-processing modules (such as \texttt{threading} or \texttt{multiprocessing}), child processes will reload the main program. Without this safeguard, the main program might be executed multiple times, leading to the creation of new processes in a loop, which could cause infinite recursion or other unexpected behavior. Therefore, when writing multi-threaded programs, especially on Windows, it's crucial to place the main program inside the \texttt{if \_\_name\_\_ == "\_\_main\_\_"} block to avoid such issues.

\section{Data Types in Python}
Python supports several built-in data types, which can be broadly categorized into numbers, strings, lists, tuples, dictionaries, and sets. Understanding these types is essential for managing data effectively in Python.

\subsection{Numbers}
Numbers in Python include integers, floats (decimal numbers), and complex numbers. Python supports standard arithmetic operations on numbers such as addition, subtraction, multiplication, and division.

\begin{lstlisting}[style=python]
x = 10       # integer
y = 3.14     # float
z = 1 + 2j   # complex number
\end{lstlisting}

Python automatically performs type conversion (implicit casting) when you perform operations between different types of numbers. For example, if you add an integer and a float, Python will automatically convert the integer to a float, ensuring that the result is also a float. Similarly, operations between floats and complex numbers will result in a complex number.

\subsubsection{Example 1: Integer and Float Addition}
When you add an integer and a float, the result is a float because Python promotes the integer to a float to avoid losing precision.

\begin{lstlisting}[style=python]
a = 10        # integer
b = 2.5       # float
result = a + b
print(result) 
print(type(result)) 
\end{lstlisting}

Output:
\begin{lstlisting}[style=cmd]
12.5
<class 'float'>
\end{lstlisting}

\subsubsection{Example 2: Float and Complex Number Addition}
When adding a float and a complex number, Python automatically converts the float to a complex number, preserving the real and imaginary parts.

\begin{lstlisting}[style=python]
c = 3.14      # float
d = 2 + 3j    # complex number
result = c + d
print(result) 
print(type(result))  
\end{lstlisting}

Output:
\begin{lstlisting}[style=cmd]
(5.14+3j)
<class 'complex'>
\end{lstlisting}

\subsubsection{Example 3: Division of Integers Results in a Float}
Even when dividing two integers, the result will always be a float in Python 3.x. This ensures that division always returns a precise result, avoiding truncation.

\begin{lstlisting}[style=python]
e = 10
f = 3
result = e / f
print(result)
print(type(result)) 
\end{lstlisting}

Output:
\begin{lstlisting}[style=cmd]
3.3333333333333335
<class 'float'>
\end{lstlisting}

\subsubsection{Example 4: Integer Division Using //}
If you want to perform integer division (where the result is an integer and truncates the decimal part), you can use the `//` operator.

\begin{lstlisting}[style=python]
g = 10
h = 3
result = g // h
print(result)
print(type(result)) 
\end{lstlisting}

Output:
\begin{lstlisting}[style=cmd]
3
<class 'int'>
\end{lstlisting}

\subsubsection{Example 5: Modulus and Power Operations}
Python also supports modulus and power operations on numbers:

\begin{lstlisting}[style=python]
i = 10
j = 3

# Modulus: Remainder of division
modulus_result = i % j
print(modulus_result) 

# Power: Raising a number to a power
power_result = i ** j
print(power_result)  
\end{lstlisting}

Output:
\begin{lstlisting}[style=cmd]
1
1000
\end{lstlisting}

\subsection{Strings}
Strings are sequences of characters. They can be created by enclosing text in either single or double quotes.

\begin{lstlisting}[style=python]
greeting = "Hello, Python!"
name = 'Alice'
\end{lstlisting}

\subsubsection{Common String Methods}
Python provides several built-in methods to manipulate strings. For example:

\begin{lstlisting}[style=python]
text = "hello world"
print(text.upper())  
print(text.replace("world", "Python"))
\end{lstlisting}

Output:
\begin{lstlisting}[style=cmd]
HELLO WORLD
hello Python
\end{lstlisting}

\subsubsection{String Formatting}
You can format strings using various methods, including f-strings, the format method, or using the old \% operator.

\begin{lstlisting}[style=python]
name = "Alice"
age = 25
print(f"My name is {name} and I am {age} years old.")  # f-string
print("My name is {} and I am {} years old.".format(name, age))  # .format()
print("My name is %s and I am %d years old." % (name, age))  # % operator
print("My name is " + str(name) + "and I am " + str(age) + "years old.")  # string concatenation
\end{lstlisting}

Output:
\begin{lstlisting}[style=cmd]
My name is Alice and I am 25 years old.
My name is Alice and I am 25 years old.
My name is Alice and I am 25 years old.
My name is Alice and I am 25 years old.
\end{lstlisting}

\section{Lists}
Lists are ordered collections of items in Python. They are mutable, which means you can change their content without creating a new list. Lists can hold items of different data types, including other lists.

\begin{lstlisting}[style=python]
# Creating a list
fruits = ["apple", "banana", "cherry"]
print(fruits[0])  

# Accessing list elements
print(fruits[1])  
print(fruits[-1])

# Slicing lists
print(fruits[1:3])  

# Unpack lists
print(*fruits)
\end{lstlisting}

Output:
\begin{lstlisting}[style=cmd]
apple
banana
cherry
['banana', 'cherry']
banana cherry
\end{lstlisting}

\subsection{Common List Methods}
Python lists come with several useful methods for manipulating their content:

\begin{lstlisting}[style=python]
# Creating a list
fruits = ["apple", "banana", "cherry"]
print("Original list:", fruits)

# Adding an item to the list
fruits.append("orange")  # Adds 'orange' to the end of the list
print("After append:", fruits)

# Inserting an item at a specific position
fruits.insert(1, "blueberry")  # Inserts 'blueberry' at index 1
print("After insert:", fruits)

# Removing an item by value
fruits.remove("banana")  # Removes the first occurrence of 'banana'
print("After remove:", fruits)

# Removing an item by index
del fruits[0]  # Removes the item at index 0 ('apple')
print("After delete:", fruits)

# Sorting the list
fruits.sort()  # Sorts the list in alphabetical order
print("After sort:", fruits)

# Reversing the list
fruits.reverse()  # Reverses the order of the list
print("After reverse:", fruits)

# Copying the list
new_fruits = fruits.copy()  # Creates a shallow copy of the list
print("Copied list:", new_fruits)
\end{lstlisting}

Output:
\begin{lstlisting}[style=cmd]
Original list: ['apple', 'banana', 'cherry']
After append: ['apple', 'banana', 'cherry', 'orange']
After insert: ['apple', 'blueberry', 'banana', 'cherry', 'orange']
After remove: ['apple', 'blueberry', 'cherry', 'orange']
After delete: ['blueberry', 'cherry', 'orange']
After sort: ['blueberry', 'cherry', 'orange']
After reverse: ['orange', 'cherry', 'blueberry']
Copied list: ['orange', 'cherry', 'blueberry']
\end{lstlisting}

\subsection{Nested Lists}
Lists can contain other lists, which allows for the creation of more complex data structures.

\begin{lstlisting}[style=python]
# Creating a nested list
matrix = [
    [1, 2, 3],
    [4, 5, 6],
    [7, 8, 9]
]

# Accessing elements in a nested list
# matrix[row][column]
print(matrix[0][1])  # Output: 2
print(matrix[1])     # Output: [4, 5, 6]
\end{lstlisting}

Output:
\begin{lstlisting}[style=cmd]
2
[4, 5, 6]
\end{lstlisting}

\section{Tuples}
Tuples are immutable sequences in Python, meaning once you create a tuple, you cannot modify its elements. They are often used to group related data together and can be of mixed data types. Tuples are defined using parentheses `()`.

\begin{lstlisting}[style=python]
# Creating a tuple
coordinates = (10, 20)
print(coordinates[0])  # Prints the first element
print(coordinates[1])  # Prints the second element

# Tuples can contain different types of data
mixed_tuple = (1, "apple", 3.14, [1, 2, 3])
print(mixed_tuple)  # Prints the entire tuple

# Nested tuples
nested_tuple = (1, (2, 3), (4, 5))
print(nested_tuple[1])  # Prints the nested tuple at index 1
print(nested_tuple[1][0])  # Prints the first element of the nested tuple at index 1

# Tuple unpacking
a, b, c = (10, 20, 30)
print(a)  # Prints the value of a
print(b)  # Prints the value of b
print(c)  # Prints the value of c

# Concatenating tuples
tuple1 = (1, 2, 3)
tuple2 = (4, 5, 6)
combined = tuple1 + tuple2
print(combined)  # Prints the concatenated tuple

# Repeating tuples
repeated = tuple1 * 3
print(repeated)  # Prints the repeated tuple

# Checking membership
print(2 in tuple1)  # Checks if 2 is in tuple1
print(7 in tuple1)  # Checks if 7 is in tuple1
\end{lstlisting}

Output:
\begin{lstlisting}[style=cmd]
10
20
(1, 'apple', 3.14, [1, 2, 3])
(2, 3)
2
10
20
30
(1, 2, 3, 4, 5, 6)
(1, 2, 3, 1, 2, 3, 1, 2, 3)
True
False
\end{lstlisting}

\section{Dictionaries}
Dictionaries are collections of key-value pairs. They allow you to store data that is associated with a unique key. Dictionaries are defined using curly braces `{}`.

\begin{lstlisting}[style=python]
# Creating a dictionary
person = {"name": "Alice", "age": 25}
print(person["name"])  # Prints the value associated with the key 'name'
print(person["age"])   # Prints the value associated with the key 'age'

# Adding a new key-value pair
person["email"] = "alice@example.com"
print(person)  # Prints the updated dictionary

# Updating an existing value
person["age"] = 26
print(person)  # Prints the dictionary with the updated age

# Removing a key-value pair
del person["email"]
print(person)  # Prints the dictionary after removing the 'email' key

# Accessing keys and values
print(person.keys())  # Prints all keys in the dictionary
print(person.values())  # Prints all values in the dictionary

# Checking if a key exists
print("name" in person)  # Checks if 'name' is a key in the dictionary
print("email" in person)  # Checks if 'email' is a key in the dictionary

# Using the get method
print(person.get("name"))  # Safely retrieves the value for 'name'
print(person.get("phone", "Not Available"))  # Returns default value if 'phone' key does not exist
\end{lstlisting}

Output:
\begin{lstlisting}[style=cmd]
Alice
25
{'name': 'Alice', 'age': 25, 'email': 'alice@example.com'}
{'name': 'Alice', 'age': 26, 'email': 'alice@example.com'}
{'name': 'Alice', 'age': 26}
dict_keys(['name', 'age'])
dict_values(['Alice', 26])
True
False
Alice
Not Available
\end{lstlisting}

\section{Sets}
Sets are unordered collections of unique elements. They are useful when you want to store unique values and perform operations like union, intersection, and difference. Sets are defined using curly braces `{}`.

\begin{lstlisting}[style=python]
# Creating a set
my_set = {1, 2, 3, 3, 4}
print(my_set)  # Prints the set, automatically removing duplicates

# Adding elements to the set
my_set.add(5)
print(my_set)  # Prints the set after adding 5

# Removing elements from the set
my_set.remove(2)
print(my_set)  # Prints the set after removing 2

# Checking membership
print(3 in my_set)  # Checks if 3 is in the set
print(2 in my_set)  # Checks if 2 is in the set

# Set operations
set1 = {1, 2, 3}
set2 = {3, 4, 5}

# Union of sets
union_set = set1 | set2
print(union_set)  # Prints the union of set1 and set2

# Intersection of sets
intersection_set = set1 & set2
print(intersection_set)  # Prints the intersection of set1 and set2

# Difference of sets
difference_set = set1 - set2
print(difference_set)  # Prints the difference between set1 and set2

# Symmetric difference of sets
symmetric_difference_set = set1 ^ set2
print(symmetric_difference_set)  # Prints the symmetric difference between set1 and set2
\end{lstlisting}

Output:
\begin{lstlisting}[style=cmd]
{1, 2, 3, 4}
{1, 2, 3, 4, 5}
{1, 3, 4, 5}
True
False
{1, 2, 3, 4, 5}
{3}
{1, 2}
{1, 2, 4, 5}
\end{lstlisting}

\section{Variables in Python}
Variables in Python are used to store data, and their values can change throughout the program. Python does not require explicit variable declarations. Variables can be classified into global and local types.

\subsection{Global and Local Variables}
Variables defined outside of functions are global variables, while those defined inside functions are local variables. Global variables can be accessed from anywhere in the code, whereas local variables are only accessible within the function where they are defined.

\begin{lstlisting}[style=python]
x = "global"  # Global variable

def example():
    x = "local"  # Local variable
    print(x) 

example()
print(x)  
\end{lstlisting}

Output:
\begin{lstlisting}[style=cmd]
local
global
\end{lstlisting}

\subsection{Call by Object Reference}
In Python, when passing arguments to functions, you are passing references to the objects, not copies of the objects. This means that changes made to mutable objects inside a function will affect the original object, while immutable objects will not be altered.

\subsubsection{Mutable Objects}
Mutable objects can be changed after their creation. Lists, dictionaries, and sets are examples of mutable objects.

\begin{lstlisting}[style=python]
def modify_list(lst):
    lst.append(4)  # Modifies the original list

original_list = [1, 2, 3]
modify_list(original_list)
print(original_list)  
\end{lstlisting}

Output:
\begin{lstlisting}[style=cmd]
[1, 2, 3, 4]
\end{lstlisting}

\subsubsection{Immutable Objects}
Immutable objects cannot be changed after their creation. Integers, floats, strings, and tuples are examples of immutable objects.

\begin{lstlisting}[style=python]
def modify_tuple(t):
    t += (4,)  # Creates a new tuple, does not modify the original tuple

original_tuple = (1, 2, 3)
modify_tuple(original_tuple)
print(original_tuple) 
\end{lstlisting}

Output:
\begin{lstlisting}[style=cmd]
(1, 2, 3)
\end{lstlisting}

\section{Floating-Point Precision Issues in Python}

In Python, when working with floating-point numbers, you may encounter results that differ slightly from what you expect. This occurs because of the way floating-point numbers are stored in a computer, which can lead to precision issues. This problem is not unique to Python—many other programming languages also experience similar behavior.

\subsection{The Cause of Floating-Point Precision Issues}

Computers represent floating-point numbers using the IEEE 754 standard, which stores numbers as binary fractions. However, decimal numbers like $0.1$ and $0.2$ cannot be exactly represented in binary, which leads to small rounding errors during storage and computation. For example, in Python, when calculating $0.1 + 0.2$, the result is $0.30000000000000004$ instead of $0.3$.

\begin{lstlisting}[style=cmd]
>>> 0.1 + 0.2
0.30000000000000004
\end{lstlisting}

\subsection{Which Programming Languages Have Similar Issues?}

This floating-point precision problem occurs in many programming languages that also use the IEEE 754 standard, including but not limited to: JavaScript, Java, C, C++, Go, Ruby, and Swift.

\subsection{The Real Impact of Floating-Point Operations}

This precision issue can affect not only simple arithmetic operations but also common iteration operations. For example, when using code like \texttt{np.arange(0.1, 0.4, 0.1)}, you might expect the output to be $0.1$, $0.2$, and $0.3$, but due to precision errors, the output could include $0.30000000000000004$ and $0.4$.

\begin{lstlisting}[style=python]
import numpy as np

for x in np.arange(0.1, 0.4, 0.1):
    print(x)
\end{lstlisting}

Output:
\begin{lstlisting}[style=cmd]
0.1
0.2
0.30000000000000004
0.4
\end{lstlisting}

As seen in the example above, the accumulation of floating-point precision errors causes the result to deviate from what you might expect, such as $0.3$ being displayed as $0.30000000000000004$.

\subsection{How to Handle Floating-Point Precision in Python}

To avoid these floating-point precision issues, Python provides the \texttt{decimal} module, which allows for exact decimal arithmetic. This is especially useful in financial or scientific calculations. For example, by using \texttt{decimal.Decimal}, you can get exact results without floating-point errors.

\begin{lstlisting}[style=python]
from decimal import Decimal

a = Decimal('0.1')
b = Decimal('0.2')
print(a + b) 
\end{lstlisting}

Output:
\begin{lstlisting}[style=cmd]
0.3
\end{lstlisting}

\subsection{Banker's Rounding and Floating-Point Precision}

One common issue arising from floating-point precision is related to rounding methods, such as Banker's Rounding. This rounding method, also known as round-half-to-even, helps to minimize cumulative rounding errors. It rounds to the nearest value, with ties broken by rounding to the nearest even number.

Python’s built-in `round()` function provides an example of Banker's Rounding behavior. Consider rounding $2.5$ and $3.5$ using Python's `round()` function:

\begin{lstlisting}[style=python]
# Example of Banker's Rounding using Python's round() function

# Round 2.5 and 3.5 using Banker's rounding
print(round(2.5)) 
print(round(3.5))  
\end{lstlisting}

Output:
\begin{lstlisting}[style=cmd]
2
4
\end{lstlisting}

If you need traditional rounding where $0.5$ always rounds up to the next integer, you can use the `decimal` module with the `ROUND\_HALF\_UP` method:

\begin{lstlisting}[style=python]
import decimal

decimal.getcontext().rounding = decimal.ROUND_HALF_UP

# Round 2.5 and 3.5 using traditional rounding
print(decimal.Decimal('2.5').quantize(decimal.Decimal('1'))) 
print(decimal.Decimal('3.5').quantize(decimal.Decimal('1')))
\end{lstlisting}

Output:
\begin{lstlisting}[style=cmd]
3
4
\end{lstlisting}

\section{Basic Control Structures}

\subsection{Logical Operators: \texttt{and}, \texttt{or}, \texttt{not}}
Logical operators in Python allow you to combine or invert conditions in your conditional statements. The three main logical operators are \texttt{and}, \texttt{or}, and \texttt{not}.

\subsubsection{\texttt{and} Operator:}
The \texttt{and} operator is used to combine two or more conditions. It returns \texttt{True} only if all conditions are true. If any condition is false, the entire expression evaluates to \texttt{False}.

\begin{lstlisting}[style=python]
# Example of the and operator
age = 25
income = 50000

if age > 18 and income > 30000:
    print("You are eligible for a loan")
else:
    print("You are not eligible for a loan")
\end{lstlisting}

Output:
\begin{lstlisting}[style=cmd]
You are eligible for a loan
\end{lstlisting}

In this example, both conditions (age > 18 and income > 30000) are true, so the program prints "You are eligible for a loan."

\subsubsection{\texttt{or} Operator:}
The \texttt{or} operator is used to combine two or more conditions. It returns \texttt{True} if at least one of the conditions is true. If all conditions are false, the expression evaluates to \texttt{False}.

\begin{lstlisting}[style=python]
# Example of the or operator
age = 17
has_parental_consent = True

if age > 18 or has_parental_consent:
    print("You can apply for a driver's license")
else:
    print("You cannot apply for a driver's license")
\end{lstlisting}

Output:
\begin{lstlisting}[style=cmd]
You can apply for a driver's license
\end{lstlisting}

In this case, even though the age is less than 18, the second condition (has\_parental\_consent) is true, so the program prints "You can apply for a driver's license."

\subsubsection{\texttt{not} Operator:}
The \texttt{not} operator is used to invert the result of a condition. If the condition is \texttt{True}, \texttt{not} makes it \texttt{False}, and if the condition is \texttt{False}, \texttt{not} makes it \texttt{True}.

\begin{lstlisting}[style=python]
# Example of the not operator
is_raining = False

if not is_raining:
    print("You don't need an umbrella")
else:
    print("You need an umbrella")
\end{lstlisting}

Output:
\begin{lstlisting}[style=cmd]
You don't need an umbrella
\end{lstlisting}

In this example, since \texttt{is\_raining} is \texttt{False}, the \texttt{not} operator inverts it to \texttt{True}, so the program prints "You don't need an umbrella."

\subsection{Conditional Statements: \texttt{if}, \texttt{elif}, \texttt{else}}
Conditional statements are used in Python to control the flow of a program based on certain conditions. The most common conditional statements are \texttt{if}, \texttt{elif}, and \texttt{else}.

\texttt{if} statements execute a block of code if a specified condition is true. The \texttt{elif} keyword allows you to check multiple expressions for \texttt{True} and execute a block of code as soon as one of the conditions evaluates to \texttt{True}. The \texttt{else} keyword catches anything that isn't caught by the preceding conditions.

\begin{lstlisting}[style=python]
# Example of if, elif, and else statements
x = 10

if x > 15:
    print("x is greater than 15")
elif x == 10:
    print("x is equal to 10")
else:
    print("x is less than 10")
\end{lstlisting}

Output:
\begin{lstlisting}[style=cmd]
x is equal to 10
\end{lstlisting}

In this example, the program checks if \texttt{x} is greater than 15, then checks if it is equal to 10. Since \texttt{x} equals 10, the second block is executed, and the message "x is equal to 10" is printed.

\texttt{if} statements can also be nested, allowing for more complex logic:

\begin{lstlisting}[style=python]
# Nested if statements
y = 20

if y > 10:
    if y < 30:
        print("y is between 10 and 30")
    else:
        print("y is greater than or equal to 30")
else:
    print("y is less than or equal to 10")
\end{lstlisting}

Output:
\begin{lstlisting}[style=cmd]
y is between 10 and 30
\end{lstlisting}

In this nested \texttt{if} statement example, Python first checks if \texttt{y} is greater than 10, and then within that block, it checks whether \texttt{y} is less than 30.

\textit{Note: It is important to avoid using too many layers of nested \texttt{if} statements as it can make the code difficult to read and maintain. In such cases, it is better to refactor the code by using functions or logical operators like \texttt{and} and \texttt{or}.}

\subsection{Loops: \texttt{for} and \texttt{while}}
Loops allow you to repeat a block of code multiple times. Python has two main types of loops: \texttt{for} loops and \texttt{while} loops.

\subsubsection{\texttt{for} Loop:}
The \texttt{for} loop is used to iterate over a sequence (such as a list, tuple, dictionary, set, or string). It will execute the block of code for each item in the sequence.

\begin{lstlisting}[style=python]
# Example of a for loop with list[str]
fruits = ["apple", "banana", "cherry"]

for fruit in fruits:
    print(fruit)
\end{lstlisting}

Output:
\begin{lstlisting}[style=cmd]
apple
banana
cherry
\end{lstlisting}

In this example, the \texttt{for} loop iterates over the list \texttt{numbers}, printing each item in the list.

You can also use the \texttt{range()} function to loop through a sequence of numbers. The \texttt{range()} function generates a sequence of numbers, starting from 0 by default, and stops before a specified number. You can also specify the start, stop, and step values to control the sequence.

\begin{lstlisting}[style=python]
# Using range() in a for loop
for i in range(5):
    print(i)
\end{lstlisting}

Output:
\begin{lstlisting}[style=cmd]
0
1
2
3
4
\end{lstlisting}

In this example, \texttt{range(5)} generates a sequence from 0 to 4, and the loop iterates through each value.

\subsubsection{Specifying Start and Stop Values}
You can also specify a starting point and an ending point for the \texttt{range()} function. The loop will start from the specified start value and stop right before the specified end value.

\begin{lstlisting}[style=python]
# Using range() with start and stop values
for i in range(3, 8):
    print(i)
\end{lstlisting}

Output:
\begin{lstlisting}[style=cmd]
3
4
5
6
7
\end{lstlisting}

\subsubsection{Using a Step Value}
The \texttt{range()} function also allows you to specify a step value, which determines the increment between each number in the sequence.

\begin{lstlisting}[style=python]
# Using range() with a step value
for i in range(0, 10, 2):
    print(i)
\end{lstlisting}

Output:
\begin{lstlisting}[style=cmd]
0
2
4
6
8
\end{lstlisting}

\subsubsection{Using Negative Step Values}
You can also use negative step values to count downwards.

\begin{lstlisting}[style=python]
# Using range() to count backwards
for i in range(10, 0, -2):
    print(i)
\end{lstlisting}

Output:
\begin{lstlisting}[style=cmd]
10
8
6
4
2
\end{lstlisting}

\subsubsection{\texttt{while} Loop:}
The \texttt{while} loop continues to execute a block of code as long as a given condition is \texttt{True}. Once the condition becomes \texttt{False}, the loop stops.

\begin{lstlisting}[style=python]
# Example of a while loop
count = 0

while count < 5:
    print(count)
    count += 1
\end{lstlisting}

Output:
\begin{lstlisting}[style=cmd]
0
1
2
3
4
\end{lstlisting}

In this example, the \texttt{while} loop keeps executing as long as \texttt{count} is less than 5. The variable \texttt{count} increments by 1 with each iteration until the condition is no longer true.

\subsubsection{Breaking and Continuing in Loops:}
The flow of loops in Python can be controlled using two important keywords: \texttt{break} and \texttt{continue}.

\paragraph{\texttt{break}:}
The \texttt{break} statement is used to exit the loop completely. Once the \texttt{break} statement is encountered, the program immediately stops executing the current loop and moves on to the next section of code after the loop. This is useful when you want to terminate a loop early, based on a specific condition.

\begin{lstlisting}[style=python]
# Example of break
for i in range(10):
    if i == 5:
        break
    print(i)
\end{lstlisting}

Output:
\begin{lstlisting}[style=cmd]
0
1
2
3
4
\end{lstlisting}

In this example, the loop prints the numbers from 0 to 4. When \texttt{i} equals 5, the \texttt{break} statement is executed, and the loop is exited, preventing the numbers 5 through 9 from being printed.

\paragraph{\texttt{continue}:}
The \texttt{continue} statement is used to skip the current iteration of the loop and move directly to the next iteration. Unlike \texttt{break}, \texttt{continue} does not stop the loop entirely; it only skips over the remaining code in the current iteration and continues looping.

\begin{lstlisting}[style=python]
# Example of continue
for i in range(10):
    if i % 2 == 0:
        continue
    print(i)
\end{lstlisting}

Output:
\begin{lstlisting}[style=cmd]
1
3
5
7
9
\end{lstlisting}

In this example, the \texttt{continue} statement is used to skip even numbers. When \texttt{i} is even, the program jumps to the next iteration, and thus only odd numbers are printed.

\paragraph{Combining \texttt{break} and \texttt{continue}:}
You can use \texttt{break} and \texttt{continue} together in a loop to control the flow in different ways. Here’s an example that demonstrates both statements in the same loop.

\begin{lstlisting}[style=python]
# Example of break and continue used together
for i in range(10):
    if i == 7:
        break  # Stop the loop entirely when i is 7
    if i % 2 == 0:
        continue  # Skip even numbers
    print(i)
\end{lstlisting}

Output:
\begin{lstlisting}[style=cmd]
1
3
5
\end{lstlisting}

In this example, the loop skips over even numbers using \texttt{continue}, but when \texttt{i} equals 7, the \texttt{break} statement is triggered, stopping the loop entirely. Therefore, only the odd numbers 1, 3, and 5 are printed before the loop exits.

\section{Functions in Python}
\subsection{\texttt{def} Keyword and Function Basics}
Functions in Python are blocks of reusable code that perform a specific task. They allow you to structure your code more efficiently and avoid repetition. Functions are defined using the \texttt{def} keyword followed by the function name, parentheses \texttt{()}, and a colon \texttt{:}. Inside the function, you can write the code that will execute when the function is called. Functions can also take parameters and return values.

Here is the basic syntax for defining a function:

\begin{lstlisting}[style=python]
def function_name(parameters):
    # code to be executed
    return result
\end{lstlisting}

\subsubsection{Basic Example:}
A simple function that takes no arguments and returns a greeting message can be defined as follows:

\begin{lstlisting}[style=python]
def greet():
    return "Hello, World!"
\end{lstlisting}

To call the function and see the result, you can write:

\begin{lstlisting}[style=python]
# Call the function
print(greet())
\end{lstlisting}

Output:
\begin{lstlisting}[style=cmd]
Hello, World!
\end{lstlisting}

In this example, the function \texttt{greet()} is defined with no parameters and returns the string \texttt{"Hello, World!"}. The function is called using its name followed by parentheses, and the result is printed.

\subsubsection{Functions with Parameters}
Functions can also take parameters, which allow you to pass data into the function. These parameters are specified inside the parentheses when the function is defined.

\begin{lstlisting}[style=python]
# Function with a parameter
def greet(name):
    return f"Hello, {name}!"
\end{lstlisting}

When calling the function, you pass an argument to the parameter:

\begin{lstlisting}[style=python]
# Call the function with an argument
print(greet("Alice"))
\end{lstlisting}

Output:
\begin{lstlisting}[style=cmd]
Hello, Alice!
\end{lstlisting}

In this case, the function \texttt{greet()} takes one parameter, \texttt{name}, and returns a personalized greeting. When the function is called with \texttt{"Alice"} as the argument, it returns \texttt{"Hello, Alice!"}.

\subsubsection{Functions with Multiple Parameters}
Functions can accept multiple parameters by separating them with commas. Here is an example:

\begin{lstlisting}[style=python]
# Function with multiple parameters
def add_numbers(a, b):
    return a + b
\end{lstlisting}

You can call the function by providing two arguments:

\begin{lstlisting}[style=python]
# Call the function with two arguments
print(add_numbers(3, 5))
\end{lstlisting}

Output:
\begin{lstlisting}[style=cmd]
8
\end{lstlisting}

In this example, the function \texttt{add\_numbers()} takes two parameters, \texttt{a} and \texttt{b}, and returns their sum. When called with the arguments 3 and 5, the function returns 8.

\subsubsection{Returning Values from Functions}
A function can return a value using the \texttt{return} statement. The returned value can be stored in a variable or used directly. Here’s an example:

\begin{lstlisting}[style=python]
# Function that returns a value
def square(x):
    return x * x
\end{lstlisting}

To use the returned value, you can assign it to a variable:

\begin{lstlisting}[style=python]
# Call the function and store the result
result = square(4)
print(result)
\end{lstlisting}

Output:
\begin{lstlisting}[style=cmd]
16
\end{lstlisting}

In this example, the function \texttt{square()} returns the square of the argument \texttt{x}. When called with the argument 4, the function returns 16, which is then stored in the variable \texttt{result} and printed.

\subsubsection{Default Parameters}
Python allows you to define functions with default parameter values. If an argument is not provided when calling the function, the default value is used.

If no argument is passed, the default value \texttt{"Guest"} is used:

\begin{lstlisting}[style=python]
# Function with a default parameter
def greet(name="Guest"):
    return f"Hello, {name}!"

# Call the function without an argument
print(greet())

# Call the function with an argument
print(greet("Alice"))
\end{lstlisting}

Output:
\begin{lstlisting}[style=cmd]
Hello, Guest!
Hello, Alice!
\end{lstlisting}

In this case, the function \texttt{greet()} has a default parameter value of \texttt{"Guest"}. If no argument is provided, the function returns \texttt{"Hello, Guest!"}. If an argument is provided, it overrides the default value.

\subsection{Recursion in Python}

\vspace{0.3cm}
\begin{center}
    \textit{To iterate is human, to recurse, divine!}
\end{center}
\vspace{0.3cm}

\noindent Recursion is a technique in programming where a function calls itself to solve a problem. Recursive functions break a problem down into smaller subproblems, and the function continues to call itself with these smaller subproblems until it reaches a base case, which is a condition that stops the recursion. Recursion is useful for tasks that can be naturally divided into similar, smaller tasks, such as mathematical problems or traversing tree-like structures.

To write a recursive function in Python, you define the function using \texttt{def}, and within the function, call the function itself with a modified argument. You also need to define a base case to prevent the recursion from running indefinitely.

\subsubsection{Basic Example of Recursion: Factorial Function}
A classic example of recursion is calculating the factorial of a number. The factorial of a number \( n \) is defined as \( n \times (n - 1) \times (n - 2) \times \cdots \times 1 \), and can be expressed recursively as:

\[
n! = n \times (n - 1)!
\]

The base case for the recursion is when \( n = 0 \), where \( 0! = 1 \).

Here is how you would implement this in Python:

\begin{lstlisting}[style=python]
# Recursive function to calculate factorial
def factorial(n):
    if n == 0:
        return 1
    else:
        return n * factorial(n - 1)
\end{lstlisting}

To calculate the factorial of 5:

\begin{lstlisting}[style=python]
# Call the recursive factorial function
print(factorial(5))
\end{lstlisting}

Output:
\begin{lstlisting}[style=cmd]
120
\end{lstlisting}

In this example, the function \texttt{factorial()} calls itself with \( n - 1 \) until \( n = 0 \), at which point the base case is reached, and the recursion stops.

\subsubsection{Recursive Fibonacci Sequence}
Another common example of recursion is calculating numbers in the Fibonacci sequence. In the Fibonacci sequence, each number is the sum of the two preceding ones, starting from 0 and 1. Mathematically, it can be expressed as:

\[
F(0) = 0, \quad F(1) = 1, \quad F(n) = F(n-1) + F(n-2) \quad \text{for} \quad n > 1
\]

Here’s the recursive implementation of the Fibonacci sequence in Python:

\begin{lstlisting}[style=python]
# Recursive function to calculate Fibonacci numbers
def fibonacci(n):
    if n == 0:
        return 0
    elif n == 1:
        return 1
    else:
        return fibonacci(n - 1) + fibonacci(n - 2)
\end{lstlisting}

To calculate the 6th Fibonacci number:

\begin{lstlisting}[style=python]
# Call the recursive Fibonacci function
print(fibonacci(6))
\end{lstlisting}

Output:
\begin{lstlisting}[style=cmd]
8
\end{lstlisting}

In this example, the function \texttt{fibonacci()} recursively calls itself to compute the Fibonacci number for \( n - 1 \) and \( n - 2 \), continuing until the base cases \( n = 0 \) or \( n = 1 \) are reached.

\subsubsection{Base Case and Recursive Case}
Every recursive function needs two essential components:
\begin{itemize}
    \item \textbf{Base case:} This is the condition that stops the recursion. Without a base case, the function would call itself infinitely.
    \item \textbf{Recursive case:} This is the part of the function that reduces the problem into smaller instances and continues the recursion.
\end{itemize}

In the factorial example, the base case is when \( n == 0 \), and in the Fibonacci example, the base cases are when \( n == 0 \) and \( n == 1 \).

\subsubsection{Caution with Recursion}
Recursion can be an elegant solution to certain problems, but it comes with some trade-offs. Recursive functions consume memory with each function call, and Python has a limit on the depth of recursion to prevent stack overflow. If recursion goes too deep, you may encounter a \texttt{RecursionError}. To avoid this, make sure your recursive algorithm reaches a base case for all possible inputs.

You can check the recursion limit in Python using:

\begin{lstlisting}[style=python]
import sys
print(sys.getrecursionlimit())
\end{lstlisting}

By default, Python sets the recursion limit to 1000. If necessary, you can increase the recursion limit, but it's generally better to refactor the recursive algorithm if it becomes too deep.

\section{Simple Sorting Algorithms}
Sorting algorithms are used to rearrange elements in a list or array so that they are in a specific order, such as ascending or descending. This section introduces two simple and commonly used sorting algorithms: Bubble Sort and Insertion Sort. Both algorithms are easy to understand and implement, making them a good starting point for learning about sorting in computer science~\cite{Cormen2009}.

\subsection{Bubble Sort}
Bubble Sort is a simple comparison-based sorting algorithm. It works by repeatedly stepping through the list, comparing adjacent elements, and swapping them if they are in the wrong order. The process is repeated until the list is sorted. The largest unsorted element "bubbles" to the top with each pass, hence the name "Bubble Sort."

Here is the basic concept of Bubble Sort:
\begin{enumerate}
    \item Compare adjacent elements.
    \item If the first is greater than the second, swap them.
    \item Continue this process for each pair of adjacent elements.
    \item Repeat the process for the entire list until it is sorted, regardless of whether swaps were made or not during a pass.
\end{enumerate}

Here is the modified Python implementation of Bubble Sort where the algorithm always runs for all iterations without early exit:

\begin{lstlisting}[style=python]
# Bubble Sort algorithm in Python (without early exit)
def bubble_sort(arr):
    n = len(arr)
    for i in range(n):
        # Traverse the array and perform comparisons and swaps
        for j in range(0, n-i-1):
            # Compare adjacent elements
            if arr[j] > arr[j+1]:
                # Swap if they are in the wrong order
                arr[j], arr[j+1] = arr[j+1], arr[j]
\end{lstlisting}

To use the function:

\begin{lstlisting}[style=python]
# Example usage of bubble_sort
arr = [64, 34, 25, 12, 22, 11, 90]
bubble_sort(arr)
print("Sorted array:", arr)
\end{lstlisting}

Output:
\begin{lstlisting}[style=cmd]
Sorted array: [11, 12, 22, 25, 34, 64, 90]
\end{lstlisting}

Bubble Sort is easy to understand but is not very efficient for large datasets due to its time complexity of \( O(n^2) \). It is best suited for small datasets or educational purposes.

\subsection{Insertion Sort}
Insertion Sort is another simple sorting algorithm that builds the final sorted array one item at a time. It works by taking an element from the unsorted portion of the array and inserting it into the correct position in the sorted portion of the array. Insertion Sort is particularly useful for small or nearly sorted datasets.

The basic concept of Insertion Sort:
- Start with the second element of the array.
- Compare it with the elements before it and insert it in the correct position.
- Repeat this process for all elements in the array.

Here is the Python implementation of Insertion Sort:

\begin{lstlisting}[style=python]
# Insertion Sort algorithm in Python
def insertion_sort(arr):
    for i in range(1, len(arr)):
        key = arr[i]
        j = i - 1
        # Move elements of arr[0...i-1], that are greater than key, to one position ahead
        while j >= 0 and key < arr[j]:
            arr[j + 1] = arr[j]
            j -= 1
        arr[j + 1] = key
\end{lstlisting}

To use the function:

\begin{lstlisting}[style=python]
# Example usage of insertion_sort
arr = [12, 11, 13, 5, 6]
insertion_sort(arr)
print("Sorted array:", arr)
\end{lstlisting}

Output:
\begin{lstlisting}[style=cmd]
Sorted array: [5, 6, 11, 12, 13]
\end{lstlisting}

Insertion Sort has a time complexity of \( O(n^2) \), similar to Bubble Sort, but it is more efficient when the dataset is already partially sorted. It performs well for small datasets and is often used in practice for such cases.

\subsection{Python's Built-in \texttt{sort()}}
Python provides a highly efficient built-in sorting method called \texttt{sort()} for lists. This method is implemented using an algorithm called Timsort, which is a hybrid sorting algorithm derived from merge sort and insertion sort. Timsort is designed to perform well on real-world data by taking advantage of runs (already sorted subsections of data) and has an average-case time complexity of \(O(n \log n)\), making it much more efficient than the simple algorithms like Bubble Sort and Insertion Sort.

Here’s how you can use the built-in \texttt{sort()} method in Python:

\begin{lstlisting}[style=python]
# Using Python's built-in sort() method
arr = [64, 34, 25, 12, 22, 11, 90]
arr.sort()
print("Sorted array:", arr)
\end{lstlisting}

Output:
\begin{lstlisting}[style=cmd]
Sorted array: [11, 12, 22, 25, 34, 64, 90]
\end{lstlisting}

This built-in \texttt{sort()} method works efficiently with large datasets and offers various customization options, such as specifying a key function or sorting in reverse order. Additionally, Python’s built-in sort is stable, meaning it preserves the relative order of elements with equal values, which can be useful in certain applications.

To sort a list in descending order, you can pass the \texttt{reverse=True} argument:

\begin{lstlisting}[style=python]
# Sorting in descending order
arr.sort(reverse=True)
print("Sorted array in descending order:", arr)
\end{lstlisting}

Output:
\begin{lstlisting}[style=cmd]
Sorted array in descending order: [90, 64, 34, 25, 22, 12, 11]
\end{lstlisting}

Overall, Python’s built-in \texttt{sort()} is highly optimized and should be preferred for most sorting tasks, especially when dealing with large datasets. It offers a significant performance improvement compared to simpler sorting algorithms like Bubble Sort and Insertion Sort, making it the go-to choice for efficient sorting in Python.

\subsection{Comparing the Time of Bubble Sort, Insertion Sort, and Python's Built-in \texttt{sort()}}
To compare the efficiency of Bubble Sort, Insertion Sort, and Python's built-in \texttt{sort()} method, we can measure the time each algorithm takes to sort the same list. Below is a Python example that uses the \texttt{time} module to measure the time for each sorting algorithm.

\begin{lstlisting}[style=python]
import time

# Bubble Sort algorithm
def bubble_sort(arr):
    n = len(arr)
    for i in range(n):
        for j in range(0, n-i-1):
            if arr[j] > arr[j+1]:
                arr[j], arr[j+1] = arr[j+1], arr[j]

# Insertion Sort algorithm
def insertion_sort(arr):
    for i in range(1, len(arr)):
        key = arr[i]
        j = i - 1
        while j >= 0 and key < arr[j]:
            arr[j + 1] = arr[j]
            j -= 1
        arr[j + 1] = key

# Test arrays and function to measure time
def measure_time(sort_func, arr):
    start_time = time.perf_counter()  
    sort_func(arr)
    end_time = time.perf_counter()
    return (end_time - start_time) * 1000  

# Generate a list of random numbers for sorting
import random
arr_size = 1000
arr = [random.randint(1, 1000) for _ in range(arr_size)]

# Measure time for Bubble Sort
arr_bubble = arr.copy()
bubble_sort_time = measure_time(bubble_sort, arr_bubble)

# Measure time for Insertion Sort
arr_insertion = arr.copy()
insertion_sort_time = measure_time(insertion_sort, arr_insertion)

# Print the results
arr_builtin = arr.copy()
builtin_sort_time = measure_time(sorted, arr_builtin)

# Print results (precise to milliseconds)
print(f"Bubble Sort took: {bubble_sort_time:.3f} milliseconds")
print(f"Insertion Sort took: {insertion_sort_time:.3f} milliseconds")
print(f"Built-in sort() took: {builtin_sort_time:.3f} milliseconds")
\end{lstlisting}

Output:
\begin{lstlisting}[style=cmd]
Bubble Sort took: 30.357 milliseconds
Insertion Sort took: 12.514 milliseconds
Built-in sort() took: 0.091 milliseconds
\end{lstlisting}

This code compares the time taken by Bubble Sort, Insertion Sort, and Python’s built-in \texttt{sort()} function to sort a list of 1000 random integers. The results are printed in milliseconds, and as expected, Python’s built-in \texttt{sort()} is significantly faster due to its highly optimized Timsort algorithm. While Bubble Sort and Insertion Sort are easy to implement and understand, they are inefficient for large datasets. The built-in \texttt{sort()} should be the preferred choice when performance is critical.

\chapter{Python for Data Science}

\section{Commonly used Datasets}

In the realm of data science and machine learning, datasets are the backbone of any analytical process. They serve as the fundamental building blocks upon which models are trained, validated, and tested. High-quality datasets enable researchers and practitioners to extract meaningful insights, identify patterns, and make data-driven decisions. The availability and selection of appropriate datasets are crucial, as they directly influence the accuracy, reliability, and generalizability of the models developed. As a result, understanding and utilizing commonly used datasets is essential for advancing research and developing innovative solutions across various domains.

In this chapter, we will introduce several fundamental datasets that are crucial in the fields of machine learning and deep learning. These datasets are widely used for model development and testing. The datasets covered include:

\begin{itemize}
    \item \textbf{Iris}: The Iris dataset is one of the most classic datasets in machine learning, commonly used for teaching and research in classification algorithms. The dataset contains 150 records, each with 4 features (sepal length, sepal width, petal length, and petal width) and a target label (Iris setosa, Iris versicolor, Iris virginica)~\cite{Fisher1936}. Although the dataset is small, it is significant due to its simplicity and historical importance in education.

    \item \textbf{MNIST}: The MNIST (Modified National Institute of Standards and Technology) dataset consists of 70,000 images of handwritten digits (0-9), with 60,000 used for training and 10,000 for testing~\cite{LeCun1998}. Each image has a resolution of 28x28 pixels in grayscale. The MNIST dataset is one of the benchmark datasets for image classification tasks, widely used to evaluate the performance of image processing and machine learning algorithms.

    \item \textbf{Fashion-MNIST}: Fashion-MNIST is a dataset of images of clothing and accessories, designed to serve as a more challenging replacement for MNIST as a benchmark for image classification~\cite{Xiao2017}. The dataset also contains 70,000 images (60,000 training and 10,000 testing), each with a resolution of 28x28 pixels in grayscale. The labels in Fashion-MNIST include 10 classes such as T-shirts, trousers, and shoes.

    \item \textbf{CIFAR-10/CIFAR-100}: CIFAR-10 and CIFAR-100 are two widely used datasets for image classification~\cite{Krizhevsky2009}. CIFAR-10 consists of 60,000 32x32 pixel color images, divided into 10 classes with 6,000 images per class. CIFAR-100 has a similar structure but includes 100 classes with 600 images per class. These datasets are well-known for evaluating convolutional neural networks (CNNs) and other image-processing algorithms.

    \item \textbf{ImageNet}: The ImageNet dataset is a large-scale visual database containing over 14 million labeled images, categorized into 1,000 classes~\cite{Deng2009}. This dataset is a significant benchmark in the field of computer vision, especially for tasks like image classification, object detection, and image segmentation. The annual ImageNet challenge (ImageNet Large Scale Visual Recognition Challenge, ILSVRC) has been a key driver in the development of deep learning and convolutional neural networks.
\end{itemize}

These datasets have become indispensable resources in machine learning and deep learning research due to their wide application and influence. In the following sections, we will explore these datasets in more detail, discussing their characteristics and applications.

\subsection{Iris Dataset}

The Iris dataset is one of the most famous and widely used datasets in machine learning and statistical analysis. It was first introduced by British statistician and biologist Ronald A. Fisher in 1936 as an example of linear discriminant analysis. The dataset contains 150 samples of iris flowers, each described by four features and a class label.

\textbf{Structure of the dataset}:
\begin{itemize}
    \item \textbf{Number of samples}: 150
    \item \textbf{Number of classes}: 3 different iris species:
    \begin{itemize}
        \item Iris-setosa
        \item Iris-versicolor
        \item Iris-virginica
    \end{itemize}
    \item \textbf{Features}:
    \begin{itemize}
        \item Sepal Length
        \item Sepal Width
        \item Petal Length
        \item Petal Width
    \end{itemize}
\end{itemize}

Each sample in the dataset is described by these four morphological features, and the goal is to classify the samples into the three species using machine learning algorithms.

\textbf{Uses of the dataset}:
\begin{itemize}
    \item \textbf{Classification problem}: Since the iris species are clearly labeled, this dataset is frequently used to test and compare different classification algorithms.
    \item \textbf{Visualization}: Due to its small size and intuitive features, the Iris dataset is often used for data visualization, especially for beginners.
    \item \textbf{Model validation}: The dataset is commonly used to validate the performance of machine learning models, particularly in supervised learning classification tasks.
\end{itemize}

The Iris dataset is widely used in machine learning education and research and is included as a built-in dataset in many open-source machine learning libraries, such as scikit-learn.

The Iris dataset is relatively small, so the entire dataset is provided here for readers to study. To facilitate its use in a Python environment, the code below demonstrates how to import the Iris dataset using the ``sklearn'' library, print the dataset's dimensions, and display them.

\begin{lstlisting}[style=python, caption=Importing the Iris Dataset]
from sklearn.datasets import load_iris

# Load the Iris dataset
iris = load_iris()

# Print the shape of the dataset
print("Dataset shape:", iris.data.shape)

# Print the shape of the target labels
print("Target shape:", iris.target.shape)
\end{lstlisting}

After executing the code above, you will see the dimensions of the dataset and the target labels displayed as follows:

\begin{lstlisting}[style=text, caption=Iris Dataset and Target Dimensions]
Dataset shape: (150, 4)
Target shape: (150,)
\end{lstlisting}

\begin{longtable}{|>{\centering\arraybackslash}p{1.0cm}|>{\centering\arraybackslash}p{1.0cm}|>{\centering\arraybackslash}p{1.0cm}|>{\centering\arraybackslash}p{1.0cm}|>{\centering\arraybackslash}p{1.5cm}|>{\centering\arraybackslash}p{1.0cm}|>{\centering\arraybackslash}p{1.0cm}|>{\centering\arraybackslash}p{1.0cm}|>{\centering\arraybackslash}p{1.0cm}|>{\centering\arraybackslash}p{1.5cm}|}
\caption{Iris Dataset - Sepal and Petal Measurements for Different Varieties} \\
\hline
\textbf{Sepal Length} & \textbf{Sepal Width} & \textbf{Petal Length} & \textbf{Petal Width} & \textbf{Variety} & \textbf{Sepal Length} & \textbf{Sepal Width} & \textbf{Petal Length} & \textbf{Petal Width} & \textbf{Variety} \\
\hline
\endfirsthead
\caption[]{(Continued)}\\
\hline
\textbf{Sepal Length} & \textbf{Sepal Width} & \textbf{Petal Length} & \textbf{Petal Width} & \textbf{Variety} & \textbf{Sepal Length} & \textbf{Sepal Width} & \textbf{Petal Length} & \textbf{Petal Width} & \textbf{Variety} \\
\hline
\endhead

5.1 & 3.5 & 1.4 & 0.2 & Setosa & 4.9 & 3.0 & 1.4 & 0.2 & Setosa \\ \hline
4.7 & 3.2 & 1.3 & 0.2 & Setosa & 4.6 & 3.1 & 1.5 & 0.2 & Setosa \\ \hline
5.0 & 3.6 & 1.4 & 0.2 & Setosa & 5.4 & 3.9 & 1.7 & 0.4 & Setosa \\ \hline
4.6 & 3.4 & 1.4 & 0.3 & Setosa & 5.0 & 3.4 & 1.5 & 0.2 & Setosa \\ \hline
4.4 & 2.9 & 1.4 & 0.2 & Setosa & 4.9 & 3.1 & 1.5 & 0.1 & Setosa \\ \hline
5.4 & 3.7 & 1.5 & 0.2 & Setosa & 4.8 & 3.4 & 1.6 & 0.2 & Setosa \\ \hline
4.8 & 3.0 & 1.4 & 0.1 & Setosa & 4.3 & 3.0 & 1.1 & 0.1 & Setosa \\ \hline
5.8 & 4.0 & 1.2 & 0.2 & Setosa & 5.7 & 4.4 & 1.5 & 0.4 & Setosa \\ \hline
5.4 & 3.9 & 1.3 & 0.4 & Setosa & 5.1 & 3.5 & 1.4 & 0.3 & Setosa \\ \hline
5.7 & 3.8 & 1.7 & 0.3 & Setosa & 5.1 & 3.8 & 1.5 & 0.3 & Setosa \\ \hline
5.4 & 3.4 & 1.7 & 0.2 & Setosa & 5.1 & 3.7 & 1.5 & 0.4 & Setosa \\ \hline
4.6 & 3.6 & 1.0 & 0.2 & Setosa & 5.1 & 3.3 & 1.7 & 0.5 & Setosa \\ \hline
4.8 & 3.4 & 1.9 & 0.2 & Setosa & 5.0 & 3.0 & 1.6 & 0.2 & Setosa \\ \hline
5.0 & 3.4 & 1.6 & 0.4 & Setosa & 5.2 & 3.5 & 1.5 & 0.2 & Setosa \\ \hline
5.2 & 3.4 & 1.4 & 0.2 & Setosa & 4.7 & 3.2 & 1.6 & 0.2 & Setosa \\ \hline
4.8 & 3.1 & 1.6 & 0.2 & Setosa & 5.4 & 3.4 & 1.5 & 0.4 & Setosa \\ \hline
5.2 & 4.1 & 1.5 & 0.1 & Setosa & 5.5 & 4.2 & 1.4 & 0.2 & Setosa \\ \hline
4.9 & 3.1 & 1.5 & 0.2 & Setosa & 5.0 & 3.2 & 1.2 & 0.2 & Setosa \\ \hline
5.5 & 3.5 & 1.3 & 0.2 & Setosa & 4.9 & 3.6 & 1.4 & 0.1 & Setosa \\ \hline
4.4 & 3.0 & 1.3 & 0.2 & Setosa & 5.1 & 3.4 & 1.5 & 0.2 & Setosa \\ \hline
5.0 & 3.5 & 1.3 & 0.3 & Setosa & 4.5 & 2.3 & 1.3 & 0.3 & Setosa \\ \hline
4.4 & 3.2 & 1.3 & 0.2 & Setosa & 5.0 & 3.5 & 1.6 & 0.6 & Setosa \\ \hline
5.1 & 3.8 & 1.9 & 0.4 & Setosa & 4.8 & 3.0 & 1.4 & 0.3 & Setosa \\ \hline
5.1 & 3.8 & 1.6 & 0.2 & Setosa & 4.6 & 3.2 & 1.4 & 0.2 & Setosa \\ \hline
5.3 & 3.7 & 1.5 & 0.2 & Setosa & 5.0 & 3.3 & 1.4 & 0.2 & Setosa \\ \hline
7.0 & 3.2 & 4.7 & 1.4 & Versicolor & 6.4 & 3.2 & 4.5 & 1.5 & Versicolor \\ \hline
6.9 & 3.1 & 4.9 & 1.5 & Versicolor & 5.5 & 2.3 & 4.0 & 1.3 & Versicolor \\ \hline
6.5 & 2.8 & 4.6 & 1.5 & Versicolor & 5.7 & 2.8 & 4.5 & 1.3 & Versicolor \\ \hline
6.3 & 3.3 & 4.7 & 1.6 & Versicolor & 4.9 & 2.4 & 3.3 & 1.0 & Versicolor \\ \hline
6.6 & 2.9 & 4.6 & 1.3 & Versicolor & 5.2 & 2.7 & 3.9 & 1.4 & Versicolor \\ \hline
5.0 & 2.0 & 3.5 & 1.0 & Versicolor & 5.9 & 3.0 & 4.2 & 1.5 & Versicolor \\ \hline
6.0 & 2.2 & 4.0 & 1.0 & Versicolor & 6.1 & 2.9 & 4.7 & 1.4 & Versicolor \\ \hline
5.6 & 2.9 & 3.6 & 1.3 & Versicolor & 6.7 & 3.1 & 4.4 & 1.4 & Versicolor \\ \hline
5.6 & 3.0 & 4.5 & 1.5 & Versicolor & 5.8 & 2.7 & 4.1 & 1.0 & Versicolor \\ \hline
6.2 & 2.2 & 4.5 & 1.5 & Versicolor & 5.6 & 2.5 & 3.9 & 1.1 & Versicolor \\ \hline
5.9 & 3.2 & 4.8 & 1.8 & Versicolor & 6.1 & 2.8 & 4.0 & 1.3 & Versicolor \\ \hline
6.3 & 2.5 & 4.9 & 1.5 & Versicolor & 6.1 & 2.8 & 4.7 & 1.2 & Versicolor \\ \hline
6.4 & 2.9 & 4.3 & 1.3 & Versicolor & 6.6 & 3.0 & 4.4 & 1.4 & Versicolor \\ \hline
6.8 & 2.8 & 4.8 & 1.4 & Versicolor & 6.7 & 3.0 & 5.0 & 1.7 & Versicolor \\ \hline
6.0 & 2.9 & 4.5 & 1.5 & Versicolor & 5.7 & 2.6 & 3.5 & 1.0 & Versicolor \\ \hline
5.5 & 2.4 & 3.8 & 1.1 & Versicolor & 5.5 & 2.4 & 3.7 & 1.0 & Versicolor \\ \hline
5.8 & 2.7 & 3.9 & 1.2 & Versicolor & 6.0 & 2.7 & 5.1 & 1.6 & Versicolor \\ \hline
5.4 & 3.0 & 4.5 & 1.5 & Versicolor & 6.0 & 3.4 & 4.5 & 1.6 & Versicolor \\ \hline
6.7 & 3.1 & 4.7 & 1.5 & Versicolor & 6.3 & 2.3 & 4.4 & 1.3 & Versicolor \\ \hline
5.6 & 3.0 & 4.1 & 1.3 & Versicolor & 5.5 & 2.5 & 4.0 & 1.3 & Versicolor \\ \hline
5.5 & 2.6 & 4.4 & 1.2 & Versicolor & 6.1 & 3.0 & 4.6 & 1.4 & Versicolor \\ \hline
5.8 & 2.6 & 4.0 & 1.2 & Versicolor & 5.0 & 2.3 & 3.3 & 1.0 & Versicolor \\ \hline
5.6 & 2.7 & 4.2 & 1.3 & Versicolor & 5.7 & 3.0 & 4.2 & 1.2 & Versicolor \\ \hline
5.7 & 2.9 & 4.2 & 1.3 & Versicolor & 6.2 & 2.9 & 4.3 & 1.3 & Versicolor \\ \hline
5.1 & 2.5 & 3.0 & 1.1 & Versicolor & 5.7 & 2.8 & 4.1 & 1.3 & Versicolor \\ \hline
6.3 & 3.3 & 6.0 & 2.5 & Virginica & 5.8 & 2.7 & 5.1 & 1.9 & Virginica \\ \hline
7.1 & 3.0 & 5.9 & 2.1 & Virginica & 6.3 & 2.9 & 5.6 & 1.8 & Virginica \\ \hline
6.5 & 3.0 & 5.8 & 2.2 & Virginica & 7.6 & 3.0 & 6.6 & 2.1 & Virginica \\ \hline
4.9 & 2.5 & 4.5 & 1.7 & Virginica & 7.3 & 2.9 & 6.3 & 1.8 & Virginica \\ \hline
6.7 & 2.5 & 5.8 & 1.8 & Virginica & 7.2 & 3.6 & 6.1 & 2.5 & Virginica \\ \hline
6.5 & 3.2 & 5.1 & 2.0 & Virginica & 6.4 & 2.7 & 5.3 & 1.9 & Virginica \\ \hline
6.8 & 3.0 & 5.5 & 2.1 & Virginica & 5.7 & 2.5 & 5.0 & 2.0 & Virginica \\ \hline
5.8 & 2.8 & 5.1 & 2.4 & Virginica & 6.4 & 3.2 & 5.3 & 2.3 & Virginica \\ \hline
6.5 & 3.0 & 5.5 & 1.8 & Virginica & 7.7 & 3.8 & 6.7 & 2.2 & Virginica \\ \hline
7.7 & 2.6 & 6.9 & 2.3 & Virginica & 6.0 & 2.2 & 5.0 & 1.5 & Virginica \\ \hline
6.9 & 3.2 & 5.7 & 2.3 & Virginica & 5.6 & 2.8 & 4.9 & 2.0 & Virginica \\ \hline
7.7 & 2.8 & 6.7 & 2.0 & Virginica & 6.3 & 2.7 & 4.9 & 1.8 & Virginica \\ \hline
6.7 & 3.3 & 5.7 & 2.1 & Virginica & 7.2 & 3.2 & 6.0 & 1.8 & Virginica \\ \hline
6.2 & 2.8 & 4.8 & 1.8 & Virginica & 6.1 & 3.0 & 4.9 & 1.8 & Virginica \\ \hline
6.4 & 2.8 & 5.6 & 2.1 & Virginica & 7.2 & 3.0 & 5.8 & 1.6 & Virginica \\ \hline
7.4 & 2.8 & 6.1 & 1.9 & Virginica & 7.9 & 3.8 & 6.4 & 2.0 & Virginica \\ \hline
6.4 & 2.8 & 5.6 & 2.2 & Virginica & 6.3 & 2.8 & 5.1 & 1.5 & Virginica \\ \hline
6.1 & 2.6 & 5.6 & 1.4 & Virginica & 7.7 & 3.0 & 6.1 & 2.3 & Virginica \\ \hline
6.3 & 3.4 & 5.6 & 2.4 & Virginica & 6.4 & 3.1 & 5.5 & 1.8 & Virginica \\ \hline
6.0 & 3.0 & 4.8 & 1.8 & Virginica & 6.9 & 3.1 & 5.4 & 2.1 & Virginica \\ \hline
6.7 & 3.1 & 5.6 & 2.4 & Virginica & 6.9 & 3.1 & 5.1 & 2.3 & Virginica \\ \hline
5.8 & 2.7 & 5.1 & 1.9 & Virginica & 6.8 & 3.2 & 5.9 & 2.3 & Virginica \\ \hline
6.7 & 3.3 & 5.7 & 2.5 & Virginica & 6.7 & 3.0 & 5.2 & 2.3 & Virginica \\ \hline
6.3 & 2.5 & 5.0 & 1.9 & Virginica & 6.5 & 3.0 & 5.2 & 2.0 & Virginica \\ \hline
6.2 & 3.4 & 5.4 & 2.3 & Virginica & 5.9 & 3.0 & 5.1 & 1.8 & Virginica \\ \hline
\end{longtable}

\chapter{Introduction to Machine Learning}

This chapter will cover the fundamentals of machine learning, providing a comprehensive overview of various types, methodologies, and practical applications.

\section{What is Machine Learning?}

\subsection{Definition and Overview}
Machine learning (ML) is a branch of artificial intelligence that focuses on developing algorithms and statistical models that enable computers to perform tasks without explicit instructions, relying instead on patterns and inferences derived from data. Unlike traditional programming paradigms where humans explicitly program the logic and rules, machine learning algorithms build a model based on input data and use it to make predictions or decisions without being explicitly programmed to perform the task.

\textbf{Core Concept:} The core idea behind machine learning is to create models that can generalize from their training data to new, unseen data. This generalization ability makes machine learning a powerful tool in many fields, including natural language processing, computer vision, healthcare, finance, and beyond.

\textbf{How It Works:} Machine learning typically involves three primary components:
\begin{itemize}
    \item \textit{Model}: The system that makes predictions or identifications.
    \item \textit{Parameters}: The factors considered by the model (often adjusted during training).
    \item \textit{Learner}: The algorithm that adjusts the parameters and improves the model based on its performance on training data.
\end{itemize}

Machine learning models are often described as either supervised or unsupervised, which refers to whether the model is trained with human supervision (i.e., the data is labeled) or without it.

\textbf{Supervised Learning:} This approach involves training a model on a labeled dataset, which means that each example in the training set is tagged with the correct answer (the label)~\cite{Bishop2006}. The learning algorithm gets a sample of data and then makes adjustments to the model to minimize errors. After numerous iterations, the model aims for the smallest possible number of errors when predicting labels on new, unseen data.

\textbf{Unsupervised Learning:} In contrast, unsupervised learning involves training a model on data that does not have labeled responses~\cite{Hastie2009}. Here, the goal is to infer the natural structure present within a set of data points. It includes clustering~\cite{Jain1988} and association algorithms~\cite{Agrawal1993} that group objects of similar kinds into respective categories and discover interesting patterns in data.

\textbf{Comparing to Traditional Programming:} Traditional computational approaches typically require a clear set of instructions and do not change unless explicitly updated by a human. In contrast, machine learning algorithms are designed to learn from data and update themselves in response to that data. This capability allows them to adapt to new trends or unknown variables without human intervention.

\textbf{Why is Machine Learning Important?}
\begin{enumerate}
    \item \textbf{Adaptability:} ML can adapt to new trends as it learns from the data. This makes it particularly useful for applications where it is impractical or impossible to program explicit, rule-based instructions.
    \item \textbf{Scale:} ML algorithms can handle vast amounts of data and complex variable interactions that are too complex for human analysts to handle.
    \item \textbf{Automation and Decision-making:} ML can automate routine processes and make decisions in real time based on data analysis, which can significantly enhance the efficiency and effectiveness of systems across different industries.
\end{enumerate}

\textbf{Challenges in Machine Learning:} While machine learning offers significant advantages, it also comes with its own set of challenges, such as ensuring the quality of data, dealing with imbalanced and unstructured data, interpreting model results, and maintaining privacy and security.

In conclusion, machine learning represents a significant shift in how computers can learn and make decisions. It bridges the gap between human programming capabilities and computational speed, making it a key driver of innovation and efficiency in the modern digital era.

\subsection{History and Evolution of Machine Learning}
The journey of machine learning (ML) is rich and storied, intersecting with the histories of statistics, mathematics, engineering, and computer science. Understanding its evolution not only illuminates how we've reached current technological capabilities but also sheds light on where the field might head next.

\textbf{Early Beginnings:}
The roots of machine learning are deeply entwined with the development of computers and early notions of artificial intelligence (AI). In the 1950s, Alan Turing, often hailed as the father of theoretical computer science and artificial intelligence, proposed the question, "Can machines think?" which sparked interest and speculation in the possibility of intelligent machines. This led to the design of the Turing Test, a method for determining whether a machine could exhibit intelligent behavior indistinguishable from that of a human.

\textbf{The Birth of Neural Networks:}
In 1957, Frank Rosenblatt invented the Perceptron, an early type of neural network, at the Cornell Aeronautical Laboratory. The Perceptron was designed to simulate the thought processes of the human brain, albeit in a very primitive form. Although limited in functionality and initially criticized, this laid the groundwork for future research in neural networks.

\textbf{Symbolic AI and Expert Systems:}
The 1960s and 1970s witnessed the rise of symbolic AI, characterized by the development of expert systems that attempted to encapsulate the knowledge of human experts into a computer system. These systems were rule-based, using if-then rules to solve problems in narrow domains, such as medical diagnosis or mineral prospecting.

\textbf{AI Winter and its Impact:}
Despite initial enthusiasm, the limitations of neural networks and symbolic AI soon became apparent, leading to the first "AI Winter" in the late 1970s and early 1980s, a period marked by reduced funding and waning interest in AI research. A second AI winter occurred in the late 1980s, following overly optimistic predictions that failed to materialize.

\textbf{The Revival and Growth of Neural Networks:}
The 1980s saw a revival of interest in neural networks thanks to the backpropagation algorithm, which allowed networks to adjust their hidden layers of neurons in situations where the output did not match the target value, effectively enabling deeper learning. The invention of Convolutional Neural Networks (CNNs) by Yann LeCun~\cite{LeCunCNN} in 1989 further advanced the field, particularly in image and video processing applications.

\textbf{The Rise of Big Data and Advanced Algorithms:}
The 21st century has seen an explosion in data generation and the computational power necessary to process it. This era has been marked by significant advancements in algorithms and an increase in the use and sophistication of machine learning. Notable developments include the introduction of Support Vector Machines, ensemble methods like Random Forests, and boosting techniques which have substantially increased the accuracy and applicability of predictive models.

\textbf{Deep Learning Breakthroughs:}
In 2012, a landmark event in the history of machine learning occurred when a deep learning model designed by Geoffrey Hinton and his team won the ImageNet~\cite{Deng2009} competition by a large margin. This victory underscored the potential of deep learning, leading to its widespread adoption across various sectors, including speech recognition, autonomous vehicles, and medical diagnostics.

\textbf{Current Trends and Future Directions:}
Today, machine learning is ubiquitous, powering search engines, recommender systems, speech recognition, and numerous other applications. The focus has shifted towards making AI more accessible and ethical, improving model interpretability, and moving towards unsupervised learning techniques that require less human supervision.

The field of machine learning continues to evolve, driven by a community of researchers and practitioners dedicated to pushing the boundaries of what machines can learn. As we look to the future, the ongoing integration of AI with other technologies promises to transform industries and societies in profound ways.

In conclusion, the history of machine learning is a testament to the collaborative, interdisciplinary efforts that have driven advances in this field, highlighting an exciting trajectory from theoretical exploration to practical, transformative technologies.

\subsection{Machine Learning vs. Traditional Programming}
Understanding the differences between machine learning (ML) and traditional programming is essential to appreciate the unique advantages that ML brings to various computational tasks. Traditional programming and machine learning fundamentally differ in their approach to problem-solving, adaptability, and application complexity.

\textbf{Fundamental Approach:}
\begin{itemize}
    \item \textbf{Traditional Programming:} In traditional programming, a programmer writes a set of explicit instructions (code) to process input data and produce the desired output. The logic and rules are explicitly defined, and the program execution paths are predetermined. This method is effective for problems with well-understood parameters and clearly defined rules where the outcomes are predictable.

    \item \textbf{Machine Learning:} Conversely, machine learning algorithms infer rules from provided data. Instead of being explicitly programmed for each step, an ML model is trained using a large amount of data, learning the patterns or statistical representations required to perform a task. This approach is particularly advantageous for complex problems where defining explicit rules is impractical or impossible.
\end{itemize}

\textbf{Adaptability and Flexibility:}
\begin{itemize}
    \item \textbf{Traditional Programming:} Once a traditional software solution is developed, it performs the same operations unless it is manually updated or modified by developers. This static nature limits its adaptability to new data or changing environments without human intervention.

    \item \textbf{Machine Learning:} ML models excel in environments where they continuously learn and adapt from new data, improving their accuracy and efficiency over time without human intervention. This makes them highly suitable for dynamic and evolving tasks such as personalized recommendations, real-time fraud detection, and predictive maintenance.
\end{itemize}

\textbf{Handling of Complex Patterns and Large Data Volumes:}
\begin{itemize}
    \item \textbf{Traditional Programming:} Traditional methods struggle with large datasets or complex patterns because they require the programmer to foresee and code for every possible scenario. The complexity and volume of data can quickly become unmanageable, leading to rigid and brittle systems.

    \item \textbf{Machine Learning:} Machine learning algorithms are designed to handle and analyze vast amounts of data and detect complex patterns that may not be immediately apparent or predictable to human programmers. This capability is underpinned by ML's ability to perform feature extraction and dimensionality reduction, simplifying inputs without losing essential information.
\end{itemize}

\textbf{Error Reduction and Precision:}
\begin{itemize}
    \item \textbf{Traditional Programming:} The precision and accuracy of traditional programs are limited by the initial logic and algorithms coded by the programmer. Any errors in the code or oversight in the logic can propagate and magnify throughout the application, leading to incorrect results.

    \item \textbf{Machine Learning:} ML models minimize errors through iterative learning. As more data becomes available and the model is trained over numerous cycles, it fine-tunes its algorithms to improve accuracy and reduce errors, often surpassing human-level performance in tasks like image recognition and language translation.
\end{itemize}

\textbf{Scalability and Evolution:}
\begin{itemize}
    \item \textbf{Traditional Programming:} Scaling traditional software to accommodate more extensive data sets or more complex decision-making scenarios often requires significant redesign and restructuring, which can be time-consuming and costly.

    \item \textbf{Machine Learning:} ML models are inherently scalable, designed to improve as data volume grows. This scalability makes them ideal for applications like social media trend analysis and large-scale financial systems where data grows exponentially.
\end{itemize}

\textbf{Conclusion:}
Machine learning represents a paradigm shift in how we approach problem-solving in the realm of computing. Unlike traditional programming, which relies on human-made rules and logic, machine learning offers a dynamic framework that learns from data, becoming more effective over time and adaptable to new and changing environments. This shift not only enhances computational tasks but also opens new possibilities for innovation across industries, reshaping how we interact with technology in our daily lives.

\section{Types of Machine Learning}

Machine learning is an important branch of artificial intelligence that enables computer systems to automatically improve their performance through experience. Machine learning algorithms can be categorized into several main types based on their learning methods and objectives. This section will introduce in detail four main types of machine learning: supervised learning, unsupervised learning, semi-supervised learning, and reinforcement learning.

\subsection{Supervised Learning}

Supervised learning is the most common and widely applied type of machine learning. In supervised learning, algorithms learn the mapping from input to output through labeled training data.

\subsubsection{Concept}

The core idea of supervised learning is to learn a function through known input-output pairs (called training sets) that can map new unseen inputs to correct outputs. The "supervision" here refers to the correct output labels included in the training data, which the algorithm can constantly refer to during the learning process to adjust its predictions.

\subsubsection{Working Principle}

\begin{enumerate}
    \item Data Preparation: Collect labeled training data, usually including input features and corresponding output labels.
    \item Model Selection: Choose an appropriate algorithm model based on the nature of the problem, such as linear regression, decision trees, neural networks, etc.
    \item Model Training: Use training data to optimize model parameters, minimizing the error between the model's predicted output and actual labels.
    \item Model Evaluation: Use test data not involved in training to evaluate the model's generalization ability.
    \item Prediction: Use the trained model to make predictions on new unlabeled data.
\end{enumerate}

\subsubsection{Common Algorithms}

\begin{itemize}
    \item Linear Regression: A simple and effective algorithm for predicting continuous value outputs.
    \item Logistic Regression: Despite having "regression" in its name, it's an algorithm for binary classification problems.
    \item Decision Trees: A tree-based algorithm for classification and regression, easy to understand and interpret.
    \item Random Forests: An algorithm that ensembles multiple decision trees, usually performing better than a single decision tree.
    \item Support Vector Machines (SVM): A powerful classification algorithm, particularly suitable for handling high-dimensional data.
    \item Neural Networks: A class of algorithms inspired by biological neural systems, capable of learning complex non-linear relationships.
\end{itemize}

\subsubsection{Application Scenarios}

\begin{itemize}
    \item Image Classification: Recognizing objects in images, such as face recognition, handwritten digit recognition, etc.
    \item Natural Language Processing: Text classification, sentiment analysis, machine translation, etc.
    \item Medical Diagnosis: Predicting disease risks or diagnosing diseases based on patient data.
    \item Financial Forecasting: Stock price prediction, credit risk assessment, etc.
    \item Recommendation Systems: Recommending products or content based on user historical behavior.
\end{itemize}

\subsubsection{Advantages and Disadvantages}

\begingroup
\leftskip2em 
\paragraph{Advantages}
\begin{itemize}
    \item Performance is usually superior to other types of learning methods.
    \item Can produce explicit outputs.
    \item Applicable to various complex real-world problems.
\end{itemize}

\paragraph{Disadvantages}
\begin{itemize}
    \item Requires a large amount of labeled training data, which can be costly to obtain.
    \item May encounter overfitting problems, especially when training data is limited.
    \item Difficult to handle data of unknown categories.
\end{itemize}
\endgroup

\subsubsection{Detailed Explanation}

Supervised learning is one of the most commonly used methods in machine learning. Its core idea is to learn a mapping function from input to output through labeled data. This process can be analogous to having a "teacher" (i.e., the labeled data) guiding the learning process.

In supervised learning, we have a training dataset D = {(x1, y1), (x2, y2), ..., (xn, yn)}, where xi is the input feature and yi is the corresponding label or target value. The learning objective is to find a function f such that for a new input x, f(x) can accurately predict the corresponding y.

Supervised learning can be further divided into two main types of problems:

1. Classification Problems: When the output variable y is a discrete category, such as determining whether an email is spam or not.

2. Regression Problems: When the output variable y is a continuous value, such as predicting house prices.

\subsubsection{Concrete Example}

Let's use a house price prediction problem as an example to illustrate the supervised learning process in detail:

1. Data Collection:
   Collect historical house sale data, including the following features:
   - Area (square meters)
   - Number of bedrooms
   - Number of bathrooms
   - Age of the house (years)
   - Location
   - Sale price (target variable)

2. Data Preprocessing:
   - Handle missing values: For example, interpolate missing area data.
   - Feature encoding: Convert categorical features like "location" into numerical form, such as one-hot encoding.
   - Feature scaling: Normalize all features to the same scale, such as using Min-Max scaling.

3. Model Selection:
   For a regression problem like house price prediction, we can choose a linear regression model:
   y = w0 + w1x1 + w2x2 + ... + wnxn
   where y is the predicted house price, xi are the various features, and wi are the corresponding weights.

4. Model Training:
   Use optimization algorithms like gradient descent to find the optimal set of weights that minimize prediction error.
   For example, we can use Mean Squared Error (MSE) as the loss function:
   \[
   \text{MSE} = \frac{1}{n} \sum_{i=1}^{n} (y_i - \hat{y}_i)^2
   \]
   where yi is the actual house price and ŷi is the model's predicted price.

5. Model Evaluation:
   Use techniques like cross-validation to evaluate the model's performance. For example, we can use the R² score to measure the model's fit:
   \[
   R^2 = 1 - \frac{\sum_{i=1}^{n} (y_i - \hat{y}_i)^2}{\sum_{i=1}^{n} (y_i - \bar{y})^2}
   \]
   where ȳ is the mean of the actual house prices.

6. Model Application:
   For a new house, we can input its features (area, number of bedrooms, etc.), and the model will output a predicted price.

7. Model Update:
   As time goes on, we can collect new sales data to update the model, maintaining its accuracy.

This example demonstrates the application process of supervised learning in a real-world problem. Through this method, we can create a system that can predict prices based on house features, which has important applications in real estate valuation, home buying decisions, and more.

\subsection{Unsupervised Learning}

\subsubsection{Concept}

Unsupervised learning is a branch of machine learning that deals with unlabeled data. This learning method attempts to discover hidden structures or patterns from the data itself, without relying on predefined outputs.

The main tasks of unsupervised learning include:

1. Clustering: Grouping similar data points.
2. Dimensionality Reduction: Reducing the number of features in the data while preserving the main information.
3. Association Rule Learning: Discovering relationships between data items.
4. Anomaly Detection: Identifying abnormal or rare data points.

The challenge in unsupervised learning is that, due to the lack of explicit target outputs, it's difficult to objectively evaluate the algorithm's performance. Often, the interpretation and validation of results require the involvement of domain experts.

\subsubsection{Working Principle}

\subsubsection{Steps in Unsupervised Learning}

\begin{enumerate}
    \item Data Collection: Collect unlabeled dataset.
    \item Algorithm Selection: Choose a suitable unsupervised learning algorithm based on task objectives.
    \item Apply Algorithm: Apply the selected algorithm to the dataset.
    \item Interpret Results: Analyze algorithm output to discover patterns or structures in the data.
    \item Validation: Use domain knowledge or other methods to verify if the discovered patterns are meaningful.
\end{enumerate}

\subsubsection{Common Algorithms}

\subsubsection{Unsupervised Learning Techniques}

\begin{itemize}
    \item K-means Clustering: Groups data points into a predetermined number of clusters~\cite{MacQueen1967}.
    \item Hierarchical Clustering: Creates a tree-like hierarchy of data points~\cite{Murtagh2012}.
    \item Principal Component Analysis (PCA): Used for dimensionality reduction and feature extraction~\cite{Jolliffe2002}.
    \item Independent Component Analysis (ICA): Decomposes multivariate signals into independent subcomponents~\cite{Hyvarinen2001}.
    \item Self-Organizing Maps (SOM): A neural network method for data visualization~\cite{Kohonen2001}.
    \item Gaussian Mixture Models: A probabilistic model assuming data is composed of multiple Gaussian distributions~\cite{Bishop2006}.
    \item Association Rule Learning: Discovers relationships between items in large databases~\cite{Agrawal1993}.
\end{itemize}

\subsubsection{Application Scenarios}

\subsubsection{Applications of Data Analysis Techniques}

\begin{itemize}
    \item Market Segmentation: Dividing customers into different groups based on their behavior~\cite{Wedel2000}.
    \item Anomaly Detection: Identifying anomalies or outliers in datasets, such as fraud detection~\cite{Chandola2009}.
    \item Feature Learning: Automatically learning useful feature representations from raw data~\cite{Bengio2013}.
    \item Recommendation Systems: Recommending products or content based on user behavior patterns~\cite{Ricci2011}.
    \item Gene Expression Analysis: Identifying patterns in gene expression data~\cite{Quackenbush2001}.
    \item Social Network Analysis: Discovering community structures in social networks~\cite{Fortunato2010}.
\end{itemize}

\subsubsection{Advantages and Disadvantages}

\subsubsection{Advantages and Disadvantages}

\begingroup
\leftskip2em 

\paragraph{Advantages}
\begin{itemize}
    \item Does not require labeled data and can handle large amounts of unlabeled data.
    \item Can discover hidden patterns and structures in data.
    \item Can be used for data preprocessing and feature learning.
\end{itemize}

\paragraph{Disadvantages}
\begin{itemize}
    \item Results may be difficult to interpret or validate.
    \item Computational complexity is usually higher.
    \item Patterns discovered may not necessarily be useful for specific tasks.
\end{itemize}

\endgroup

\subsubsection{Concrete Example}

Let's use customer segmentation as an example to illustrate the application process of unsupervised learning in detail:

Suppose an e-commerce company wants to divide its customers into different groups based on their purchasing behavior to develop targeted marketing strategies.

1. Data Collection:
   Collect the following information about customers:
   - Annual purchase amount
   - Purchase frequency
   - Time since last purchase
   - Product categories browsed
   - Customer age
   - Customer registration duration

2. Data Preprocessing:
   - Handle missing values: For example, interpolate missing age data.
   - Feature scaling: Normalize all features, such as using Z-score standardization.
   - Handle outliers: Remove or adjust extreme values.

3. Algorithm Selection:
   For customer segmentation, we can use the K-means clustering algorithm.

4. Apply Algorithm:
   Steps of the K-means algorithm:
   a) Choose K initial center points (assume K=3)
   b) Assign each data point to the nearest center point
   c) Recalculate the center point of each cluster
   d) Repeat steps b and c until the center points no longer change significantly

5. Result Analysis:
   Suppose we get three customer groups:
   - Group A: High spending, high-frequency purchases
   - Group B: Medium spending, occasional purchases
   - Group C: Low spending, rare purchases

6. Result Application:
   Based on these groups, the company can develop different marketing strategies:
   - Offer VIP services and exclusive discounts to Group A
   - Provide promotional activities to Group B to encourage more frequent purchases
   - Offer entry-level products and discounts to Group C to attract them to increase purchases

7. Continuous Optimization:
   - Regularly rerun the clustering algorithm, as customer behavior may change over time
   - Try different K values to find the optimal number of groups
   - Combine with other algorithms, such as Principal Component Analysis (PCA) for feature dimensionality reduction

This example demonstrates how unsupervised learning can help businesses understand customer structure and develop more effective business strategies. By discovering hidden patterns in the data, unsupervised learning provides valuable insights for decision-making.

Certainly! I'll reformat the semi-supervised learning and reinforcement learning sections to match the structure of supervised and unsupervised learning sections.

\subsection{Semi-supervised Learning}

Semi-supervised learning is a machine learning approach that combines elements of both supervised and unsupervised learning. It utilizes a small amount of labeled data along with a large amount of unlabeled data to train models.

\subsubsection{Concept}

The core idea of semi-supervised learning~\cite{ouali2020overviewdeepsemisupervisedlearning} is to leverage the abundant unlabeled data to improve the performance of models trained on limited labeled data. This approach is particularly useful in scenarios where obtaining labeled data is expensive or time-consuming, but unlabeled data is plentiful.

\subsubsection{Working Principle}

\subsubsection{Semi-Supervised Learning Process}

\begin{enumerate}
    \item Data Preparation: Collect a small set of labeled data and a large set of unlabeled data.
    \item Initial Training: Train an initial model using the labeled data.
    \item Pseudo-labeling: Use the initial model to generate pseudo-labels for part of the unlabeled data.
    \item Model Update: Retrain the model using both the original labeled data and the pseudo-labeled data.
    \item Iteration: Repeat the pseudo-labeling and model update steps until convergence or a set number of iterations.
\end{enumerate}

\subsubsection{Common Algorithms}

\subsubsection{Semi-Supervised Learning Methods}

\begin{itemize}
    \item Self-training: The model uses its predictions to generate labels for unlabeled data~\cite{Yarowsky1995}.
    \item Co-training: Multiple models learn from each other, each focusing on different aspects of the data~\cite{Blum1998}.
    \item Generative Models: Use generative models to model the data distribution~\cite{Kingma2014}.
    \item Graph-based Methods: Construct graphs based on relationships between data points, then propagate labels on the graph~\cite{Zhu2002}.
    \item Semi-supervised SVMs: A variant of SVMs that incorporates unlabeled data~\cite{Joachims1999}.
\end{itemize}

\subsubsection{Application Scenarios}

1. Text Classification: Improving classification performance using large amounts of unlabeled text data~\cite{Nigam2000}.

2. Image Recognition: Training models with few labeled images and many unlabeled images~\cite{Rasmus2015}.

3. Speech Recognition: Enhancing recognition accuracy using large amounts of untranscribed speech data~\cite{Zhang2015}.

4. Bioinformatics: Predicting gene functions using few genes with known functions and many with unknown functions~\cite{Xia2014}.

5. Web Page Classification: Automatically classifying web pages using a few manually classified pages and many unclassified ones~\cite{Blum1998}.

\subsubsection{Advantages and Disadvantages}

\begingroup
\leftskip2em 

\paragraph{Advantages}
\begin{itemize}
    \item Can significantly reduce the need for labeled data.
    \item Often performs better than supervised learning using only labeled data.
    \item Can leverage large amounts of easily obtainable unlabeled data.
\end{itemize}

\paragraph{Disadvantages}
\begin{itemize}
    \item Algorithm design and implementation can be complex.
    \item May introduce incorrect labels, leading to performance degradation.
    \item Theoretical foundations are relatively weak, and effectiveness depends on specific problems and data distributions.
\end{itemize}

\endgroup

\subsubsection{Concrete Example}

Let's use a text classification problem to illustrate the application of semi-supervised learning:

Suppose we're building a news article classification system to categorize articles into four classes: "Politics", "Sports", "Technology", and "Entertainment". We have a small number of manually labeled articles and a large number of unlabeled articles.

1. Data Preparation:
   - Labeled data: 1,000 annotated news articles
   - Unlabeled data: 100,000 unannotated news articles

2. Feature Extraction:
   - Use TF-IDF (Term Frequency-Inverse Document Frequency) to convert text into numerical features
   - Features include: word frequency, article length, and presence of specific keywords

3. Initial Model Training:
   Train an initial Support Vector Machine (SVM) classifier using the 1,000 labeled articles

4. Self-training Process:
   a) Use the initial SVM to predict labels for the 100,000 unlabeled articles
   b) Select the 10,000 articles with the highest prediction confidence
   c) Add these high-confidence predictions to the training set
   d) Retrain the SVM using the augmented training set (now 11,000 articles)
   e) Repeat steps a-d for 5 iterations or until performance stabilizes

5. Model Evaluation:
   - Use 5-fold cross-validation to evaluate model performance
   - Compare the performance of the SVM trained only on labeled data vs. the semi-supervised model

6. Results Analysis:
   Suppose we observe:
   - Initial SVM (1,000 labeled articles): 75
   - Semi-supervised SVM (after 5 iterations): 85
   - "Politics" and "Sports" categories show the highest improvement
   - "Technology" and "Entertainment" still have some confusion

7. Practical Application:
   - Deploy the final model to classify incoming news articles
   - Implement a feedback loop where human editors occasionally verify and correct classifications, providing newly labeled data

8. Continuous Improvement:
   - Periodically retrain the model with newly acquired labeled data
   - Experiment with other semi-supervised methods like graph-based label propagation
   - Enhance feature extraction by incorporating word embeddings

This example demonstrates how semi-supervised learning can significantly improve classification performance by leveraging a large amount of unlabeled data, which is often readily available in real-world scenarios.

\subsection{Reinforcement Learning}

Reinforcement learning~\cite{mnih2013playingatarideepreinforcement} is a type of machine learning where an agent learns to make decisions by interacting with an environment. It mimics the way humans and animals learn through trial and error.

\subsubsection{Concept}

In reinforcement learning, an intelligent agent learns by taking actions in an environment and observing the results. Each action receives feedback from the environment, usually in the form of rewards or penalties. The agent's goal is to learn a policy that maximizes long-term cumulative rewards.

\subsubsection{Working Principle}

\subsubsection{Agent Interaction Process}

\begin{enumerate}
    \item State Perception: The agent observes the current state of the environment.
    \item Action Selection: Based on the current policy, the agent chooses an action.
    \item Action Execution: The chosen action is executed in the environment.
    \item Feedback Reception: The agent receives a reward or penalty from the environment.
    \item State Transition: The environment transitions to a new state.
    \item Policy Update: The action policy is updated based on the experience gained.
    \item Repetition: The above steps are repeated, continuously improving the policy.
\end{enumerate}

\subsubsection{Key Concepts}

\subsubsection{Key Concepts in Reinforcement Learning}

\begin{itemize}
    \item Markov Decision Process (MDP): A mathematical framework for describing reinforcement learning problems~\cite{Bellman1957}.
    \item Value Function: Estimates the expected future rewards from a given state~\cite{Sutton2018}.
    \item Policy: A rule that determines what action to take in a given state~\cite{Sutton1988}.
    \item Exploration vs. Exploitation: Balancing between trying new actions (exploration) and choosing known good actions (exploitation)~\cite{Kaelbling1996}.
    \item Temporal Difference Learning: Updating value estimates based on current estimates and observed rewards.
\end{itemize}

\subsubsection{Common Algorithms}

\subsubsection{Reinforcement Learning Algorithms}

\begin{itemize}
    \item Q-learning: A model-free temporal difference learning algorithm~\cite{Watkins1992}.
    \item SARSA: Similar to Q-learning but considers the actual next action taken~\cite{Rummery1994}.
    \item Policy Gradient Methods: Directly optimize the policy without using a value function~\cite{Sutton2000}.
    \item Actor-Critic Methods: Combine policy gradient methods with value function estimation~\cite{Konda2000}.
    \item Deep Q-Network (DQN): Combines Q-learning with deep learning~\cite{Mnih2015}.
    \item Proximal Policy Optimization (PPO): A popular policy optimization algorithm~\cite{Schulman2017}.
\end{itemize}

\subsubsection{Application Scenarios}

\begin{itemize}
    \item Game AI: Such as AlphaGo in the game of Go~\cite{Silver2016}.
    \item Robot Control: Learning complex motion and manipulation tasks~\cite{Levine2016}.
    \item Autonomous Driving: Learning to make decisions in various traffic situations~\cite{Kendall2019}.
    \item Resource Management: Such as energy optimization in data centers~\cite{Mao2016}.
    \item Recommendation Systems: Learning strategies to maximize long-term user satisfaction~\cite{Zhao2019}.
    \item Financial Trading: Developing automated trading strategies~\cite{Moody2001}.
\end{itemize}

\subsubsection{Advantages and Disadvantages}

\begingroup
\leftskip2em 

\paragraph{Advantages}
\begin{itemize}
    \item Can learn complex sequential decision-making problems.
    \item Does not require large amounts of labeled data.
    \item Can continuously learn and adapt to new environments.
\end{itemize}

\paragraph{Disadvantages}
\begin{itemize}
    \item The Training process can be unstable and requires a lot of trial and error.
    \item Slow convergence on some problems.
    \item Difficult to apply to problems with sparse or delayed reward signals.
\end{itemize}

\endgroup

This reformatted structure for semi-supervised learning and reinforcement learning sections now matches the layout of the supervised and unsupervised learning sections, providing a consistent format throughout the chapter.

\paragraph{Concrete Example:}

Let's illustrate reinforcement learning with an example of training an AI to play a simple maze game:

\paragraph{Game Setup:}
\begin{itemize}
    \item The maze is a 10x10 grid
    \item Starting point: upper left corner (0,0)
    \item Goal: lower right corner (9,9)
    \item Walls block certain paths
    \item The agent can move Up, Down, Left, or Right
\end{itemize}

\paragraph{Environment Definition:}
\begin{itemize}
    \item State: Agent's position (x, y coordinates)
    \item Actions: Up, Down, Left, Right
    \item Rewards:
    \begin{itemize}
        \item Reaching the goal: +100
        \item Hitting a wall: -10
        \item Each step: -1 (to encourage finding the shortest path)
    \end{itemize}
\end{itemize}

\paragraph{Q-table Initialization:}
Create a 100x4 table (100 states, 4 actions), initialize all values to 0.

\paragraph{Learning Parameters:}
\begin{itemize}
    \item Learning rate ($\alpha$): 0.1
    \item Discount factor ($\gamma$): 0.9
    \item $\epsilon$ for $\epsilon$-greedy strategy: initial value 0.9, decaying by 0.005 each episode
\end{itemize}

\paragraph{Training Process (Q-learning algorithm):}
\begin{itemize}
    \item For 1000 episodes:
    \begin{itemize}
        \item[(a)] Start at (0,0)
        \item[(b)] While not at (9,9):
        \begin{itemize}
            \item Choose action: $\epsilon$ probability of random action, else best Q-value action
            \item Execute action, observe new state, and reward
            \item Update Q-value: \[ Q(s,a) = Q(s,a) + \alpha \left[ r + \gamma \cdot \max(Q(s',a')) - Q(s,a) \right] \]
        \end{itemize}
        \item[(c)] Decrease $\epsilon$ by 0.005 (min 0.1)
    \end{itemize}
\end{itemize}

\paragraph{Evaluation:}
\begin{itemize}
    \item After training, run 100 test episodes without exploration
    \item Record success rate and average steps to goal
\end{itemize}

\paragraph{Results Analysis:}
\begin{itemize}
    \item Suppose we observe:
    \begin{itemize}
        \item First 100 episodes: 20\% success rate, average 80 steps when successful
        \item Last 100 episodes: 95\% success rate, average 20 steps
        \item Final test run: 98\% success rate, average 18 steps
    \end{itemize}
\end{itemize}

\paragraph{Visualization:}
\begin{itemize}
    \item Create a heatmap of the maze showing the most frequent path
    \item Plot learning curve (episode vs. steps to goal)
\end{itemize}

\paragraph{Further Improvements:}
\begin{itemize}
    \item Implement experience replay to improve sample efficiency
    \item Use a Deep Q-Network to handle larger, more complex mazes
    \item Add dynamic obstacles to test adaptability
\end{itemize}

\section{Key Concepts and Terminologies}

In this section, you will be introduced to the key concepts and terminologies commonly used in machine learning. Understanding these terms is crucial for anyone working in this field, as they form the foundation of both theoretical and applied ML practices.

\subsection{Features and Labels}
In machine learning, \textbf{features} refer to the input variables or attributes that are used by the model to make predictions. These can include any kind of data —numerical, categorical, or even textual — depending on the problem being solved. Features are often represented as vectors or matrices and are fed into the model to allow it to learn patterns from the data. 

\textbf{Labels}, on the other hand, represent the target variable or the outcome that the model is trying to predict. For supervised learning, the model uses labeled data (features paired with labels) to understand the relationship between the inputs and the desired output. For example, in an image classification task, the pixels of the image are the features, and the object category (e.g., "cat" or "dog") is the label.

\textbf{Feature Engineering} is the process of transforming raw data into meaningful features that can improve the performance of machine learning models. For instance, in natural language processing, raw text is transformed into numerical vectors, such as word embeddings, that can be fed into models.

\subsection{Training Data and Test Data}
To evaluate a machine learning model's performance accurately, the dataset is generally split into \textbf{training} and \textbf{test} sets. The \textbf{training data} is used to train the model, meaning that the model learns patterns from this subset by adjusting its internal parameters. The model must generalize well, i.e., it must not simply memorize the training data but learn to make accurate predictions on unseen data.

The \textbf{test data}, which the model has never seen during training, is used to evaluate its generalization capabilities. This split ensures that the model is assessed on data it has not encountered before, indicating how well it will perform on real-world, unseen datasets. Typically, an 80/20 or 70/30 split between training and test data is used, although cross-validation techniques can provide a more robust evaluation.

In addition to training and test data, a \textbf{validation set} is often used to tune hyperparameters and evaluate model performance during the training process. It helps prevent overfitting by providing feedback on how well the model generalizes before it is exposed to the test data.

\subsection{Model, Algorithm, and Hyperparameters}
A \textbf{model} in machine learning refers to the mathematical structure or system that learns to make predictions or decisions based on data. For example, a linear regression model might predict numerical outcomes based on a set of input features, while a convolutional neural network (CNN) might classify images.

An \textbf{algorithm} refers to the process or method used to train the model. Common algorithms include linear regression, decision trees, and gradient descent, among others. The algorithm defines how the model updates its internal parameters based on the input data.

\textbf{Hyperparameters} are the external parameters that are not learned from the training data but are set by the user. These include settings like learning rate, the number of layers in a neural network, or the number of trees in a random forest. Proper tuning of hyperparameters is critical to optimizing a model’s performance, and techniques such as grid search or Bayesian optimization are often used to identify the best hyperparameters for a given task.

After training, models are evaluated using metrics such as \textbf{accuracy}, \textbf{precision}, \textbf{recall}, and the \textbf{F1 score}, depending on the problem type. For regression tasks, metrics like \textbf{mean squared error} (MSE) or \textbf{R-squared} are commonly used. These metrics provide insight into how well the model is performing on unseen data.

\section{Machine Learning Process}

Machine learning follows a structured and iterative process, involving several key stages that guide the development and deployment of models. This section provides an overview of these stages, from data collection to model deployment.

\subsection{Data Collection}
\textbf{Data collection} is the first step in the machine learning process. It involves gathering relevant and high-quality data that will be used to train the model. The quality and quantity of the data directly influence the performance of the model, making this step critical. 

Sources of data can vary and include:
\begin{itemize}
    \item \textbf{Public datasets} (e.g., from Kaggle, UCI Machine Learning Repository)
    \item \textbf{APIs} (e.g., Twitter API for text data)
    \item \textbf{Custom collection} from sensors, logs, or manual entry.
\end{itemize}
Key considerations during data collection include ensuring the data is relevant, representative, and free of bias. In real-world projects, data collection is often one of the most time-consuming phases due to the need for data cleaning and formatting before use.

\subsection{Data Preprocessing}
Once the data is collected, it undergoes \textbf{preprocessing} to prepare it for use by the machine learning algorithm. Raw data is often incomplete, inconsistent, or noisy, making preprocessing a necessary step to ensure quality input for the model.

Common preprocessing techniques include:
\begin{itemize}
    \item \textbf{Data Cleaning}: Handling missing data (e.g., by imputing values), removing duplicates, and correcting inconsistencies.
    \item \textbf{Normalization and Standardization}: Scaling numerical features so that they have a uniform range or distribution, which improves the performance of many machine learning algorithms.
    \item \textbf{Encoding Categorical Data}: Converting categorical features into numerical values using techniques like one-hot encoding or label encoding.
    \item \textbf{Splitting the Data}: Dividing the dataset into training, validation, and test sets to ensure that the model generalizes well to unseen data.
\end{itemize}

\subsection{Model Selection}
Choosing the right \textbf{model} for a particular task depends on the nature of the problem and the data. This process involves selecting a suitable algorithm that can learn from the data and perform well in the target application.

Factors to consider when selecting a model include:
\begin{itemize}
    \item \textbf{Problem Type}: Classification, regression, clustering, etc. For instance, decision trees, logistic regression, and neural networks are common for classification tasks.
    \item \textbf{Model Complexity}: Simple models (like linear regression) are easier to interpret but may underfit complex data, while more complex models (like deep neural networks) are powerful but prone to overfitting.
    \item \textbf{Data Size}: Algorithms like k-nearest neighbors (KNN) can become computationally expensive with large datasets, while models like stochastic gradient descent (SGD) can scale well.
    \item \textbf{Training Time}: Some algorithms (e.g., support vector machines) take more time to train, so resource constraints must be considered.
\end{itemize}

\subsection{Training and Evaluation}
In the \textbf{training} phase, the model learns patterns from the training data by adjusting its internal parameters to minimize a predefined loss function. This process is typically iterative, involving multiple epochs and optimization algorithms (e.g., gradient descent) that aim to find the best parameters.

Key elements of this stage include:
\begin{itemize}
    \item \textbf{Hyperparameter Tuning}: Adjusting non-learnable parameters (e.g., learning rate, batch size) to improve model performance. This is often done using grid search or randomized search.
    \item \textbf{Cross-validation}: A technique to assess the model's performance and ensure that it generalizes well. K-fold cross-validation is commonly used, where the data is split into multiple folds, and the model is trained and tested on each fold.
    \item \textbf{Evaluation Metrics}: Different metrics are used to measure the model's effectiveness, depending on the problem. For classification tasks, accuracy, precision, recall, and F1-score are popular, while for regression tasks, mean squared error (MSE) or R-squared is typically used.
\end{itemize}

\subsection{Model Deployment and Monitoring}
After a model has been trained and evaluated, the final step is \textbf{deployment}, where the model is integrated into a production environment to make predictions on real-world data.

Key aspects of this phase include:
\begin{itemize}
    \item \textbf{Model Serving}: Setting up infrastructure to expose the model as a service, often through APIs. This allows other systems or applications to send new data to the model and receive predictions in real time.
    \item \textbf{Monitoring}: Once deployed, continuous monitoring of the model is critical to ensure that it maintains its accuracy over time. Model \textbf{drift} (where the model’s performance deteriorates due to changes in data patterns) is a common issue in production environments, requiring retraining or adjustments to the model.
    \item \textbf{Scalability}: Ensuring that the model can handle increasing amounts of data or more requests in real-time scenarios. Techniques like distributed computing or cloud-based deployment can help.
\end{itemize}

\section{Common Algorithms in Machine Learning}

\subsection{Linear Regression}
Linear regression is one of the most fundamental algorithms in machine learning~\cite{montgomery2012introduction}. It is primarily used for predictive analysis and finding relationships between variables. The equation for a simple linear regression model can be written as:
\[
y = \beta_0 + \beta_1 x + \epsilon
\]
Where \(y\) is the dependent variable, \(x\) is the independent variable, \(\beta_0\) is the intercept, \(\beta_1\) is the coefficient (slope), and \(\epsilon\) is the error term.

To implement linear regression in Python using PyTorch, you can follow this example:
\begin{lstlisting}[style=python]
import torch
import torch.nn as nn
import torch.optim as optim

# Sample data
x_train = torch.tensor([[1.0], [2.0], [3.0], [4.0]])
y_train = torch.tensor([[2.0], [4.0], [6.0], [8.0]])

# Model definition
class LinearRegressionModel(nn.Module):
    def __init__(self):
        super(LinearRegressionModel, self).__init__()
        self.linear = nn.Linear(1, 1)
    
    def forward(self, x):
        return self.linear(x)

model = LinearRegressionModel()

# Loss and optimizer
criterion = nn.MSELoss()
optimizer = optim.SGD(model.parameters(), lr=0.01)

# Training loop
for epoch in range(1000):
    model.train()
    optimizer.zero_grad()
    outputs = model(x_train)
    loss = criterion(outputs, y_train)
    loss.backward()
    optimizer.step()

    if epoch % 100 == 0:
        print(f'Epoch {epoch}, Loss: {loss.item()}')
\end{lstlisting}

Output:
\begin{lstlisting}[style=cmd]
Epoch 0, Loss: 44.164981842041016
Epoch 100, Loss: 0.10525545477867126
Epoch 200, Loss: 0.05778397619724274
Epoch 300, Loss: 0.03172262758016586
Epoch 400, Loss: 0.017415346577763557
Epoch 500, Loss: 0.009560829028487206
Epoch 600, Loss: 0.0052487668581306934
Epoch 700, Loss: 0.0028815295081585646
Epoch 800, Loss: 0.0015819233376532793
Epoch 900, Loss: 0.000868460105266422
\end{lstlisting}

This code defines a simple linear regression model using PyTorch, where we create a model, define a loss function, and use gradient descent to optimize the model’s weights.

\subsection{Logistic Regression}
Logistic regression is used for binary classification problems. Unlike linear regression, it models the probability that an observation belongs to a particular class by using the logistic function~\cite{Hastie2009}:
\[
P(y=1|x) = \frac{1}{1+e^{-(\beta_0 + \beta_1 x)}}
\]
In PyTorch, a logistic regression model can be implemented similarly to linear regression but using a sigmoid activation function for the output:
\begin{lstlisting}[style=python]
import torch.nn.functional as F

class LogisticRegressionModel(nn.Module):
    def __init__(self):
        super(LogisticRegressionModel, self).__init__()
        self.linear = nn.Linear(1, 1)

    def forward(self, x):
        return torch.sigmoid(self.linear(x))

model = LogisticRegressionModel()
\end{lstlisting}
Here, the logistic function \( \text{sigmoid} \) is used to map outputs to probabilities between 0 and 1, suitable for binary classification.

\subsection{Decision Trees and Random Forests}
Decision trees are non-parametric supervised learning algorithms used for classification and regression~\cite{breiman1984classification}. They split the data into subsets based on feature values to make predictions. Random forests are an ensemble method that combines multiple decision trees to improve prediction accuracy~\cite{breiman2001random}.

In PyTorch, decision trees and random forests are not natively supported, but libraries like Scikit-learn can be used to implement them:
\begin{lstlisting}[style=python]
from sklearn.ensemble import RandomForestClassifier

# Sample data
X_train = [[1, 2], [2, 3], [3, 4], [4, 5]]
y_train = [0, 1, 0, 1]

clf = RandomForestClassifier(n_estimators=10)
clf.fit(X_train, y_train)
print(clf.predict([[3, 3]]))
\end{lstlisting}

\subsection{Support Vector Machines}
Support Vector Machines (SVM)~\cite{cortes1995support} are supervised learning models used for classification and regression analysis. They work by finding a hyperplane that best separates the data into different classes.

Here is a simple example using Scikit-learn to implement an SVM:
\begin{lstlisting}[style=python]
from sklearn import svm

# Sample data
X_train = [[1, 2], [2, 3], [3, 4], [4, 5]]
y_train = [0, 1, 0, 1]

clf = svm.SVC(kernel='linear')
clf.fit(X_train, y_train)
print(clf.predict([[3, 3]]))
\end{lstlisting}

\subsection{Clustering Algorithms}
Clustering algorithms, such as K-means~\cite{arthur2007k}, are used for unsupervised learning where the goal is to group data into clusters based on similarity. K-means works by iteratively assigning data points to clusters and adjusting the cluster centroids.

An example of K-means clustering with Scikit-learn:
\begin{lstlisting}[style=python]
from sklearn.cluster import KMeans

X = [[1, 2], [2, 3], [3, 4], [4, 5]]

kmeans = KMeans(n_clusters=2, random_state=0)
kmeans.fit(X)
print(kmeans.predict([[3, 3]]))
\end{lstlisting}

\section{Challenges in Machine Learning}

\subsection{Overfitting and Underfitting}
Overfitting occurs when a machine learning model learns the details and noise in the training data to the extent that it negatively impacts the performance of new data. The model becomes too specific and fails to generalize. Underfitting, on the other hand, happens when the model is too simple to capture the underlying patterns of the data.

To mitigate overfitting, techniques such as cross-validation, regularization (L1, L2), and dropout (for neural networks) are commonly used. Here's an example of applying L2 regularization (also known as weight decay) in PyTorch:
\begin{lstlisting}[style=python]
import torch.optim as optim

# Defining the optimizer with weight decay (L2 regularization)
optimizer = optim.SGD(model.parameters(), lr=0.01, weight_decay=1e-5)
\end{lstlisting}
For underfitting, the solution usually involves increasing model complexity, adding more features, or training for longer periods.

\subsection{Bias-Variance Tradeoff}
The bias-variance tradeoff is a central challenge in machine learning. A model with high bias makes strong assumptions about the data and tends to underfit, leading to poor performance on both the training and testing data. In contrast, a model with high variance is too sensitive to the fluctuations in the training data, leading to overfitting.

The goal is to find the right balance between bias and variance, which can be achieved through model complexity tuning, regularization, and ensuring an appropriate amount of training data.

\subsection{Data Quality and Quantity Issues}
High-quality data is essential for machine learning models. Problems such as missing data, noisy data, and imbalanced datasets can significantly degrade the performance of a model. Similarly, having an insufficient quantity of data leads to difficulties in training models, often resulting in underfitting or unstable models.

Techniques such as data augmentation, synthetic data generation, and balancing datasets using methods like SMOTE (Synthetic Minority Over-sampling Technique) can help improve data quality and quantity.

Here’s an example of how to handle missing data using Pandas:
\begin{lstlisting}[style=python]
import pandas as pd

# Filling missing values with the column mean
df.fillna(df.mean(), inplace=True)
\end{lstlisting}

\subsection{Ethical Concerns and Bias in AI}
Ethical issues in AI arise when models inadvertently encode biases present in the training data. This bias can lead to unfair outcomes, particularly when the data reflects societal biases, such as racial or gender discrimination.

Addressing these biases requires careful data curation, transparency in model design, and fairness-aware algorithms that adjust for these biases. AI developers must be aware of these ethical concerns to ensure their models are fair and just.

\section{Machine Learning Tools and Libraries}

\subsection{Overview of Popular Libraries}
There are several popular libraries used for machine learning, each offering different capabilities:

\textbf{Scikit-learn:} Primarily for classical machine learning algorithms such as linear regression, decision trees, and clustering.

\textbf{TensorFlow:} A deep learning framework developed by Google that supports both CPU and GPU computation, widely used in production environments.

\textbf{PyTorch:} Developed by Facebook, PyTorch is a deep learning library that is widely used in research due to its dynamic computation graph, ease of use, and flexibility.

Here is how to import these libraries:
\begin{lstlisting}[style=python]
import sklearn
import tensorflow as tf
import torch
\end{lstlisting}

\subsection{Choosing the Right Tool for Your Task}
The choice of tool or library depends on the specific machine-learning task. For example:
- If you are working with traditional machine learning algorithms (like decision trees or clustering), Scikit-learn is typically the best choice due to its simplicity.
- For deep learning models, PyTorch or TensorFlow are preferred.
- TensorFlow is often chosen for large-scale production systems, while PyTorch is commonly used in research and development.

\subsection{Integration with Data Processing Tools}
Machine learning models often require substantial data processing before model training. Libraries such as Pandas and NumPy are essential for data manipulation and preprocessing.

Here's an example of using Pandas to load and preprocess a dataset:
\begin{lstlisting}[style=python]
import pandas as pd
import numpy as np

# Loading data
df = pd.read_csv('data.csv')

# Preprocessing: standardizing the data
df_standardized = (df - df.mean()) / df.std()
\end{lstlisting}

\section{Practical Examples and Case Studies}

\subsection{Example Projects Using Supervised Learning}
Supervised learning is commonly used in applications like spam detection, sentiment analysis, and image classification. A classic example is predicting housing prices based on features such as size, number of rooms, and location.

Here’s an example project where we use a simple linear regression model to predict housing prices:
\begin{lstlisting}[style=python]
# Assuming X_train, y_train contain the feature and label data
from sklearn.linear_model import LinearRegression

model = LinearRegression()
model.fit(X_train, y_train)
predictions = model.predict(X_test)
\end{lstlisting}

\subsection{Unsupervised Learning in Action}
Unsupervised learning is used in tasks where the goal is to uncover hidden patterns in data, such as customer segmentation and anomaly detection. One common algorithm is K-means clustering, where we cluster data points into a pre-specified number of groups.

Here’s how to apply K-means clustering to a dataset:
\begin{lstlisting}[style=python]
from sklearn.cluster import KMeans

kmeans = KMeans(n_clusters=3)
kmeans.fit(data)
labels = kmeans.predict(data)
\end{lstlisting}

\subsection{Real-world Applications of Reinforcement Learning}
Reinforcement learning is used in scenarios where an agent learns to take actions in an environment to maximize cumulative reward. It is commonly applied in areas like robotics, game-playing, and autonomous vehicles.

An example of reinforcement learning is Google’s AlphaGo, which uses this technique to play and defeat human players in the game of Go. Reinforcement learning models like Q-learning are implemented using PyTorch or TensorFlow, allowing the agent to learn optimal policies through trial and error.

Here’s the basic structure of a Q-learning algorithm:
\begin{lstlisting}[style=python]
import numpy as np

# Initialize Q-table with zeros
Q_table = np.zeros([state_size, action_size])

# Update rule for Q-learning
def update_Q(Q_table, state, action, reward, next_state, alpha, gamma):
    best_next_action = np.argmax(Q_table[next_state])
    Q_table[state, action] = Q_table[state, action] + alpha * (reward + gamma * Q_table[next_state, best_next_action] - Q_table[state, action])
\end{lstlisting}

\section{Summary and Future Directions}

\subsection{Recap of Key Points}
In this chapter, we covered the following important machine-learning concepts:

\begin{itemize}
    \item We explored common algorithms in machine learning, such as linear regression, logistic regression, decision trees, random forests, support vector machines, and clustering algorithms.
    \item We discussed the key challenges faced in machine learning, including overfitting, underfitting, bias-variance tradeoff, data quality, and ethical concerns related to bias in AI.
    \item We reviewed popular machine learning libraries such as Scikit-learn, TensorFlow, and PyTorch, emphasizing how to choose the right tool for specific tasks.
    \item Practical case studies in supervised, unsupervised, and reinforcement learning were examined to highlight real-world applications of these techniques.
\end{itemize}

Understanding these core topics prepares you for more advanced machine learning topics, where these foundational algorithms and tools will be applied in increasingly complex scenarios.

\subsection{Emerging Trends in Machine Learning}
Machine learning is a rapidly evolving field, and new trends are constantly emerging. Some of the key trends shaping the future of machine learning include:

\textbf{AutoML:} Automating machine learning processes has been a significant focus in recent years. AutoML~\cite{feurer2015efficient} tools allow non-experts to build machine learning models with minimal effort, automating tasks such as feature engineering, model selection, and hyperparameter tuning.

\textbf{Explainable AI (XAI):} With the increasing adoption of machine learning in critical areas like healthcare and finance, there is a growing need for models that are interpretable and transparent. XAI focuses on creating models that can explain their decisions in ways humans can understand~\cite{arrieta2020explainable}.

\textbf{Federated Learning:} Federated learning~\cite{mcmahan2017communication} is a technique that enables training models across decentralized devices without needing to centralize data. This approach enhances data privacy and security, making it suitable for industries like healthcare and finance.

\textbf{Deep Reinforcement Learning:} Reinforcement learning combined with deep learning techniques has been making headlines due to its success in areas such as robotics, game AI, and autonomous driving. This trend is expected to grow as more real-world applications are explored.

\textbf{AI Ethics and Fairness:} As machine learning becomes more embedded in society, there is a growing focus on ensuring that AI systems are fair, transparent, and ethical. Regulatory frameworks are being developed to ensure the responsible deployment of AI technologies~\cite{mittelstadt2016ethics}.

\subsection{Further Reading and Resources}
To deepen your understanding of machine learning, consider exploring the following books, articles, and online resources:

\textbf{Books:}
\begin{itemize}
    \item \textit{Pattern Recognition and Machine Learning} by Christopher M. Bishop — A comprehensive introduction to the concepts of machine learning, suitable for advanced learners.
    \item \textit{Deep Learning} by Ian Goodfellow, Yoshua Bengio, and Aaron Courville — An in-depth guide to deep learning, covering foundational concepts and modern applications.
\end{itemize}

\textbf{Articles:}
\begin{itemize}
    \item "A Few Useful Things to Know About Machine Learning" by Pedro Domingos — A well-known article that covers essential tips for practitioners in the field.
    \item "Deep Learning for AI" by Yann LeCun, Yoshua Bengio, and Geoffrey Hinton — A foundational paper on deep learning that outlines the concepts driving this field.
\end{itemize}

\textbf{Online Resources:}
\begin{itemize}
    \item \textbf{Coursera:} Machine Learning by Andrew Ng — A popular course that provides an excellent introduction to the field of machine learning.
    \item \textbf{Fast.ai:} Practical deep learning for coders — A hands-on course designed to teach deep learning with PyTorch.
    \item \textbf{Kaggle:} An online platform where you can participate in machine learning competitions, learn from other practitioners, and work on real-world datasets.
\end{itemize}

\chapter{Understanding Neural Networks}

\section{Introduction to Neural Networks}
Neural networks (NN), central to deep learning, are computational models inspired by the structure and functioning of the human brain. Each neural network consists of interconnected layers of artificial neurons, which process and transmit information. The foundation of neural networks lies in their ability to learn complex patterns from data, enabling them to perform tasks like image recognition, language translation, and decision-making with unprecedented accuracy.

Neural networks have become indispensable in modern machine learning because they can approximate any continuous function, making them highly versatile in solving both linear and nonlinear problems. They excel in domains with large amounts of data, where traditional algorithms might struggle to capture intricate relationships between features. For example, deep learning models, particularly those based on neural networks, have set new benchmarks in fields like speech recognition, autonomous driving, and medical diagnosis.

\section{The Perceptron}

\subsection{Mathematical Foundations of the Perceptron}

The perceptron, introduced by Frank Rosenblatt in 1958~\cite{rosenblatt1958perceptron}, is the simplest form of a neural network. Mathematically, it consists of an input vector $X = (x_1, x_2, \dots, x_n)$, corresponding weights $W = (w_1, w_2, \dots, w_n)$, a bias term $b$, and an activation function $\phi$. The output of the perceptron is computed as:

\[
y = \phi(W \cdot X + b),
\]

where $\phi$ is often a step function (Heaviside function) that outputs a binary decision based on whether the weighted sum is greater than a threshold.

This model can be visualized as a single neuron in which input data is passed through the weighted sum and bias, followed by an activation function. Despite its simplicity, the perceptron is capable of solving linearly separable problems, such as classifying points on opposite sides of a line in a 2D space.

\subsection{Limitations of the Perceptron}

The primary limitation of the perceptron is its inability to solve non-linearly separable problems, such as the XOR problem. The XOR operation, which outputs true only when inputs differ, cannot be represented by a single linear decision boundary. This limitation was famously demonstrated by Minsky and Papert in 1969~\cite{minsky1969perceptrons}, which led to a decline in neural network research until the advent of multi-layer perceptrons (MLPs) and backpropagation algorithms.

By introducing hidden layers between the input and output, the MLP overcomes this limitation. These hidden layers allow for the creation of more complex decision boundaries, making neural networks capable of solving more intricate problems.

\section{Multilayer Perceptrons (MLP)}

\subsection{Architecture of MLP}

The Multilayer Perceptron (MLP) is a class of feedforward artificial neural networks that consists of at least three layers: an input layer, one or more hidden layers, and an output layer. Each layer contains fully connected neurons. MLP learns by adjusting the weights of these connections during training.

The key advantage of MLPs over simpler models like single-layer perceptrons is their ability to model complex, non-linear relationships in data. The addition of hidden layers allows MLPs to capture intricate patterns through hierarchical feature extraction. Each hidden layer transforms the input data in ways that make it easier for the subsequent layer to classify or predict.

During the learning process, MLPs use backpropagation, an algorithm that computes the gradient of the loss function concerning each weight by employing the chain rule. This allows the network to iteratively update weights in the direction that minimizes the error, enhancing the model's accuracy and robustness.

\subsection{Activation Functions}
A critical aspect of the MLP's power lies within its activation functions, which introduce non-linearity into the NN. Without non-linear activation functions, the MLP would essentially behave like a linear regression model, rendering it incapable of solving complex tasks. Common activation functions include:

\textbf{Sigmoid Function}  
The sigmoid function~\cite{NARAYAN199769}, defined as \( \sigma(x) = \frac{1}{1 + e^{-x}} \), outputs values between 0 and 1, making it useful for binary classification tasks. However, sigmoid suffer from vanishing gradients, which can slow down learning in deeper networks. This makes it less commonly used in modern deep-learning architectures.

\textbf{Hyperbolic Tangent (Tanh)}  
The tanh function, defined as \( \tanh(x) = \frac{e^x - e^{-x}}{e^x + e^{-x}} \), produces outputs between -1 and 1. Like the sigmoid, tanh also faces vanishing gradient issues, though it tends to work better than sigmoid in practice as its outputs are zero-centered, allowing for a more balanced gradient flow.

\textbf{Rectified Linear Unit (ReLU)}  
ReLU~\cite{agarap2019deeplearningusingrectified}, defined as \( \text{ReLU}(x) = \max(0, x) \), has become the most popular activation function for hidden layers. It mitigates the vanishing gradient problem by maintaining gradients for positive values of \(x\), allowing for faster and more efficient training of deep networks. However, ReLU can suffer from "dead neurons" where gradients become zero for negative inputs, which makes variants like Leaky ReLU and Parametric ReLU useful in some scenarios.

\textbf{Softmax}  
For classification tasks involving multiple categories, \textit{the softmax function is often used in the output layer}. It converts the raw output of the network into a probability distribution, ensuring that the sum of all outputs equals one. This is particularly useful in multi-class classification problems~\cite{pearce2021understandingsoftmaxconfidenceuncertainty}.

Each activation function has its specific use case, and the choice of function depends on the nature of the problem being solved, the depth of the network, and the computational resources available. The selection of the activation function also affects the speed and stability of model convergence during training.

\section{Feedforward Neural Networks}
Feedforward Neural Networks (FNN) are the simplest form of artificial neural networks, where information moves in one direction—from the input layer through the hidden layers (if any) to the output layer. Unlike recurrent networks, there are no cycles or feedback loops in FNNs. This makes them ideal for tasks like classification, regression, and function approximation, where the goal is to map inputs to outputs in a structured, linear manner.

\subsection{Forward Propagation}
Forward propagation is the process by which data flows through the network during both training and inference. In this process, the input data is multiplied by the corresponding weights and then passed through an activation function. This happens layer by layer, with the information being transformed at each hidden layer until it reaches the output layer.

Mathematically, for each neuron \(i\) in a given layer \(l\), the output is computed as:
\[
z_i^{(l)} = \sum_{j} w_{ij}^{(l)} x_j^{(l-1)} + b_i^{(l)}
\]
where \(w_{ij}^{(l)}\) are the weights, \(x_j^{(l-1)}\) is the input from the previous layer, and \(b_i^{(l)}\) is the bias term for that neuron. This weighted sum \(z_i^{(l)}\) is then passed through an activation function \(f\), typically ReLU, sigmoid, or tanh, to introduce non-linearity:
\[
a_i^{(l)} = f(z_i^{(l)})
\]
This process repeats for every layer until the final output is produced at the output layer.

\subsection{Loss Functions and Objective Functions}
Loss functions are essential in quantifying how well a model’s predicted outputs match the actual targets. The loss function calculates the error between the predicted output and the true label, serving as the objective function to be minimized during training.

\noindent\textbf{Common Loss Functions}:

\textbf{Mean Squared Error (MSE)}: Used for regression tasks, it calculates the average squared difference between predicted and true values. Mathematically, it is expressed as:
  \[
  \text{MSE} = \frac{1}{N} \sum_{i=1}^{N} (y_i - \hat{y_i})^2
  \]
  where \(y_i\) is the actual value and \(\hat{y_i}\) is the predicted value.
  
\textbf{Cross-Entropy Loss}: Primarily used for classification tasks, cross-entropy measures the difference between the true probability distribution and the predicted one. For binary classification, it is defined as:
  \[
  \text{Cross-Entropy} = -\frac{1}{N} \sum_{i=1}^{N} [y_i \log(\hat{y_i}) + (1 - y_i)\log(1 - \hat{y_i})]
  \]
  Cross-entropy penalizes confident but incorrect predictions more heavily, making it effective for classification models.

The gradients of the loss function concerning the model’s parameters are calculated and used to update the weights via optimization algorithms like gradient descent or its variants. Minimizing the loss function over multiple iterations ensures that the model improves its predictions by learning the optimal set of weights.

\section{Backpropagation and Gradient Descent}

\subsection{Gradient Descent Algorithm}
Gradient Descent~\cite{zhang2019gradientdescentbasedoptimization} is an optimization technique used to minimize the cost function of a neural network by iteratively adjusting the model's parameters (weights and biases). The core idea is to move in the direction of the negative gradient of the cost function to find the global or local minimum. Mathematically, it updates each parameter \(w\) by:

\[
w := w - \eta \frac{\partial J(w)}{\partial w}
\]

where \( \eta \) is the learning rate, and \( \frac{\partial J(w)}{\partial w} \) is the gradient of the cost function concerning the weight \(w\). The learning rate determines the size of the step we take toward the minimum; if it's too large, the algorithm may overshoot, and if it's too small, convergence will be slow.

\subsection{Backpropagation Algorithm}
Backpropagation is the key algorithm used to compute the gradients needed for gradient descent in a neural network. It works by calculating the gradient of the loss function concerning each weight, and iteratively adjusting weights to minimize the error. This is achieved using the chain rule to propagate errors backward from the output layer to the input layer.

\noindent\textbf{Steps of Backpropagation}

\begin{itemize}
  \item \textbf{Forward Pass}: Input data is passed through the network layer by layer, computing activation until the output is produced. The loss function \(J(\hat{y}, y)\) is calculated based on the difference between the predicted output \(\hat{y}\) and the true output \(y\).
  \item \textbf{Backward Pass}: The partial derivatives of the loss function concerning the weights are calculated. Using the chain rule, we can compute the gradient of the cost function for each layer starting from the output:
   \[
   \frac{\partial J}{\partial w} = \frac{\partial J}{\partial \hat{y}} \cdot \frac{\partial \hat{y}}{\partial z} \cdot \frac{\partial z}{\partial w}
   \]
   where \(z\) is the weighted input to the neuron, and \(\hat{y}\) is the predicted output.
  \item \textbf{Weight Update}: The gradients are used to update the weights, moving them in the direction that minimizes the loss function:
   \[
   w' = w - \eta \frac{\partial J}{\partial w}
   \]
\end{itemize}

This process is repeated for multiple training epochs until the network converges to a minimal loss.

\subsection{Optimization Techniques (SGD, Adam, etc.)}
Several variations of the gradient descent algorithm have been developed to improve convergence and performance:

\noindent\textbf{Stochastic Gradient Descent (SGD)}

SGD updates the model parameters using a single training example at each iteration, rather than calculating the gradient over the entire dataset. This makes SGD much faster for large datasets but introduces more variance in updates, leading to a noisier convergence:
\[
w' = w - \eta \frac{\partial J(w)}{\partial w}(x^{(i)}, y^{(i)})
\]
where \(x^{(i)}, y^{(i)}\) represents one training example.

\noindent\textbf{Adam (Adaptive Moment Estimation)}

Adam~\cite{kingma2017adammethodstochasticoptimization} is a more advanced optimization algorithm that adapts the learning rate for each parameter. It combines the benefits of both AdaGrad (adaptive learning rates) and RMSProp (exponentially decaying average of past gradients). Adam maintains two moving averages of the gradient (first and second moments) to scale the learning rate:
\[
m_t = \beta_1 m_{t-1} + (1 - \beta_1) g_t
\]
\[
v_t = \beta_2 v_{t-1} + (1 - \beta_2) g_t^2
\]
where \(g_t\) is the gradient at time step \(t\), and \(\beta_1, \beta_2\) are decay rates. The adaptive learning rate makes Adam particularly effective for non-stationary and noisy objectives.

\section{Weight Initialization and Regularization}

\subsection{Importance of Weight Initialization}
Proper weight initialization is essential for efficient training of deep neural networks. If the weights are too large or too small, it can lead to issues such as vanishing or exploding gradients, which can severely slow down or even prevent model convergence. Two commonly used initialization techniques are:

\subsubsection{Xavier Initialization:} This method is used primarily for networks with sigmoid or tanh activation functions~\cite{glorot2010understanding}. It ensures that the variance of the outputs of each layer remains constant by drawing the initial weights from a distribution with variance:
\[
v^2 = \frac{1}{N}
\]
where \(N\) is the number of input neurons to the layer. This helps to prevent the gradients from becoming too small as they propagate through the network.

\subsubsection{He Initialization:} For ReLU and its variants, He initialization is more appropriate~\cite{he2015delvingdeeprectifierssurpassing}. It adjusts the variance of the weights to:
\[
v^2 = \frac{2}{N}
\]
where \(N\) is the number of inputs to the neuron. This helps maintain variance throughout the network, allowing for faster convergence, especially in deep networks.

\subsection{Overfitting and Regularization}
Overfitting occurs when a model performs well on the training data but poorly on unseen data. Regularization techniques help to reduce overfitting by penalizing overly complex models, encouraging the network to generalize better to new data.

\subsubsection{L1 and L2 Regularization}
\paragraph{L1 Regularization (Lasso):} L1 regularization~\cite{rs15061670} adds a penalty proportional to the absolute value of the weights:
\[
J = J_0 + \lambda \sum |w_i|
\]
This encourages sparsity in the network by driving some weights to zero, which can be useful for feature selection.

\paragraph{L2 Regularization (Ridge or Weight Decay):} L2 regularization~\cite{hoerl1970ridge} adds a penalty proportional to the square of the weights:
\[
J = J_0 + \lambda \sum w_i^2
\]
L2 regularization tends to shrink all weights but does not force them to zero, promoting smoother, more distributed weight values. It helps in stabilizing learning and preventing large swings in weights during training.

\subsubsection{Dropout}
Dropout~\cite{srivastava2014dropout} is a widely used regularization technique that works by randomly "dropping" a subset of neurons during each training iteration. Each neuron is kept active with a probability \(p\) (a hyperparameter typically between 0.5 and 0.8). This prevents the network from relying too heavily on any single neuron and forces it to learn more robust features. During inference, the full network is used, but the weights are scaled to account for the dropped neurons during training.

\section{Convolutional Neural Networks (CNNs)}

\subsection{Introduction to CNNs}
Convolutional Neural Networks (CNNs)~\cite{oshea2015introductionconvolutionalneuralnetworks} are a specialized class of neural networks designed to process data with a grid-like topology, such as images. CNNs have revolutionized fields like image recognition, computer vision, and video analysis due to their ability to automatically and efficiently extract spatial hierarchies of features. CNNs reduce the need for manual feature extraction by learning both low-level features like edges and high-level features like shapes during training.

CNNs differ from traditional feedforward neural networks in their architecture, utilizing convolutional layers instead of fully connected layers. This enables the network to be more parameter-efficient, making it better suited for high-dimensional inputs such as images, which can contain millions of pixels. CNNs are widely used in applications such as object detection, facial recognition, medical imaging, and even natural language processing when processing structured data.

\subsection{Convolutional Layers and Pooling}
The core component of CNNs is the convolutional layer. In a convolutional layer, small filters (or kernels) slide over the input data, performing element-wise multiplications and summing the results to produce feature maps. Each filter is responsible for detecting different features, such as edges, textures, or more complex patterns, depending on the layer's depth in the network.

Mathematically, for an input feature map \( I \) and a filter \( F \), the convolution operation is defined as:
\[
S(i,j) = (I * F)(i,j) = \sum_m \sum_n I(i+m,j+n) F(m,n)
\]
where \( S(i,j) \) is the result of the convolution at position \( (i,j) \), and \( m, n \) are the dimensions of the filter. 

Convolutional layers are often followed by non-linear activation functions like ReLU (Rectified Linear Unit), which introduce non-linearity into the model and allow it to learn more complex representations.

\subsubsection{Pooling Layers:} Pooling layers are another essential part of CNNs, typically used after convolutional layers to reduce the spatial dimensions of the feature maps, which helps decrease the number of parameters and computational complexity. The most common form of pooling is max pooling, where a filter selects the maximum value from a region of the feature map. Mathematically, max pooling for a 2x2 region can be described as:
\[
P(i,j) = \max \{ S(i,j), S(i+1,j), S(i,j+1), S(i+1,j+1) \}
\]
Pooling helps retain the most important information while discarding less relevant details, making the network more robust to minor translations and distortions in the input data.

\subsection{Applications of CNNs}
CNNs have found widespread applications due to their strong ability to process visual data:
\begin{itemize}
    \item \textbf{Image Classification:} CNNs are used to classify images, from basic object recognition to more complex tasks like identifying tumors in medical imaging~\cite{Krizhevsky2012}.
    \item \textbf{Object Detection:} Advanced CNN architectures such as Faster R-CNN~\cite{Ren2015} and YOLO (You Only Look Once)~\cite{Redmon2016} enable real-time object detection and localization in images and videos.
    \item \textbf{Facial Recognition:} CNNs are used in facial recognition systems, where they extract unique features from faces for identification and verification purposes~\cite{Taigman2014}.
    \item \textbf{Medical Imaging:} In the healthcare sector, CNNs are utilized to analyze medical images such as X-rays, MRIs, and CT scans to detect diseases and abnormalities~\cite{Shen2017}.
    \item \textbf{Natural Language Processing:} While CNNs are primarily used for image-related tasks, they are also applied in NLP for tasks like sentence classification by treating text as a structured grid~\cite{Kim2014}.
\end{itemize}

The combination of convolutional layers for feature extraction and pooling layers for dimensionality reduction makes CNNs highly effective for handling large-scale image data and other spatially structured data, making them a cornerstone of modern machine learning.

\section{Recurrent Neural Networks (RNNs) and Variants}

\subsection{Introduction to RNNs}
Recurrent Neural Networks (RNNs)~\cite{schmidt2019recurrentneuralnetworksrnns} are a class of neural networks designed to model sequential data by maintaining a memory of previous inputs. Unlike feedforward networks, RNNs allow information to persist, making them well-suited for tasks involving sequences, such as time series prediction, natural language processing, and speech recognition. The key feature of RNNs is their use of loops in the network, enabling the network to pass information from one step to the next.

Mathematically, the hidden state \( h_t \) at time step \( t \) is computed based on the input \( x_t \) and the hidden state from the previous time step \( h_{t-1} \):
\[
h_t = f(W_{xh} x_t + W_{hh} h_{t-1} + b_h)
\]
where \( W_{xh} \) and \( W_{hh} \) are weight matrices, \( b_h \) is the bias, and \( f \) is the activation function (typically tanh or ReLU).

While RNNs excel at capturing temporal dependencies, they struggle with long-term dependencies due to issues like vanishing and exploding gradients. This led to the development of more advanced architectures like LSTMs and GRUs.

\subsection{Long Short-Term Memory (LSTM)}
Long Short-Term Memory (LSTM) networks~\cite{hochreiter1997long} is a type of RNN specifically designed to handle long-term dependencies in sequential data. LSTMs mitigate the vanishing gradient problem by introducing a memory cell \( C_t \) that retains information across many time steps. Each LSTM unit contains three gates: the forget gate, input gate, and output gate, which control the flow of information into and out of the cell.

The equations governing an LSTM cell at time step \( t \) are as follows:
\[
f_t = \sigma(W_f \cdot [h_{t-1}, x_t] + b_f)
\]
\[
i_t = \sigma(W_i \cdot [h_{t-1}, x_t] + b_i)
\]
\[
\tilde{C}_t = \tanh(W_C \cdot [h_{t-1}, x_t] + b_C)
\]
\[
C_t = f_t \ast C_{t-1} + i_t \ast \tilde{C}_t
\]
\[
o_t = \sigma(W_o \cdot [h_{t-1}, x_t] + b_o)
\]
\[
h_t = o_t \ast \tanh(C_t)
\]
Here, \( f_t \), \( i_t \), and \( o_t \) represent the forget, input, and output gates, respectively, while \( C_t \) is the cell state. The LSTM architecture allows the network to retain information over long sequences, making it ideal for tasks like language modeling, machine translation, and video analysis.

\subsection{Gated Recurrent Units (GRU)}
Gated Recurrent Units (GRUs)~\cite{cho2014learning} is another variant of RNNs, similar to LSTMs but with a simplified architecture. GRUs combine the forget and input gates into a single update gate and merge the cell state with the hidden state. This results in fewer parameters and a faster training process compared to LSTMs, while still retaining the ability to capture long-term dependencies.

The key equations for a GRU are:
\[
z_t = \sigma(W_z \cdot [h_{t-1}, x_t] + b_z)
\]
\[
r_t = \sigma(W_r \cdot [h_{t-1}, x_t] + b_r)
\]
\[
\tilde{h}_t = \tanh(W_h \cdot [r_t \ast h_{t-1}, x_t] + b_h)
\]
\[
h_t = z_t \ast h_{t-1} + (1 - z_t) \ast \tilde{h}_t
\]
Here, \( z_t \) is the update gate and \( r_t \) is the reset gate. GRUs are particularly useful for time series analysis, speech recognition, and tasks where computational efficiency is important.

\subsection{Applications of RNNs, LSTMs, and GRUs}
RNNs and their variants are widely used across domains that require sequential data modeling. Common applications include:
\begin{itemize}
    \item \textbf{Natural Language Processing:} RNNs, LSTMs, and GRUs are used for tasks such as language translation, text generation, and sentiment analysis.
    \item \textbf{Time Series Prediction:} RNNs are applied to financial forecasting, weather prediction, and stock market analysis.
    \item \textbf{Speech Recognition:} LSTMs and GRUs are employed in speech-to-text systems to capture temporal dependencies in audio signals.
    \item \textbf{Video Analysis:} These architectures are used to analyze sequential frames in video data, enabling tasks like action recognition and video captioning.
\end{itemize}

\section{Advanced Neural Network Architectures}

\subsection{Autoencoders}
Autoencoders~\cite{bank2021autoencoders} are a type of unsupervised neural network architecture designed to learn efficient codings of input data. The network consists of two main parts: the encoder, which compresses the input data into a latent representation, and the decoder, which reconstructs the original data from this representation. Autoencoders are commonly used for tasks such as dimensionality reduction, anomaly detection, and data denoising.

The architecture of an autoencoder can be represented as follows:
\[
h = f(W_e \cdot x + b_e)
\]
\[
\hat{x} = g(W_d \cdot h + b_d)
\]
where \(x\) is the input, \(h\) is the encoded representation, and \(\hat{x}\) is the reconstructed input. The goal of training an autoencoder is to minimize the difference between \(x\) and \(\hat{x}\), typically using a loss function like mean squared error (MSE).

Variational Autoencoders (VAEs) are an extension of basic autoencoders that introduce a probabilistic approach to the latent space, allowing for more meaningful interpolation and generation of new data samples.

\subsection{Generative Adversarial Networks (GANs)}
Generative Adversarial Networks (GANs)~\cite{goodfellow2014generativeadversarialnetworks} consist of two neural networks that compete with each other in a game-like scenario: the generator and the discriminator. The generator creates synthetic data samples, while the discriminator tries to distinguish between real and generated data. The goal of the generator is to fool the discriminator by producing data that is indistinguishable from real data.

The generator learns by minimizing the following loss function:
\[
L_G = - \log(D(G(z)))
\]
where \(G(z)\) is the generator’s output given noise input \(z\), and \(D\) is the discriminator's prediction.

GANs are widely used in tasks such as image generation, video synthesis, and style transfer. Variants like Deep Convolutional GANs (DCGANs) and Conditional GANs (CGANs) have expanded their application scope, allowing for more controlled and realistic data generation.

\subsection{Transformers}
Transformers~\cite{vaswani2023attentionneed} are a type of neural network architecture designed to handle sequential data without relying on recurrent layers. Instead, they use a mechanism called self-attention, which allows the model to weigh the importance of different parts of the input sequence. This makes Transformers particularly effective for tasks involving long-range dependencies.

The self-attention mechanism can be expressed as:
\[
\text{Attention}(Q, K, V) = \text{softmax} \left( \frac{QK^T}{\sqrt{d_k}} \right) V
\]
where \(Q\) (queries), \(K\) (keys), and \(V\) (values) are the inputs, and \(d_k\) is the dimension of the key vectors.

Transformers are widely used in natural language processing tasks, particularly for sequence-to-sequence tasks such as language translation and text summarization. The architecture was popularized by models like BERT and GPT, which have set new benchmarks in NLP by leveraging massive pre-trained models and fine-tuning them for specific tasks.

\subsection{Applications of Advanced Neural Architectures}
\begin{itemize}
    \item \textbf{Autoencoders:} Used for anomaly detection, data compression, and generative tasks like image denoising and inpainting.
    \item \textbf{GANs:} Applied in image and video generation, deepfake technology, and creative AI systems for art generation.
    \item \textbf{Transformers:} Power state-of-the-art language models, machine translation systems, and question-answering tasks.
\end{itemize}

\section{Hyperparameter Tuning}

\subsection{Common Hyperparameters}
Hyperparameter tuning is a critical step in improving model performance and ensuring generalization to unseen data. Hyperparameters are settings that must be specified before the learning process begins, and they directly influence the behavior of the training process and the capacity of the model. Some of the most important hyperparameters include:

\subsubsection{Learning Rate:} The learning rate determines the size of the steps taken during gradient descent to minimize the loss function. A low learning rate results in slow convergence, while a high learning rate can cause the model to overshoot the optimal values, leading to divergence. Optimizing this hyperparameter is essential for training efficiency and stability.

\subsubsection{Batch Size:} Batch size defines how many samples are processed before the model's internal parameters (weights) are updated. Small batch sizes can introduce noise into gradient estimates but often lead to faster convergence. Larger batch sizes provide more stable gradients but require more memory and can lead to slower updates.

\subsubsection{Number of Layers and Neurons:} The architecture of a neural network, such as the number of layers and the number of neurons per layer, defines its capacity to learn complex patterns. More layers and neurons can capture higher-level abstractions but may also lead to overfitting if not controlled through regularization.

\subsubsection{Dropout Rate:} Dropout is a regularization technique where randomly selected neurons are ignored during training to prevent overfitting. The dropout rate specifies the probability of dropping a neuron in each layer. Adjusting this hyperparameter can significantly affect the generalization ability of the model.

\subsubsection{Weight Initialization:} Proper weight initialization can prevent issues such as vanishing and exploding gradients. Techniques like Xavier or He initialization are commonly used to ensure that the variance of activations is consistent across layers, allowing for smoother learning.

\subsubsection{Optimizer:} The choice of the optimizer (e.g., Stochastic Gradient Descent, Adam, RMSprop) controls how the model's parameters are updated based on the gradients. Some optimizers are better suited for specific tasks or architectures, and tuning their hyperparameters (e.g., momentum, beta parameters) can improve convergence.

\subsection{Grid Search, Random Search, and Bayesian Optimization}
There are several methods to optimize hyperparameters. The following techniques are commonly used to find the best hyperparameter configurations:

\subsubsection{Grid Search:} Grid search is an exhaustive method for hyperparameter tuning where all possible combinations of a predefined set of hyperparameter values are evaluated. While it guarantees finding the best combination, grid search can be computationally expensive, especially when dealing with multiple hyperparameters or large models. It works best when the search space is small or when computational resources are abundant.

\subsubsection{Random Search:} In random search, hyperparameters are randomly sampled from predefined distributions. Unlike grid search, random search does not systematically test all combinations, but it often finds good solutions faster because it covers a wider range of the hyperparameter space. Empirical studies have shown that random search can be more efficient than grid search for high-dimensional hyperparameter spaces.

\subsubsection{Bayesian Optimization:} Bayesian optimization~\cite{snoek2012practicalbayesianoptimizationmachine} models the relationship between hyperparameters and the model’s performance as a probabilistic function. It selects hyperparameter settings based on previous evaluations, balancing exploration (trying new regions of the hyperparameter space) with exploitation (focusing on promising areas). This approach is more efficient than grid or random search, particularly for complex models with many hyperparameters. Bayesian optimization is ideal when computational resources are limited or when training is time-consuming.

\subsubsection{Choosing an Optimization Method}
The choice of hyperparameter optimization method depends on the complexity of the model and the available computational resources:
\begin{itemize}
    \item \textbf{Grid Search:} Suitable for small models and when all hyperparameter values need to be tested.
    \item \textbf{Random Search:} More practical for larger search spaces where only a subset of hyperparameter combinations need to be evaluated.
    \item \textbf{Bayesian Optimization:} Best suited for models where each evaluation is computationally expensive, and an informed search is necessary to optimize quickly.
\end{itemize}

\section{Practical Considerations}

\subsection{Hardware Requirements for Training Neural Networks}
Training neural networks, particularly deep learning models, requires specialized hardware to handle the high computational demands. The primary options for hardware include:

\subsubsection{Graphics Processing Units (GPUs):} GPUs are widely used for training neural networks due to their ability to perform parallel computations efficiently. They are particularly effective for tasks involving large datasets and models such as Convolutional Neural Networks (CNNs) and Recurrent Neural Networks (RNNs). High-end GPUs, such as the NVIDIA A100, are ideal for large-scale deep-learning tasks, offering high throughput for both training and inference.

\subsubsection{Tensor Processing Units (TPUs):} TPUs, developed by Google~\cite{jouppi2017indatacenterperformanceanalysistensor}, are specialized hardware designed for machine learning tasks. TPUs are highly optimized for tensor operations and provide even faster performance than GPUs for certain tasks, such as training large transformer models and CNNs. TPUs are particularly suited for projects deployed on Google Cloud, where they can be scaled easily for large datasets and complex models.

\subsubsection{Memory and Storage:} Training large neural networks requires substantial memory resources. Models with millions or billions of parameters demand high VRAM (16GB to 40GB or more) to prevent memory bottlenecks. Fast storage solutions such as NVMe SSDs are critical for loading large datasets quickly, especially when performing data-intensive tasks like training on large image or text corpora.

\subsection{Training and Inference Speed Considerations}
Training and inference speed depend heavily on the hardware used and the optimization of the neural network architecture. Key considerations include:

\subsubsection{Batch Size and Learning Rate:} Batch size and learning rate are critical hyperparameters that influence training speed. Larger batch sizes reduce the variance in gradient updates but require more memory. The learning rate controls the speed of convergence, and optimizing this can prevent unnecessary training epochs.

\subsubsection{Model Parallelism and Data Parallelism:} For very large models, such as those used in NLP (e.g., GPT-3), it is common to use parallelism strategies. Model parallelism splits the model across multiple devices, while data parallelism distributes the training data across multiple devices. These techniques are essential for scaling models across multiple GPUs or TPUs to accelerate training.

\subsubsection{Inference Optimization:} For inference tasks, it is essential to minimize latency. Optimizations like pruning, quantization, and model compression reduce the size of the model, making inference faster and more efficient, especially when deploying models on edge devices or in production environments with limited resources.

\subsection{Model Deployment and Scalability}
Once a model is trained, deploying it effectively requires careful consideration of the deployment environment and scalability. Key aspects include:

\subsubsection{Scalability:} In large-scale applications, such as recommendation systems or real-time analytics, the ability to scale neural network models across multiple devices or cloud platforms is essential. TPUs, particularly in Google Cloud environments, offer seamless scalability for deep learning tasks. For GPU-based deployments, frameworks like NVIDIA’s TensorRT help optimize models for different GPUs, enhancing inference speed and performance.

\subsubsection{Deployment in Cloud Environments:} Cloud platforms like AWS, Google Cloud, and Azure provide flexible and scalable environments for deploying machine learning models. These platforms offer pre-configured instances with GPUs or TPUs, enabling rapid scaling as demand increases. Leveraging serverless functions and containerized services (e.g., Kubernetes) allows for dynamic scaling of machine learning services.

\subsubsection{Edge Deployment and Energy Efficiency:} For applications like autonomous vehicles and IoT devices, deploying models on edge devices is essential. Accelerators such as Edge TPUs and mobile GPUs offer efficient deployment solutions by optimizing models for lower power consumption and real-time processing.

\subsubsection{Monitoring and Maintenance:} After deployment, it is crucial to monitor model performance in production environments. Continuous integration with MLOps frameworks ensures models are retrained and updated as needed, maintaining high accuracy and efficiency over time.

\chapter{Getting Started with Deep Learning}

\section{AI Environment Setup}
In this section, we will go through the necessary steps to set up your environment for deep learning tasks. Whether you are using a CPU or a GPU, having the correct libraries and tools installed is essential for the smooth development and execution of machine learning models.

\subsection{Installing Deep Learning Libraries}
To get started with deep learning, you'll need to install the necessary Python libraries. The two most popular libraries for deep learning are \textbf{TensorFlow} and \textbf{PyTorch}. Here’s how you can install them using \texttt{pip}, Python’s package installer:

\textbf{Installing TensorFlow:}
\begin{lstlisting}[style=cmd]
pip install tensorflow
\end{lstlisting}

\textbf{Installing PyTorch:}
The installation command for PyTorch depends on your system configuration (CPU or GPU). You can specify these options on the official PyTorch website (\url{https://pytorch.org/}).

For example, to install PyTorch with CUDA (for NVIDIA GPU support):
\begin{lstlisting}[style=cmd]
pip install torch torchvision torchaudio --index-url https://download.pytorch.org/whl/cu118
\end{lstlisting}

\textbf{Setting up Virtual Environments:}
It’s highly recommended to create a virtual environment for your deep learning projects. This helps you manage dependencies and avoid conflicts between different projects. Here’s how to set up a virtual environment:

\begin{lstlisting}[style=cmd]
# Install virtualenv if not already installed
pip install virtualenv

# Create a virtual environment (replace 'env_name' with your chosen name)
virtualenv env_name

# Activate the virtual environment
# For Windows:
env_name\Scripts\activate
# For macOS/Linux:
source env_name/bin/activate

# Install deep learning libraries inside the virtual environment
pip install tensorflow
pip install torch
\end{lstlisting}

Remember to always activate the virtual environment before working on your project, and deactivate it after finishing your session by running:
\begin{lstlisting}[style=cmd]
deactivate
\end{lstlisting}

\subsection{GPU Setup for Deep Learning}
Deep learning models can benefit greatly from GPU acceleration, especially when working with large datasets and complex models. \textbf{NVIDIA GPUs} are commonly used for deep learning tasks because they support CUDA, a parallel computing platform, and cuDNN, a library for deep neural networks.

\textbf{Installing CUDA and cuDNN:}
To leverage GPU acceleration in TensorFlow or PyTorch, you’ll need to set up CUDA and cuDNN. Follow these steps:

\begin{enumerate}
    \item Download and install the NVIDIA drivers for your GPU from \url{https://www.nvidia.com/drivers}.
    \item Download CUDA from \url{https://developer.nvidia.com/cuda-downloads}. Select the version compatible with your GPU and operating system.
    \item Download cuDNN from \url{https://developer.nvidia.com/cudnn}. You’ll need to register for a free NVIDIA Developer account to access the download.
    \item After installing CUDA, copy the cuDNN files to the appropriate directories:
    \begin{lstlisting}[style=cmd]
    # Copy cuDNN files to the CUDA directory
    cp <cuDNN_path>/libcudnn* <CUDA_path>/lib64/
    cp <cuDNN_path>/include/cudnn*.h <CUDA_path>/include/
    \end{lstlisting}
\end{enumerate}

For a more detailed setup guide, refer to the official documentation:
\begin{itemize}
    \item \textbf{CUDA Installation Guide:} \url{https://docs.nvidia.com/cuda/}
    \item \textbf{cuDNN Installation Guide:} \url{https://docs.nvidia.com/deeplearning/cudnn/}
\end{itemize}

Once CUDA and cuDNN are installed, you can verify that PyTorch is using the GPU by running the following Python code:
\begin{lstlisting}[style=python]
import torch
if torch.cuda.is_available():
    print("CUDA is available. PyTorch is using GPU!")
else:
    print("CUDA is not available. PyTorch is using CPU.")
\end{lstlisting}

Similarly, for TensorFlow, you can verify GPU availability with:
\begin{lstlisting}[style=python]
import tensorflow as tf
print("Num GPUs Available: ", len(tf.config.experimental.list_physical_devices('GPU')))
\end{lstlisting}

\section{Introduction to TensorFlow and PyTorch}

\subsection{Overview of TensorFlow}
\textbf{TensorFlow} is an open-source deep learning framework developed by Google. It is widely used for both research and production environments, offering a flexible architecture for building and deploying machine learning models. The TensorFlow ecosystem includes tools like \texttt{TensorFlow Extended} (TFX) for deploying ML pipelines, \texttt{TensorFlow Lite} for mobile and embedded devices, and \texttt{TensorFlow.js} for running ML models in the browser.

One of the main features of TensorFlow is the use of computational graphs. This allows for optimization across a wide range of hardware devices, such as CPUs, GPUs, and TPUs.

\textbf{Basic Neural Network in TensorFlow:}
The following example demonstrates how to create a simple feedforward neural network using TensorFlow’s \texttt{Keras} API:
\begin{lstlisting}[style=python]
import tensorflow as tf
from tensorflow.keras import layers

# Define a simple neural network model
model = tf.keras.Sequential([
    layers.Dense(128, activation='relu', input_shape=(784,)),
    layers.Dense(64, activation='relu'),
    layers.Dense(10, activation='softmax')
])

# Compile the model
model.compile(optimizer='adam',
              loss='sparse_categorical_crossentropy',
              metrics=['accuracy'])

# Dummy data for demonstration
x_train = tf.random.normal((1000, 784))  # 1000 samples of 784 features
y_train = tf.random.uniform((1000,), maxval=10, dtype=tf.int32)  # 1000 labels

# Train the model
model.fit(x_train, y_train, epochs=5)
\end{lstlisting}
In this example, we define a simple fully connected neural network using two hidden layers with ReLU activation and an output layer with softmax activation for classification. The model is compiled with the Adam optimizer and trained on randomly generated data.

\subsection{Overview of PyTorch}
\textbf{PyTorch} is an open-source deep learning framework developed by Facebook’s AI Research lab. It has gained significant popularity in the research community due to its flexibility and dynamic computational graph, making it easier to debug and modify models on the go. PyTorch also integrates well with Python’s scientific libraries, such as NumPy and SciPy.

Unlike TensorFlow, which relies on static computational graphs, PyTorch’s dynamic graph (also known as define-by-run) allows computations to be defined as they happen, providing greater flexibility, especially for research and experimentation.

\textbf{Basic Neural Network in PyTorch:}
The following example demonstrates how to create a simple feedforward neural network using PyTorch:
\begin{lstlisting}[style=python]
import torch
import torch.nn as nn
import torch.optim as optim

# Define a simple neural network model
class SimpleNN(nn.Module):
    def __init__(self):
        super(SimpleNN, self).__init__()
        self.fc1 = nn.Linear(784, 128)
        self.fc2 = nn.Linear(128, 64)
        self.fc3 = nn.Linear(64, 10)

    def forward(self, x):
        x = torch.relu(self.fc1(x))
        x = torch.relu(self.fc2(x))
        x = torch.softmax(self.fc3(x), dim=1)
        return x

# Initialize the model, loss function, and optimizer
model = SimpleNN()
criterion = nn.CrossEntropyLoss()
optimizer = optim.Adam(model.parameters(), lr=0.001)

# Dummy data for demonstration
x_train = torch.randn(1000, 784)  # 1000 samples of 784 features
y_train = torch.randint(0, 10, (1000,))  # 1000 labels

# Training loop
for epoch in range(5):
    optimizer.zero_grad()
    outputs = model(x_train)
    loss = criterion(outputs, y_train)
    loss.backward()
    optimizer.step()

    if epoch % 1 == 0:
        print(f'Epoch {epoch+1}, Loss: {loss.item()}')
\end{lstlisting}
In this PyTorch example, we define a simple neural network with two hidden layers. The model is trained using stochastic gradient descent with the Adam optimizer, and the loss is computed using cross-entropy loss on randomly generated data.

\section{Building Your First Deep Learning Model}

In this section, we will guide you through building a simple deep-learning model, covering dataset preparation, defining the model architecture, and training it.

\subsection{Dataset Preparation}
Deep learning models require large amounts of data to train effectively. Some of the most common datasets used in the field are:
\begin{itemize}
    \item \textbf{MNIST:} A dataset of handwritten digits (28x28 grayscale images) used for classification tasks~\cite{LeCun1998}.
    \item \textbf{CIFAR-10:} A dataset containing 60,000 color images in 10 classes, with each image sized 32x32 pixels~\cite{Krizhevsky2009}.
\end{itemize}

In TensorFlow, datasets like MNIST can be loaded directly using the \texttt{tensorflow.keras.datasets} module:
\begin{lstlisting}[style=python]
import tensorflow as tf

# Load the MNIST dataset
(x_train, y_train), (x_test, y_test) = tf.keras.datasets.mnist.load_data()

# Normalize the pixel values between 0 and 1
x_train, x_test = x_train / 255.0, x_test / 255.0
\end{lstlisting}

Similarly, in PyTorch, datasets like CIFAR-10 can be loaded using \texttt{torchvision.datasets}:
\begin{lstlisting}[style=python]
import torchvision
import torchvision.transforms as transforms

# Transform to normalize the data
transform = transforms.Compose([
    transforms.ToTensor(),
    transforms.Normalize((0.5, 0.5, 0.5), (0.5, 0.5, 0.5))
])

# Load the CIFAR-10 dataset
trainset = torchvision.datasets.CIFAR10(root='./data', train=True,
                                        download=True, transform=transform)
trainloader = torch.utils.data.DataLoader(trainset, batch_size=32,
                                          shuffle=True)
\end{lstlisting}

These libraries provide convenient access to widely used datasets, allowing you to focus on model building.

\subsection{Defining the Model}
The next step is to define a simple feedforward neural network, also known as a fully connected network. In both TensorFlow and PyTorch, this involves stacking layers, specifying activation functions, and configuring the output units based on the task.

\textbf{TensorFlow: Defining a Feedforward Neural Network}
\begin{lstlisting}[style=python]
import tensorflow as tf
from tensorflow.keras import layers

# Define the model
model = tf.keras.Sequential([
    layers.Flatten(input_shape=(28, 28)),  # Flatten the input (for MNIST)
    layers.Dense(128, activation='relu'),  # Hidden layer
    layers.Dense(10, activation='softmax') # Output layer for 10 classes
])
\end{lstlisting}
Here, the input is flattened to a 1D vector, followed by a dense hidden layer with ReLU activation and an output layer with 10 units (for the 10 classes in MNIST), using softmax for classification.

\textbf{PyTorch: Defining a Feedforward Neural Network}
\begin{lstlisting}[style=python]
import torch.nn as nn
import torch.nn.functional as F

# Define the model
class SimpleNN(nn.Module):
    def __init__(self):
        super(SimpleNN, self).__init__()
        self.fc1 = nn.Linear(28 * 28, 128)  # Input layer (for MNIST)
        self.fc2 = nn.Linear(128, 10)       # Output layer for 10 classes

    def forward(self, x):
        x = x.view(-1, 28*28)  # Flatten the input
        x = F.relu(self.fc1(x))  # Hidden layer with ReLU
        x = F.log_softmax(self.fc2(x), dim=1)  # Output layer with softmax
        return x

model = SimpleNN()
\end{lstlisting}
In PyTorch, the input is also flattened, followed by a hidden layer with ReLU activation, and the output layer uses \texttt{log\_softmax} for classification.

\subsection{Compiling and Training the Model}
Once the model is defined, the next step is to compile it with an optimizer and loss function, and then train it.

\textbf{TensorFlow: Compiling and Training the Model}
\begin{lstlisting}[style=python]
# Compile the model
model.compile(optimizer='adam',
              loss='sparse_categorical_crossentropy',
              metrics=['accuracy'])

# Train the model
model.fit(x_train, y_train, epochs=5, batch_size=32)
\end{lstlisting}
In TensorFlow, we compile the model with the Adam optimizer and a cross-entropy loss function, which is suitable for classification tasks. The model is then trained using the \texttt{fit} method.

\textbf{PyTorch: Training the Model}
In PyTorch, the training process is more explicit, as we need to define the training loop manually:
\begin{lstlisting}[style=python]
import torch.optim as optim

# Define the optimizer and loss function
optimizer = optim.Adam(model.parameters(), lr=0.001)
criterion = nn.CrossEntropyLoss()

# Training loop
for epoch in range(5):
    for inputs, labels in trainloader:
        optimizer.zero_grad()   # Zero the gradients
        outputs = model(inputs) # Forward pass
        loss = criterion(outputs, labels) # Compute loss
        loss.backward()         # Backward pass
        optimizer.step()        # Update weights

    print(f'Epoch {epoch+1}, Loss: {loss.item()}')
\end{lstlisting}
In PyTorch, we define the optimizer and loss function manually, then write a loop to feed data through the model, compute the loss, backpropagate the gradients, and update the model weights.

Both frameworks allow us to effectively train deep learning models, with TensorFlow offering a more high-level API and PyTorch providing more granular control.

\section{Evaluating and Fine-Tuning Your Model}

After training a deep learning model, it's essential to evaluate its performance and fine-tune it to improve accuracy and generalization. In this section, we will cover how to evaluate the model, save and load models, and perform basic hyperparameter tuning.

\subsection{Evaluating Model Performance}
Evaluating a model involves assessing its performance on unseen test data. A key goal of this process is to determine whether the model can generalize well to new data. Several metrics are commonly used to evaluate model performance, including accuracy, loss, and confusion matrices.

\textbf{TensorFlow: Evaluating on Test Data}
In TensorFlow, evaluating a model on test data is straightforward:
\begin{lstlisting}[style=python]
# Evaluate the model on the test set
test_loss, test_acc = model.evaluate(x_test, y_test)
print(f'Test accuracy: {test_acc}')
\end{lstlisting}
The \texttt{evaluate} function returns the loss and accuracy on the test dataset.

\textbf{PyTorch: Evaluating on Test Data}
In PyTorch, we evaluate the model using the following loop:
\begin{lstlisting}[style=python]
correct = 0
total = 0
model.eval()  # Set the model to evaluation mode

with torch.no_grad():  # Disable gradient computation
    for data in testloader:
        inputs, labels = data
        outputs = model(inputs)
        _, predicted = torch.max(outputs.data, 1)
        total += labels.size(0)
        correct += (predicted == labels).sum().item()

print(f'Accuracy on test set: {100 * correct / total}%')
\end{lstlisting}
In this code, we loop over the test dataset, compute the predicted classes, and compare them to the true labels to compute accuracy.

\textbf{Confusion Matrices:}
For classification tasks, confusion matrices provide more detailed insight into model performance by showing how many samples were correctly or incorrectly classified for each class. A confusion matrix can be generated in Python using the \texttt{sklearn.metrics} library:
\begin{lstlisting}[style=python]
from sklearn.metrics import confusion_matrix
import seaborn as sns
import matplotlib.pyplot as plt

# True labels and predicted labels
y_true = [2, 0, 2, 2, 0, 1]
y_pred = [0, 0, 2, 2, 0, 2]

# Create confusion matrix
conf_matrix = confusion_matrix(y_true, y_pred)

# Plot the confusion matrix
sns.heatmap(conf_matrix, annot=True, cmap='Blues')
plt.show()
\end{lstlisting}

\subsection{Saving and Loading Models}
After training a model, it’s essential to save it for future use. This allows you to resume training later or use the model for inference without retraining from scratch.

\textbf{TensorFlow: Saving and Loading Models}
TensorFlow provides simple methods for saving and loading models:
\begin{lstlisting}[style=python]
# Save the model
model.save('my_model.h5')

# Load the model
loaded_model = tf.keras.models.load_model('my_model.h5')

# Use the loaded model to make predictions
predictions = loaded_model.predict(x_test)
\end{lstlisting}
In TensorFlow, models are saved in the HDF5 format by default and can be reloaded easily for further use.

\textbf{PyTorch: Saving and Loading Models}
In PyTorch, saving and loading models are slightly different, using \texttt{torch.save} and \texttt{torch.load}:
\begin{lstlisting}[style=python]
# Save the model
torch.save(model.state_dict(), 'model.pth')

# Load the model
model = SimpleNN()
model.load_state_dict(torch.load('model.pth'))
model.eval()  # Set the model to evaluation mode

# Use the loaded model for inference
outputs = model(inputs)
\end{lstlisting}
In PyTorch, we save the model’s state dictionary, which contains all the model parameters, and reload it when needed.

\subsection{Basic Hyperparameter Tuning}
Hyperparameters are values that define how a model is trained. Some of the most common hyperparameters include:
\begin{itemize}
    \item \textbf{Learning rate:} Determines how quickly the model’s weights are updated during training.
    \item \textbf{Batch size:} Defines the number of samples processed before updating the model’s weights.
    \item \textbf{Number of epochs:} Refers to how many times the model sees the entire training dataset.
\end{itemize}

Tuning these hyperparameters can significantly affect model performance.

\textbf{Learning Rate:}
A learning rate that is too high may cause the model to converge too quickly to a suboptimal solution, while a learning rate that is too low may result in slow convergence. In TensorFlow, the learning rate can be adjusted when defining the optimizer:
\begin{lstlisting}[style=python]
# Adjust the learning rate in the Adam optimizer
model.compile(optimizer=tf.keras.optimizers.Adam(learning_rate=0.0001),
              loss='sparse_categorical_crossentropy',
              metrics=['accuracy'])
\end{lstlisting}

In PyTorch, you can modify the learning rate when setting up the optimizer:
\begin{lstlisting}[style=python]
optimizer = optim.Adam(model.parameters(), lr=0.0001)
\end{lstlisting}

\textbf{Batch Size:}
Increasing the batch size can lead to faster training, but larger batches require more memory. A smaller batch size can result in more frequent updates and may help improve generalization but could increase training time.

In TensorFlow:
\begin{lstlisting}[style=python]
model.fit(x_train, y_train, epochs=5, batch_size=64)  # Batch size is 64
\end{lstlisting}

In PyTorch, the batch size is defined in the \texttt{DataLoader}:
\begin{lstlisting}[style=python]
trainloader = torch.utils.data.DataLoader(trainset, batch_size=64, shuffle=True)
\end{lstlisting}

\textbf{Number of Epochs:}
The number of epochs determines how many times the model will iterate over the entire training dataset. Increasing the number of epochs may improve model accuracy, but too many epochs can lead to overfitting.

You can experiment with these hyperparameters to find the best combination for your specific dataset and model.

\section{Implementing More Complex Models}

In this section, we will explore more advanced neural network architectures, such as Convolutional Neural Networks (CNNs) for image processing and Recurrent Neural Networks (RNNs) for sequential data tasks.

\subsection{Convolutional Neural Networks (CNNs)}
Convolutional Neural Networks (CNNs) are specialized neural networks designed for processing grid-like data such as images. CNNs are widely used for image classification, object detection, and other image-related tasks. They are designed to automatically capture spatial hierarchies in data through the use of convolutional layers, pooling layers, and fully connected layers.

\textbf{CNN Architecture:}
A typical CNN consists of:
\begin{itemize}
    \item \textbf{Convolutional Layers:} Apply convolution operations to extract features from the input data.
    \item \textbf{Pooling Layers:} Reduce the dimensionality of feature maps, retaining important features.
    \item \textbf{Fully Connected Layers:} Perform classification based on the extracted features.
\end{itemize}

\textbf{TensorFlow: Implementing a Simple CNN}
Here’s how you can build a simple CNN for image classification using TensorFlow:

\begin{lstlisting}[style=python]
import tensorflow as tf
from tensorflow.keras import layers, models

# Define the CNN model
model = models.Sequential([
    layers.Conv2D(32, (3, 3), activation='relu', input_shape=(32, 32, 3)),
    layers.MaxPooling2D((2, 2)),
    layers.Conv2D(64, (3, 3), activation='relu'),
    layers.MaxPooling2D((2, 2)),
    layers.Conv2D(64, (3, 3), activation='relu'),
    layers.Flatten(),
    layers.Dense(64, activation='relu'),
    layers.Dense(10, activation='softmax')  # 10 output classes
])

# Compile the model
model.compile(optimizer='adam',
              loss='sparse_categorical_crossentropy',
              metrics=['accuracy'])

# Print the model summary
model.summary()
\end{lstlisting}

In this example, the CNN consists of three convolutional layers with ReLU activation and two pooling layers to down-sample the feature maps. The network ends with fully connected layers for classification into 10 classes.

\textbf{PyTorch: Implementing a Simple CNN}
Here’s how to build the same CNN using PyTorch:

\begin{lstlisting}[style=python]
import torch
import torch.nn as nn
import torch.nn.functional as F

# Define the CNN model
class SimpleCNN(nn.Module):
    def __init__(self):
        super(SimpleCNN, self).__init__()
        self.conv1 = nn.Conv2d(3, 32, 3, padding=1)  # Input: 3 channels (RGB), Output: 32 filters
        self.pool = nn.MaxPool2d(2, 2)               # Max pooling with a 2x2 window
        self.conv2 = nn.Conv2d(32, 64, 3, padding=1)
        self.conv3 = nn.Conv2d(64, 64, 3, padding=1)
        self.fc1 = nn.Linear(64 * 8 * 8, 64)         # Fully connected layer
        self.fc2 = nn.Linear(64, 10)                 # Output layer for 10 classes

    def forward(self, x):
        x = self.pool(F.relu(self.conv1(x)))
        x = self.pool(F.relu(self.conv2(x)))
        x = self.pool(F.relu(self.conv3(x)))
        x = x.view(-1, 64 * 8 * 8)  # Flatten the tensor
        x = F.relu(self.fc1(x))
        x = F.log_softmax(self.fc2(x), dim=1)
        return x

# Create the model instance
model = SimpleCNN()
\end{lstlisting}

In this PyTorch implementation, we define a CNN with three convolutional layers followed by max pooling and then flatten the feature maps to feed into two fully connected layers for classification.

\subsection{Recurrent Neural Networks (RNNs)}
Recurrent Neural Networks (RNNs)~\cite{schmidt2019recurrentneuralnetworksrnns} are designed to handle sequential data by maintaining a hidden state that can capture information from previous time steps. They are commonly used for tasks such as time series forecasting, language modeling, and text generation. More advanced versions of RNNs, such as Long Short-Term Memory (LSTM) and Gated Recurrent Units (GRU), are capable of capturing long-term dependencies in sequences.

\textbf{RNN Architecture:}
A typical RNN consists of:
\begin{itemize}
    \item \textbf{Input Layers:} Accept sequential input data (e.g., time series, text sequences).
    \item \textbf{Recurrent Layers:} Process sequential data by maintaining a hidden state.
    \item \textbf{Output Layers:} Produce predictions based on the processed sequences.
\end{itemize}

\textbf{TensorFlow: Implementing an LSTM for Text Classification}
Here’s how you can implement an LSTM model for text classification using TensorFlow:

\begin{lstlisting}[style=python]
import tensorflow as tf
from tensorflow.keras import layers

# Define an LSTM model for text classification
model = tf.keras.Sequential([
    layers.Embedding(input_dim=10000, output_dim=64),  # Embedding layer for word vectors
    layers.LSTM(128),  # LSTM layer with 128 units
    layers.Dense(10, activation='softmax')  # Output layer for classification
])

# Compile the model
model.compile(optimizer='adam',
              loss='sparse_categorical_crossentropy',
              metrics=['accuracy'])

# Print the model summary
model.summary()
\end{lstlisting}

In this example, we use an embedding layer to convert words into vectors, followed by an LSTM layer to process the sequential data, and an output layer with 10 units for classification.

\textbf{PyTorch: Implementing an LSTM for Text Classification}
Here’s how you can implement an LSTM model using PyTorch:

\begin{lstlisting}[style=python]
import torch
import torch.nn as nn

# Define the LSTM model for text classification
class LSTMModel(nn.Module):
    def __init__(self):
        super(LSTMModel, self).__init__()
        self.embedding = nn.Embedding(10000, 64)  # Embedding layer
        self.lstm = nn.LSTM(64, 128, batch_first=True)  # LSTM layer
        self.fc = nn.Linear(128, 10)  # Fully connected output layer

    def forward(self, x):
        x = self.embedding(x)  # Convert words to vectors
        lstm_out, _ = self.lstm(x)  # Pass through LSTM
        x = lstm_out[:, -1, :]  # Get the output of the last time step
        x = self.fc(x)  # Pass through the fully connected layer
        return x

# Create the model instance
model = LSTMModel()
\end{lstlisting}

In this PyTorch implementation, we define an LSTM model with an embedding layer, an LSTM layer to process the sequences, and a fully connected layer for the classification task.

\section{Using Pre-trained Models and Transfer Learning}

Transfer learning is a powerful technique in deep learning that allows you to leverage pre-trained models on large datasets to solve new tasks with limited data. Instead of training a model from scratch, you can use an existing model trained on a large dataset (e.g., ImageNet) and fine-tune it for your specific task. This approach is particularly useful when working with small datasets or when training deep models that would otherwise take a long time to converge.

\subsection{Introduction to Transfer Learning}
\textbf{Transfer Learning} refers to the process of taking a pre-trained model and reusing it for a new, but related, task. In deep learning workflows, this approach is beneficial because:
\begin{itemize}
    \item It reduces the amount of training time and data required.
    \item Pre-trained models have already learned useful features from a large dataset, which can be repurposed for new tasks.
    \item It improves performance on tasks where you don’t have enough labeled data.
\end{itemize}

Transfer learning typically involves the following steps:
\begin{itemize}
    \item Load a pre-trained model that has been trained on a large dataset.
    \item Replace the final layer(s) to fit your specific task (e.g., change the number of output classes).
    \item Fine-tune the model on your custom dataset, either by training only the new layers or by fine-tuning the entire model.
\end{itemize}

\subsection{Using Pre-trained Models in TensorFlow and PyTorch}
Both TensorFlow and PyTorch provide pre-trained models \cite{chen2024deeplearningmachinelearning} through their respective libraries, such as \texttt{tensorflow.keras.applications} and \texttt{torchvision.models}.

\textbf{TensorFlow: Using a Pre-trained Model (VGG16)}
In TensorFlow, you can load pre-trained models from \texttt{tensorflow.keras.applications} and fine-tune them on your custom dataset. Here’s how to use the VGG16~\cite{simonyan2015deepconvolutionalnetworkslargescale} model pre-trained on ImageNet:

\begin{lstlisting}[style=python]
import tensorflow as tf
from tensorflow.keras.applications import VGG16
from tensorflow.keras import layers, models

# Load the pre-trained VGG16 model, exclude the top layer (for custom output)
base_model = VGG16(weights='imagenet', include_top=False, input_shape=(224, 224, 3))

# Freeze the base model layers
base_model.trainable = False

# Create a new model with custom top layers
model = models.Sequential([
    base_model,
    layers.Flatten(),
    layers.Dense(128, activation='relu'),
    layers.Dense(10, activation='softmax')  # 10 output classes for classification
])

# Compile the model
model.compile(optimizer='adam',
              loss='sparse_categorical_crossentropy',
              metrics=['accuracy'])

# Fine-tune the model on your custom dataset
# x_train and y_train should be your custom data
model.fit(x_train, y_train, epochs=5, batch_size=32)
\end{lstlisting}

In this example, we load the VGG16 model, freeze its pre-trained layers, and add custom layers to adapt it for a new classification task. You can also unfreeze the layers and fine-tune the entire model if needed.

\textbf{PyTorch: Using a Pre-trained Model (ResNet)}
Similarly, in PyTorch, you can load pre-trained models from \texttt{torchvision.models} and fine-tune them. Below is an example using the ResNet18 model:

\begin{lstlisting}[style=python]
import torch
import torch.nn as nn
import torchvision.models as models

# Load the pre-trained ResNet18 model
resnet = models.resnet18(pretrained=True)

# Freeze all layers of the base model
for param in resnet.parameters():
    param.requires_grad = False

# Replace the final fully connected layer with a new one for your task
num_ftrs = resnet.fc.in_features
resnet.fc = nn.Linear(num_ftrs, 10)  # 10 output classes

# Define the loss function and optimizer
criterion = nn.CrossEntropyLoss()
optimizer = torch.optim.Adam(resnet.fc.parameters(), lr=0.001)

# Train the model on your custom dataset
for epoch in range(5):
    for inputs, labels in trainloader:
        optimizer.zero_grad()
        outputs = resnet(inputs)
        loss = criterion(outputs, labels)
        loss.backward()
        optimizer.step()

    print(f'Epoch {epoch+1}, Loss: {loss.item()}')
\end{lstlisting}

In this example, we load the ResNet18 model, freeze the pre-trained layers, and replace the final fully connected layer with one suited for our custom task (e.g., 10 output classes). We then fine-tune the model on our dataset.

\subsection{Fine-tuning a Pre-trained Model on a Custom Dataset}
Fine-tuning a pre-trained model involves two primary strategies:
\begin{itemize}
    \item \textbf{Feature extraction:} Freeze the pre-trained model’s layers and only train the new layers you’ve added. This is useful when you don’t have much data or want to preserve the learned features from the base model.
    \item \textbf{Fine-tuning:} Unfreeze some or all of the layers of the pre-trained model and train them alongside the new layers. This allows the entire model to adapt to the new dataset.
\end{itemize}

Here’s how to fine-tune a model by unfreezing some layers (e.g., the top few convolutional layers) in TensorFlow:

\begin{lstlisting}[style=python]
# Unfreeze the top layers of the base model
for layer in base_model.layers[-4:]:
    layer.trainable = True

# Recompile the model with a lower learning rate for fine-tuning
model.compile(optimizer=tf.keras.optimizers.Adam(learning_rate=1e-5),
              loss='sparse_categorical_crossentropy',
              metrics=['accuracy'])

# Continue training (fine-tuning) the model
model.fit(x_train, y_train, epochs=10, batch_size=32)
\end{lstlisting}

In PyTorch, you can similarly unfreeze certain layers and adjust the optimizer to fine-tune the entire model.

\begin{lstlisting}[style=python]
# Unfreeze some layers of ResNet for fine-tuning
for name, param in resnet.named_parameters():
    if "layer4" in name:  # Unfreeze the last residual block (layer4)
        param.requires_grad = True

# Re-define the optimizer to include the unfrozen layers
optimizer = torch.optim.Adam(filter(lambda p: p.requires_grad, resnet.parameters()), lr=1e-5)

# Fine-tune the model
for epoch in range(10):
    for inputs, labels in trainloader:
        optimizer.zero_grad()
        outputs = resnet(inputs)
        loss = criterion(outputs, labels)
        loss.backward()
        optimizer.step()

    print(f'Epoch {epoch+1}, Loss: {loss.item()}')
\end{lstlisting}

Fine-tuning allows the model to better adapt to your custom dataset while still benefiting from the features learned in the pre-training phase.

\section{Practical Considerations}

When moving from research to production, there are several important considerations, including how to deploy models effectively and how to optimize them for speed and performance. This section will cover common deployment techniques and methods to enhance model efficiency.

\subsection{Model Deployment}
Deploying a deep learning model involves making the model accessible for inference in a production environment, whether it's on a server, in the cloud, or on edge devices.

\textbf{Deployment Techniques:}
There are various tools and techniques for deploying machine learning models. Two popular methods are:

\textbf{TensorFlow Serving:} TensorFlow Serving is a flexible, high-performance system for serving machine learning models in production environments. It provides an easy way to serve TensorFlow models and is compatible with TensorFlow Extended (TFX) pipelines.
\begin{lstlisting}[style=cmd]
# Install TensorFlow Serving
sudo apt-get install tensorflow-model-server

# Serve a model
tensorflow_model_server --rest_api_port=8501 --model_name=my_model --model_base_path=/path/to/model/
\end{lstlisting}

\textbf{TorchScript:} For PyTorch models, TorchScript is a way to serialize and optimize deployment models. It converts PyTorch models into a format that can be run in a high-performance environment, independent of Python.
\begin{lstlisting}[style=python]
import torch

# Convert a PyTorch model to TorchScript
scripted_model = torch.jit.script(model)

# Save the scripted model
scripted_model.save("model.pt")

# Load and run the model in a production environment
loaded_model = torch.jit.load("model.pt")
loaded_model(input_tensor)
\end{lstlisting}

\textbf{Cloud Services for Model Deployment:}
Cloud platforms like AWS, Google Cloud, and Azure provide robust services for deploying and scaling machine learning models.

\begin{itemize}
    \item \textbf{AWS Sagemaker:} AWS Sagemaker~\cite{aws_sagemaker} allows you to train and deploy machine learning models at scale. It simplifies the process of building, training, and deploying models in the cloud.
    \item \textbf{Google Cloud AI Platform:} Google Cloud offers AI Platform~\cite{google_ai_platform}, which integrates with TensorFlow and other ML frameworks to deploy and manage models at scale.
    \item \textbf{Azure Machine Learning:} Azure~\cite{azure_machine_learning} provides services for training, deploying, and managing machine learning models, including support for PyTorch, TensorFlow, and Scikit-learn.
\end{itemize}

\subsection{Optimizing for Speed and Performance}
Optimizing deep learning models is crucial when deploying them in production environments where speed and resource efficiency are important. Several techniques can be used to reduce model size and improve inference speed without significantly sacrificing accuracy.

\textbf{Quantization:}
Quantization reduces the precision of the numbers used to represent model weights and activations, typically from 32-bit floating-point to 8-bit integers. This can significantly reduce the model size and improve inference speed, especially on resource-constrained devices like mobile phones or edge devices.

\textbf{TensorFlow Quantization:}
TensorFlow supports post-training quantization, where you can quantize a trained model:
\begin{lstlisting}[style=python]
import tensorflow as tf

# Load the trained model
model = tf.keras.models.load_model('my_model.h5')

# Convert the model to a TensorFlow Lite model with quantization
converter = tf.lite.TFLiteConverter.from_keras_model(model)
converter.optimizations = [tf.lite.Optimize.DEFAULT]
quantized_model = converter.convert()

# Save the quantized model
with open('quantized_model.tflite', 'wb') as f:
    f.write(quantized_model)
\end{lstlisting}

\textbf{PyTorch Quantization:}
In PyTorch, you can apply dynamic quantization, which quantizes the model weights to 8-bit integers:
\begin{lstlisting}[style=python]
import torch.quantization

# Apply dynamic quantization
quantized_model = torch.quantization.quantize_dynamic(
    model, {torch.nn.Linear}, dtype=torch.qint8
)

# Save the quantized model
torch.jit.save(torch.jit.script(quantized_model), "quantized_model.pt")
\end{lstlisting}

\textbf{Model Pruning:}
Model pruning involves removing weights or entire neurons from the network that have little contribution to the model's output. This reduces the number of parameters, making the model smaller and faster without significantly impacting accuracy.

\textbf{TensorFlow Pruning:}
TensorFlow offers pruning capabilities through the TensorFlow Model Optimization Toolkit:
\begin{lstlisting}[style=python]
import tensorflow_model_optimization as tfmot

# Apply pruning to the model
prune_low_magnitude = tfmot.sparsity.keras.prune_low_magnitude
pruned_model = prune_low_magnitude(model)

# Compile and train the pruned model
pruned_model.compile(optimizer='adam', loss='sparse_categorical_crossentropy')
pruned_model.fit(x_train, y_train, epochs=5)
\end{lstlisting}

\textbf{PyTorch Pruning:}
In PyTorch, you can prune certain layers or entire networks:
\begin{lstlisting}[style=python]
import torch.nn.utils.prune as prune

# Prune 50% of the connections in the first linear layer
prune.l1_unstructured(model.fc1, name='weight', amount=0.5)

# Remove the pruning reparametrization to make it permanent
prune.remove(model.fc1, 'weight')
\end{lstlisting}

Optimizing models through techniques like quantization and pruning can greatly enhance performance, particularly when deploying models to edge devices or cloud environments with limited computational resources.

\chapter{Data Visualization}

Data visualization plays a critical role in data analysis and scientific research. By transforming complex data into intuitive graphs and charts, data visualization helps us easily identify trends, patterns, and anomalies, thereby enhancing the accuracy of decision-making. It is not only an effective way to present the results of data but also a vital tool for data exploration. Data visualization allows complex datasets to be displayed concisely and understandably, enabling audiences to quickly grasp the core information. Whether used in business analysis, academic research, or everyday decision support, data visualization is an indispensable tool. Through data visualization, we can more clearly understand the story behind the data and effectively share these insights with others.

\section{The Importance of Data Visualization}

In a particular grade, we have the height data of 200 students listed below. When looking at these numbers arranged in a table, it’s challenging to discern any meaningful patterns or trends due to the sheer volume and scattered distribution of the data. The overall distribution, central tendencies, and potential outliers are all hidden within this dense set of numbers, making it difficult to interpret by eye. In such cases, using visualization tools like histograms can help us better understand the data distribution, allowing us to see the concentration, range, and shape of the data more clearly, thus uncovering any underlying patterns.

\begin{table}[H]
    \centering
    \renewcommand{\arraystretch}{1} 
    \setlength{\tabcolsep}{3pt} 
    \begin{tabular}{|*{20}{c|}} 
        \hline
        174 & 168 & 176 & 185 & 167 & 167 & 185 & 177 & 165 & 175 & 165 & 165 & 172 & 150 & 152 & 164 & 159 & 173 & 160 & 155 \\ \hline
        184 & 167 & 170 & 155 & 164 & 171 & 158 & 173 & 163 & 167 & 163 & 188 & 169 & 159 & 178 & 157 & 172 & 150 & 156 & 171 \\ \hline
        177 & 171 & 168 & 166 & 155 & 162 & 165 & 180 & 173 & 152 & 173 & 166 & 163 & 176 & 180 & 179 & 161 & 166 & 173 & 179 \\ \hline
        165 & 168 & 158 & 158 & 178 & 183 & 169 & 180 & 173 & 163 & 173 & 185 & 169 & 185 & 143 & 178 & 170 & 167 & 170 & 150 \\ \hline
        167 & 173 & 184 & 164 & 161 & 164 & 179 & 173 & 164 & 175 & 170 & 179 & 162 & 166 & 166 & 155 & 172 & 172 & 170 & 167 \\ \hline
        155 & 165 & 166 & 161 & 168 & 174 & 188 & 171 & 172 & 169 & 150 & 169 & 170 & 194 & 168 & 173 & 169 & 158 & 181 & 177 \\ \hline
        177 & 160 & 184 & 155 & 175 & 191 & 160 & 164 & 170 & 164 & 154 & 170 & 159 & 174 & 160 & 185 & 162 & 166 & 178 & 157 \\ \hline
        172 & 183 & 153 & 171 & 172 & 177 & 157 & 156 & 175 & 172 & 172 & 173 & 163 & 172 & 172 & 162 & 188 & 174 & 158 & 176 \\ \hline
        160 & 177 & 181 & 161 & 179 & 174 & 178 & 188 & 167 & 162 & 161 & 161 & 169 & 173 & 172 & 178 & 170 & 184 & 167 & 197 \\ \hline
        176 & 161 & 159 & 174 & 167 & 177 & 174 & 169 & 161 & 154 & 165 & 178 & 172 & 157 & 171 & 173 & 161 & 171 & 170 & 158 \\ \hline
    \end{tabular}
    \caption{Heights of 200 Students (in cm)}
\end{table}

\begin{figure}[H]
    \centering
    \includegraphics[width=1.0\textwidth]{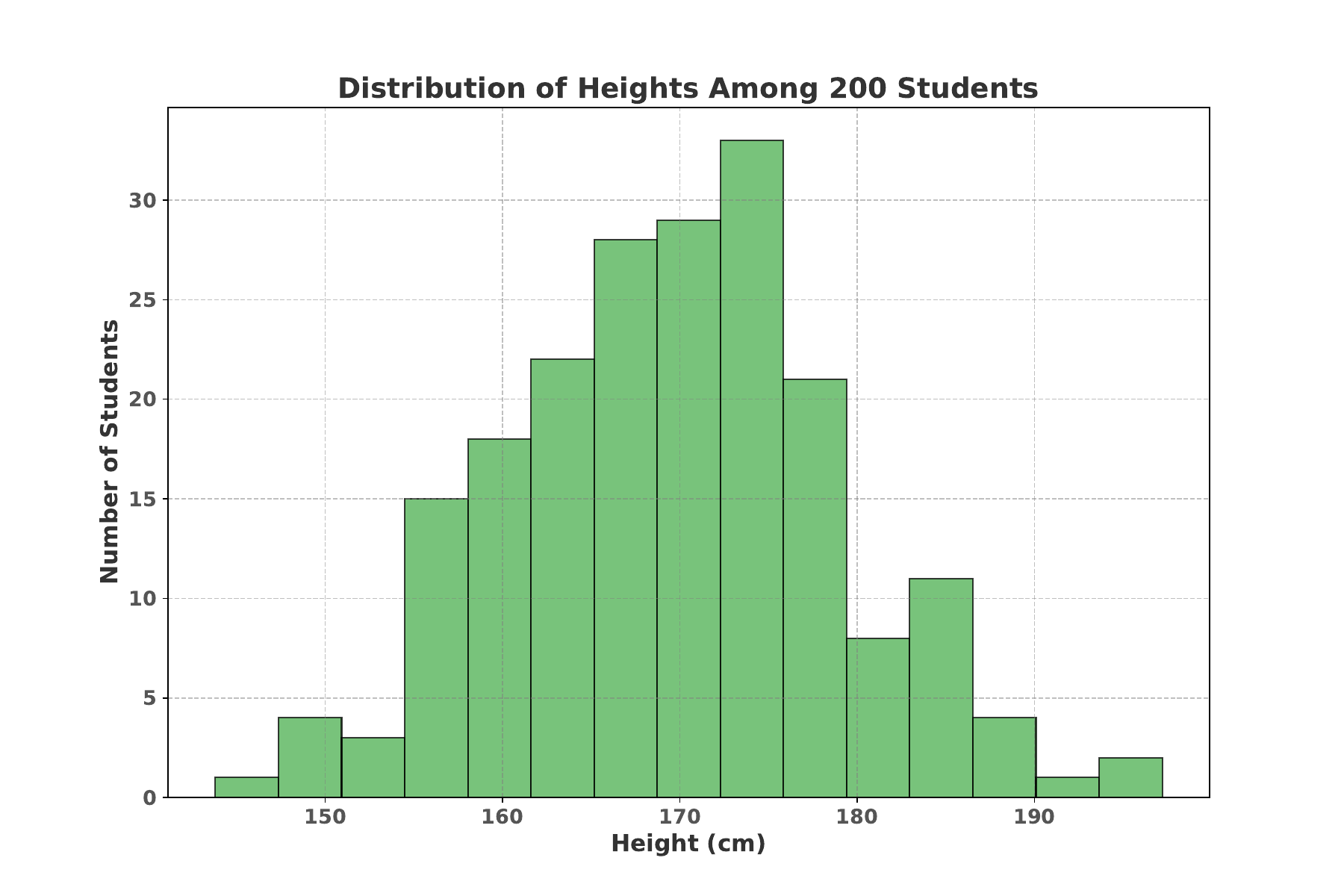}
    \caption{Histogram of Student Heights}
\end{figure}

This chart shows the distribution of heights among 200 students. As illustrated, the majority of heights are concentrated between 160 cm and 180 cm, with a clear peak around 170 cm, forming an evident normal distribution. Visualization like this allows us to easily observe the distribution, identify the central tendencies, and spot any anomalies. If we were to rely solely on raw data, such patterns would be much harder to detect. Thus, data visualization is crucial for understanding and analyzing large datasets; it simplifies complex information and helps us make quick, accurate decisions.

The code presented in this section is designed to generate and visualize a distribution of student heights using Python. It begins by importing the necessary libraries, setting a random seed for reproducibility, and generating a dataset of heights following a normal distribution. The heights are then displayed in a formatted manner with 20 numbers per row for better readability. Finally, the code creates a histogram with improved aesthetics, including custom colors, labels, and grid settings, to visually represent the distribution of heights among 200 students.

\begin{lstlisting}[style=python]
import numpy as np
import matplotlib.pyplot as plt


# Set a random seed for reproducibility
np.random.seed(42)

# Generate random heights following a normal distribution
heights = np.random.normal(loc=170, scale=10, size=200)

# Print the data in a format of 20 numbers per row
for i in range(0, len(heights), 20):
    for j in range(20):
        print(int(heights[i + j]), end=" ")
    print()  # New line after each row

# Plot a histogram with improved aesthetics
plt.figure(figsize=(12, 8))  # Increase the figure size
plt.hist(heights, bins=15, color='#4CAF50', edgecolor='black', alpha=0.75)  # Use a pleasing green color

# Add title and labels with improved font settings
plt.title('Distribution of Heights Among 200 Students', fontsize=18, fontweight='bold', color='#333333')
plt.xlabel('Height (cm)', fontsize=15, fontweight='bold', color='#333333')
plt.ylabel('Number of Students', fontsize=15, fontweight='bold', color='#333333')

# Customize grid and ticks for better readability
plt.grid(True, linestyle='--', alpha=0.5, color='gray')
plt.xticks(fontsize=13, fontweight='bold', color='#555555')
plt.yticks(fontsize=13, fontweight='bold', color='#555555')

# Display the histogram
plt.show()
\end{lstlisting}

\section{Commonly used Visualization Charts}

\subsection{Line Chart}

A Line Chart is a fundamental type of statistical chart used to display data points connected by straight line segments, showing trends over a continuous variable, often time. It is commonly used in data visualization to observe the behavior of variables over time or another continuous scale.

In this section, we will draw a simple line chart representing a quadratic function. The function is of the form \( y = ax^2 + bx + c \). For this example, we will set \( a = 1 \), \( b = 0 \), and \( c = 0 \), resulting in \( y = x^2 \). We will plot the values of this function for \( x \) ranging from -10 to 10.

\begin{lstlisting}[style=python]
import matplotlib.pyplot as plt
import numpy as np

# Define the range for x
x = np.linspace(-10, 10, 100)
# Define the quadratic function y = x^2
y = x**2

# Plotting the line chart
plt.figure(figsize=(10, 6))
plt.plot(x, y, marker='o')
plt.title('Line Chart of Quadratic Function $y = x^2$')
plt.xlabel('x')
plt.ylabel('y')
plt.grid(True)
plt.show()
\end{lstlisting}

In the above code, the \texttt{x} array represents values ranging from -10 to 10, and the corresponding \texttt{y} array represents the values calculated using the quadratic function \( y = x^2 \). We use the \texttt{plt.plot()} function to draw a line chart, with markers (\texttt{marker='o'}) highlighting each calculated point.

The chart's title is "Line Chart of Quadratic Function \( y = x^2 \)", with the x-axis labeled "x" and the y-axis labeled "y". The grid is enabled using \texttt{plt.grid(True)} to help better visualize the position of each data point.

Below is the line chart illustrating the quadratic function \( y = x^2 \) over the range -10 to 10:

\begin{center}[H]
\includegraphics[width=1.0\textwidth]{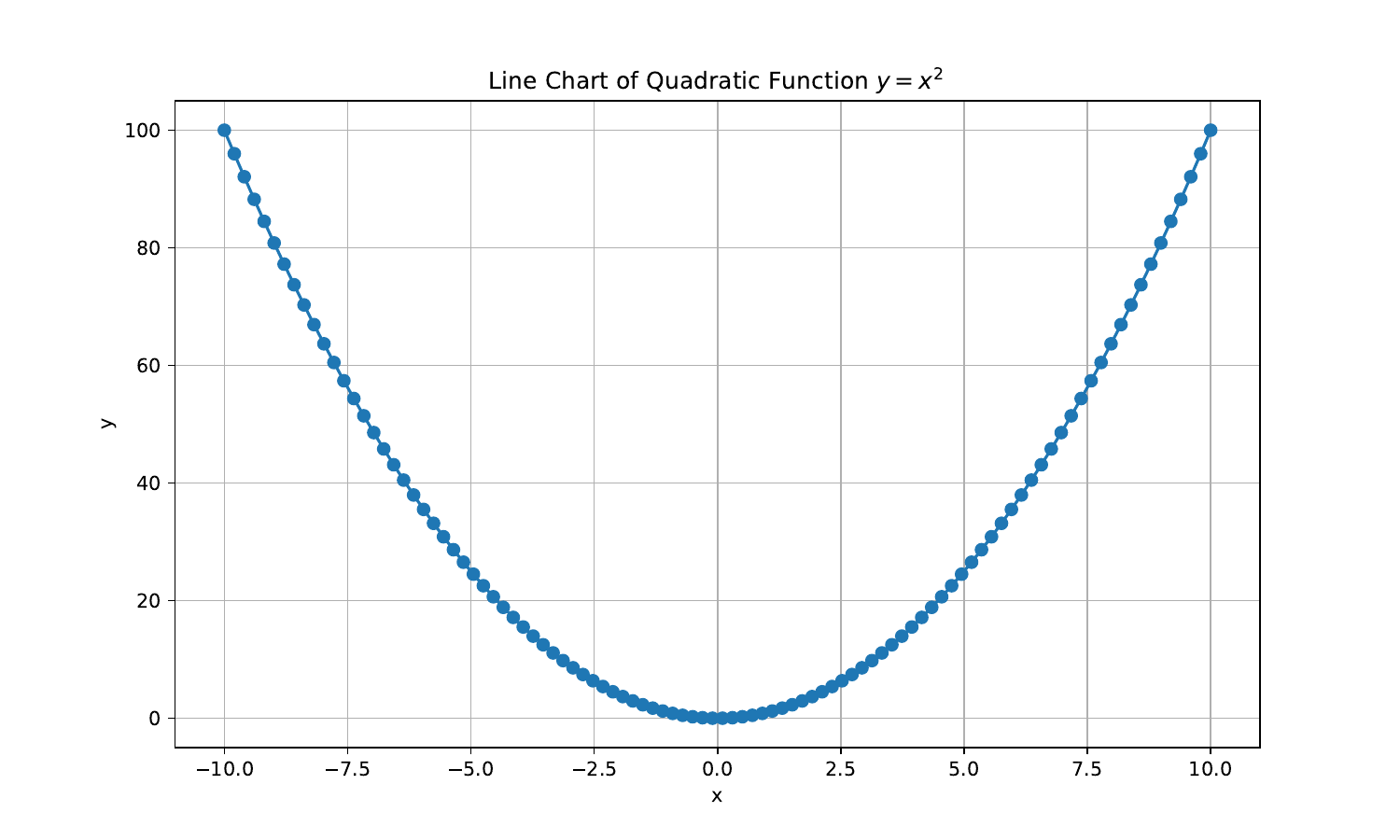}
\end{center}

This chart shows the typical parabolic shape of a quadratic function, demonstrating the squared relationship between the independent variable \( x \) and the dependent variable \( y \).

\subsection{Pie Chart}

A Pie Chart is a circular statistical graphic divided into slices to represent numerical proportions. The size of each slice corresponds to its contribution to the total sum. Pie charts are commonly used in statistics, business, and media to display the relative proportions of different categories within a dataset.

In this section, we will use a pie chart to illustrate the probability distribution of the sum of points obtained when rolling three six-sided dice. Below is the Python code to generate a pie chart that represents the probabilities of different sums when rolling three dice:

\begin{lstlisting}[style=python]
import matplotlib.pyplot as plt
from collections import Counter
import numpy as np

# Calculate the probabilities
sums = [sum(roll) for roll in np.ndindex(6, 6, 6)]
sum_counts = Counter(sums)
total_rolls = 6**3
probabilities = {sum_value: count / total_rolls * 100 for sum_value, count in sum_counts.items()}

# Data for the pie chart
labels = probabilities.keys()
sizes = probabilities.values()

# Plotting the pie chart with a colormap
plt.figure(figsize=(10,10))
cmap = plt.get_cmap('jet')
colors = cmap(np.linspace(0, 1, len(labels)))
plt.pie(sizes, labels=labels, colors=colors, autopct='%1.1f%%', startangle=140)
plt.title('Probability Distribution of Sums for Three Dice Rolls')
plt.show()
\end{lstlisting}

This code simulates the probability distribution of sums when rolling three dice and visualizes it using a pie chart. Each slice represents the probability of a specific sum occurring.

\begin{figure}[H]
    \centering
    \includegraphics[width=1.0\textwidth]{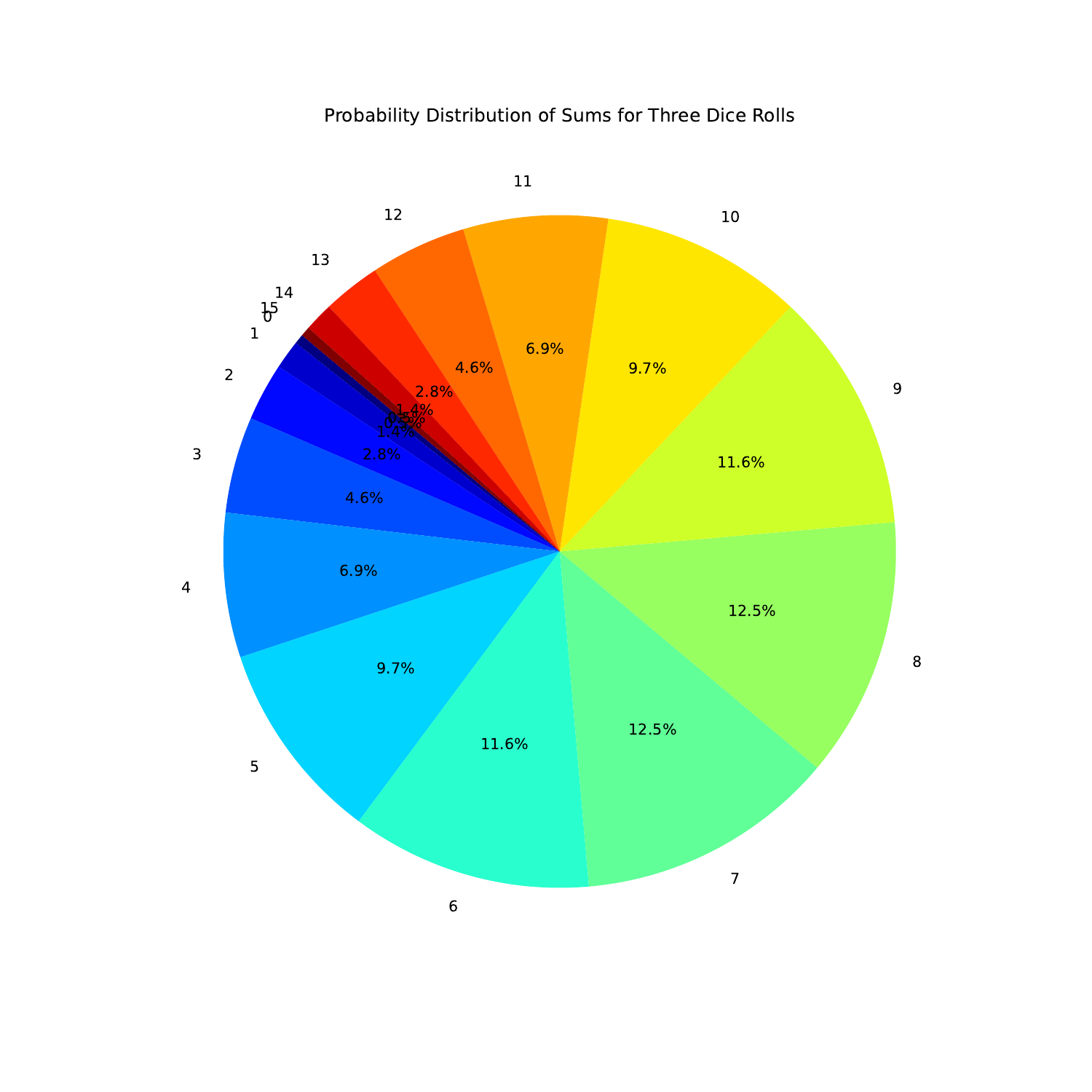}
    \caption{Probability Distribution of Sums for Three Dice Rolls}
\end{figure}

\subsection{Bar Chart}

A Bar Chart is a graphical representation of data using rectangular bars. The length of each bar is proportional to the value it represents. Bar charts are commonly used to compare different categories or to track changes over time.

In this section, we will use a bar chart to illustrate the probability distribution of the sum of points obtained when rolling three six-sided dice.

\begin{lstlisting}[style=python]
import matplotlib.pyplot as plt
from collections import Counter
import numpy as np

# Calculate the probabilities
sums = [sum(roll) for roll in np.ndindex(6, 6, 6)]
sum_counts = Counter(sums)
total_rolls = 6**3
probabilities = {sum_value: count / total_rolls * 100 for sum_value, count in sum_counts.items()}

# Data for the bar chart
labels = list(probabilities.keys())
sizes = list(probabilities.values())

# Plotting the bar chart with a colormap
plt.figure(figsize=(10,6))
cmap = plt.get_cmap('viridis')
colors = cmap(np.linspace(0, 1, len(labels)))
plt.bar(labels, sizes, color=colors)
plt.xlabel('Sum of Dice Rolls')
plt.ylabel('Probability (%)')
plt.title('Probability Distribution of Sums for Three Dice Rolls')
plt.show()
\end{lstlisting}

This code calculates the probability distribution for sums when rolling three dice and visualizes it using a bar chart with colors derived from the ``viridis'' colormap. You can see that the number of points from the three dice combinations conforms to the normal distribution. The use of visualization presents this mathematical law.

\begin{figure}[H]
    \centering
    \includegraphics[width=1.0\textwidth]{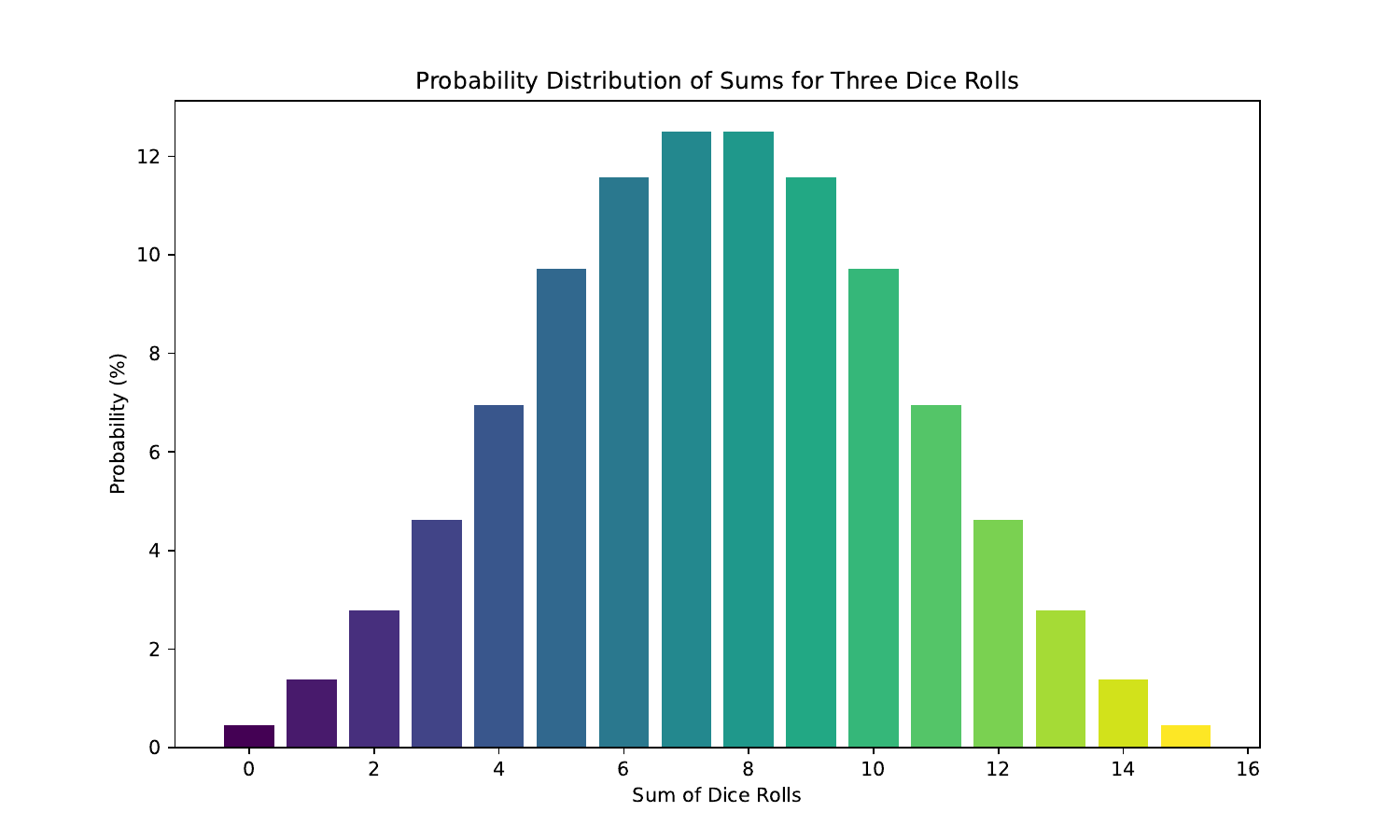}
    \caption{Probability Distribution of Sums for Three Dice Rolls}
\end{figure}

\subsection{Histogram}

A histogram is a commonly used statistical chart to display the distribution of data. It divides the data into continuous intervals (called "bins") and counts the number of data points within each interval (frequency), which reflects the frequency distribution of the data. Unlike a bar chart, the horizontal axis of a histogram represents continuous values, making it suitable for numerical data.

In this example, we simulate rolling five dice 2000 times and use a histogram to show the distribution of the sum of each roll.

\begin{lstlisting}[style=python]
import numpy as np
import matplotlib.pyplot as plt

# Simulate rolling five dice 2000 times
num_dice = 5
num_rolls = 2000
results = np.random.randint(1, 7, (num_rolls, num_dice))
sums = np.sum(results, axis=1)

# Plot the histogram
plt.hist(sums, bins=range(5, 31), edgecolor='black')
plt.title('Sum Distribution of Rolling Five Dice 2000 Times')
plt.xlabel('Sum')
plt.ylabel('Frequency')
plt.show()
\end{lstlisting}

The resulting histogram shows the distribution of the sum of the dice rolls. As seen, the distribution of the sums forms a bell curve, consistent with the Central Limit Theorem.

\begin{figure}[H]
    \centering
    \includegraphics[width=1.0\textwidth]{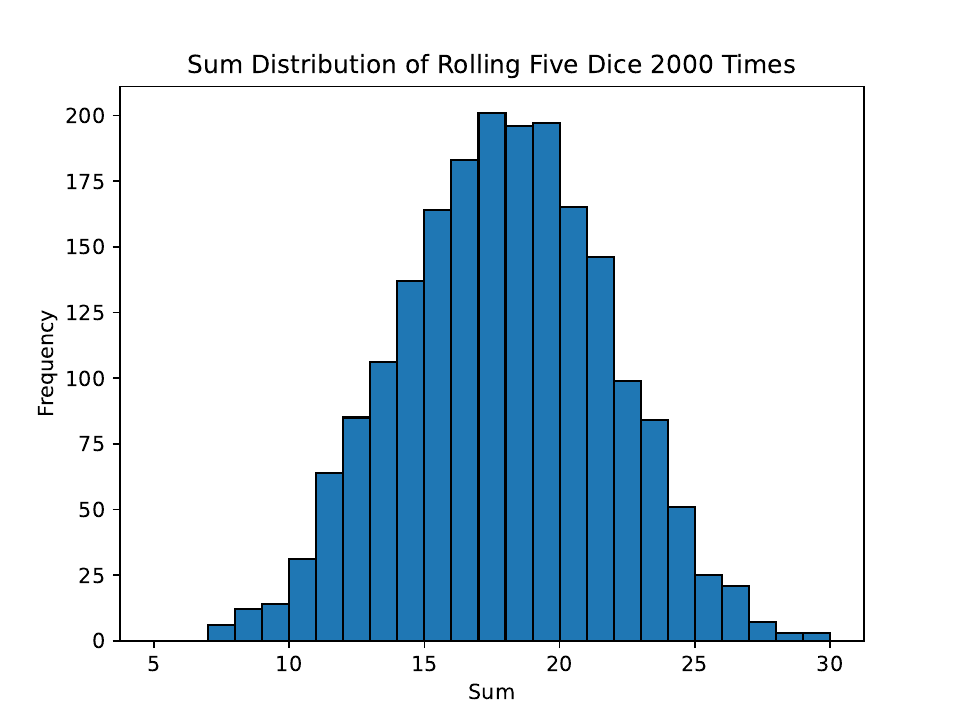}
    \caption{Sum Distribution of Rolling Five Dice 2000 Times}
    \label{fig:histogram}
\end{figure}

Figure \ref{fig:histogram} shows the generated histogram.

\subsection{Parallel Coordinates}

Parallel coordinate is a method used to visualize multi-dimensional data. Each variable (dimension of the data) is represented as a vertical axis, and all axes are arranged parallel to each other. Each data point is represented by a line that connects its value on each axis. By observing these lines, we can analyze the characteristics of different data points across various variables and identify relationships and trends within the data.

\subsection{Iris Dataset}

The Iris dataset~\cite{Fisher1936} is a classic dataset in the field of machine learning. It contains three different species of iris flowers: Iris setosa, Iris versicolor, and Iris virginica. The dataset consists of 150 records, with each record having four features: sepal length, sepal width, petal length, and petal width. This dataset is often used for demonstrating classification algorithms.

The following code uses the Iris dataset to plot a parallel coordinates chart, showing the differences in characteristics between the different species of iris flowers:

\begin{lstlisting}[style=python]
import pandas as pd
import matplotlib.pyplot as plt
from pandas.plotting import parallel_coordinates
from sklearn.datasets import load_iris
import seaborn as sns

# Load the Iris dataset
iris = load_iris()
df = pd.DataFrame(iris.data, columns=iris.feature_names)
df['species'] = iris.target

# Map the species from numbers to names
df['species'] = df['species'].map({0: 'Setosa', 1: 'Versicolor', 2: 'Virginica'})

# Set the aesthetic style of the plots
sns.set(style="whitegrid")

# Define a custom color palette
palette = sns.color_palette("Set2", 3)

# Create a figure with a specific size
plt.figure(figsize=(14, 8))

# Plot the parallel coordinates chart with enhanced aesthetics
parallel_coordinates(
    df,
    'species',
    color=palette,
    linewidth=1.5,
    alpha=0.7
)

# Customize the plot
plt.title('Parallel Coordinates Plot of the Iris Dataset', fontsize=18, fontweight='bold', pad=20)
plt.xlabel('Features', fontsize=14, labelpad=10)
plt.ylabel('Measurements (cm)', fontsize=14, labelpad=10)

# Customize the ticks
plt.xticks(fontsize=12)
plt.yticks(fontsize=12)

# Enhance the legend
plt.legend(title='Species', title_fontsize='13', fontsize='11', loc='upper right', frameon=True)

# Add subtle grid lines
plt.grid(True, which='both', linestyle='--', linewidth=0.5, alpha=0.7)

# Tight layout for better spacing
plt.tight_layout()

# Display the plot
plt.show()
\end{lstlisting}

\begin{figure}[H]
    \centering
    \includegraphics[width=1.0\textwidth]{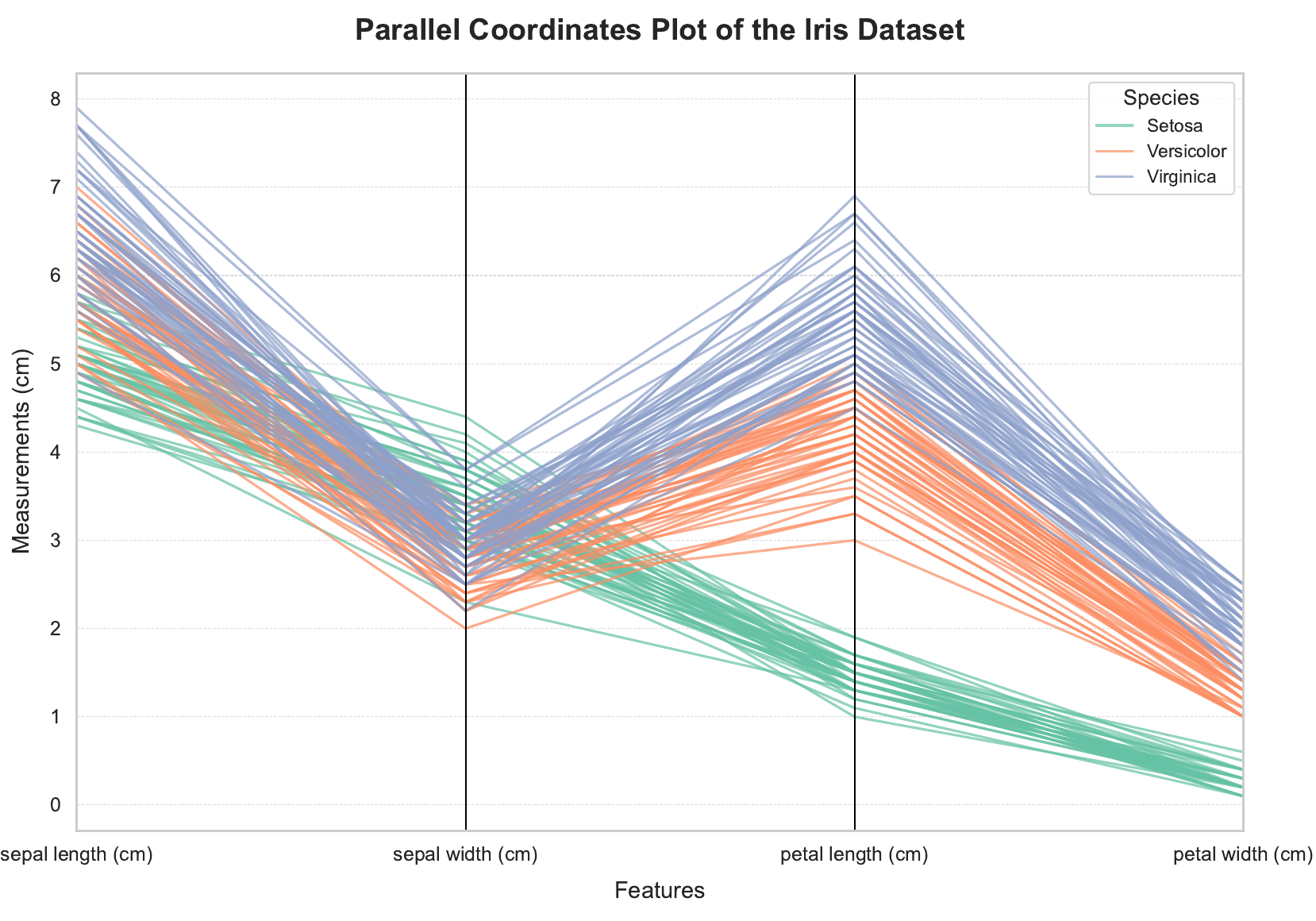}
    \caption{Parallel Coordinates of the Iris Dataset}
    \label{fig:parallel_coordinates}
\end{figure}

Figure \ref{fig:parallel_coordinates} shows the generated parallel coordinates chart, illustrating the differences in characteristics between the various species of iris flowers.

From the parallel coordinates plot, several patterns in the Iris dataset can be observed:

\begin{itemize}
    \item \textbf{Distinctiveness of Setosa}: The Setosa species is separated from the other two species (Versicolor and Virginica) across all feature axes. Specifically, in petal length and petal width, Setosa has significantly lower values, making it easily distinguishable from the other species.
    
    \item \textbf{Overlap Between Versicolor and Virginica}: The lines for Versicolor and Virginica overlap considerably in sepal length and sepal width, making it difficult to differentiate between these two species based on these features alone. However, there are differences in other features that help distinguish them.
    
    \item \textbf{Higher Discriminative Power of Petal Features}: Although Versicolor and Virginica overlap on some features, petal length and petal width provide better separation. For instance, Virginia tends to have larger values in both petal length and width compared to Versicolor.
    
\end{itemize}

In summary, Sentosa is easily distinguishable from the other species based on all features, while Versicolor and Virginica can be more effectively differentiated using petal-related features.

\subsection{Conclusion}

In this section, we explored the importance of data visualization and its critical role in communicating information. By using charts, graphs, and other visual tools, we can transform complex data into easily comprehensible visual representations, allowing readers to quickly and intuitively grasp the key insights. As we have seen, effective visualization not only enhances data readability but also reveals hidden patterns and trends, facilitating deeper analysis.

Moreover, the significance of User Interface (UI) and User Experience (UX) design in the process of data visualization cannot be overlooked. A well-designed UI ensures that the Graphical User Interface (GUI) is simple and intuitive, enabling users to interact with data effortlessly. At the same time, strong UX design guarantees a smooth and satisfying interaction, improving the overall user experience. This synergy between UI, UX, and data visualization enhances the efficiency of information retrieval and creates a more engaging and effective way for users to connect with data.

Together, data visualization, UI/UX, and GUI form a cohesive ecosystem. Through thoughtful interface design and interaction experiences, we can fully harness the potential of data, achieving both efficient information delivery and deeper insights.

In summary, data visualization provides a powerful tool for presenting information more clearly within complex, diverse environments. The integration of UI/UX and GUI further elevates the quality of data presentation and interaction. As technology continues to advance, the methods of visualizing data and designing user interfaces will become more diverse, helping us make better decisions in a data-driven world.

\chapter*{Conclusion}

By studying this book, we believe that you have gained a comprehensive understanding of machine learning, deep learning, and the related hardware and programming environments. From fundamental concepts to implementing specific algorithms, from theory to practice, we have aimed to provide you with a complete learning path. Whether you are a beginner or someone with prior knowledge, this content should help you better grasp and master the cutting-edge technologies in computer science today.

However, this is just the beginning. As technology evolves rapidly, the knowledge and applications in machine learning and deep learning continue to deepen and expand. In future volumes, we will delve further into more advanced techniques, strategies for model optimization, and the various challenges and solutions in real-world applications. We look forward to continuing this learning journey with you as we explore this vast area of technology together.

We hope that by studying this book, you not only gained a fundamental understanding of machine learning and deep learning but also found inspiration for applying them in practical situations. Let this book serve as a cornerstone in your journey into the world of intelligent computing, and may it motivate you to keep moving forward in your future learning and research.

\bibliographystyle{ieeetr}
\bibliography{sample}

\end{document}